\documentclass[manuscript,screen,final]{acmart}

\AtBeginDocument{%
  \providecommand\BibTeX{{%
    \normalfont B\kern-0.5em{\scshape i\kern-0.25em b}\kern-0.8em\TeX}}}

\setcopyright{acmcopyright}
\copyrightyear{2018}
\acmYear{2018}
\acmDOI{XXXXXXX.XXXXXXX}

\usepackage{graphicx}
\graphicspath{{./}} 
\usepackage{subcaption}
\usepackage{booktabs} 
\usepackage{longtable}
\usepackage{wrapfig}
\usepackage{makecell}
\usepackage{comment}
\usepackage{wrapfig}

\usepackage{latexsym}
\usepackage{amsmath} 
\usepackage{mathtools}       
\usepackage{stmaryrd}        
\usepackage{bm}              
\providecommand{\argmax}{\operatornamewithlimits{argmax}} 
\providecommand{\argmin}{\operatornamewithlimits{argmin}} 
\DeclareMathOperator{\Tr}{Tr}     
\DeclareMathOperator{\Cond}{Cond} 
\DeclareMathOperator{\diag}{diag} 
\providecommand{\R}{\mathbb{R}} 
\providecommand{\E}{\mathbb{E}} 
\providecommand{\T}{\mathrm{T}} 
\providecommand{\ind}[1]{\mathbb{I}\{#1\}} 
\renewcommand{\geq}{\geqslant} 
\renewcommand{\leq}{\leqslant} 
\DeclarePairedDelimiterX{\inner}[2]{\langle}{\rangle}{#1, #2}
\DeclarePairedDelimiter{\norm}{\lVert}{\rVert}
\DeclarePairedDelimiter{\abs}{\lvert}{\rvert}
\usepackage{amsthm}
\usepackage{xcolor}
\usepackage{ifthen}
\usepackage{cleveref}
\usepackage[]{hyperref}
\newtheorem{theorem}{Theorem}[section]

\theoremstyle{definition}
\newtheorem{definition}[theorem]{Definition}

\usepackage{algorithm}
\usepackage{algpseudocode}

\DeclareMathOperator{\sign}{sign}

\providecommand{\wracma}{\texttt{WRA-CMA}}
\providecommand{\wraslsqp}{\texttt{WRA-AGA}}
\providecommand{\wracmaadv}{\texttt{WRA-CMA+ADV}}
\providecommand{\wraslsqpadv}{\texttt{WRA-AGA+ADV}}

\providecommand{\acma}{\texttt{ADV-CMA-ES}}
\providecommand{\zopgda}{\texttt{ZO-Min--Max}}
\providecommand{\cmaes}{\texttt{CMA-ES}}
\providecommand{\slsqp}{\texttt{SLSQP}}

\providecommand{\yw}{\hat{y}}
\providecommand{\ywset}{\hat{Y}}

\providecommand{\ny}{N_\omega}
\begin{document}

\title[Covariance Matrix Adaptation Evolutionary Strategy with Worst-Case Ranking Approximation for Min--Max Optimization]{Covariance Matrix Adaptation Evolutionary Strategy with Worst-Case Ranking Approximation for Min--Max Optimization and its Application to Berthing Control Tasks}
\author{Atsuhiro Miyagi}
\email{mygath00@pub.taisei.co.jp}
\orcid{0000-0002-7729-8496}
\affiliation{%
  \institution{Taisei corporation,  \& University of Tsukuba }
  \streetaddress{344-1 Nase}
  \city{Yokohama}
  \state{Kanagawa}
  \postcode{245-0051}
  \country{Japan}}
  
\author{Yoshiki Miyauchi}
\email{yoshiki_miyauchi@naoe.eng.osaka-u.ac.jp}
\orcid{0000-0002-5314-8848}
\author{Atsuo Maki}
\email{maki@naoe.eng.osaka-u.ac.jp}
\orcid{0000-0002-2819-1297}
\affiliation{%
  \institution{Department of Naval Architecture and Ocean Engineering, Graduate School of Engineering, Osaka University}
  \streetaddress{2-1 Yamadaoka}
  \city{Suita}
  \state{Osaka}
  \postcode{565-0971}
  \country{Japan}}
  
\author{Kazuto Fukuchi}
\email{fukuchi@cs.tsukuba.ac.jp}
\orcid{0000-0003-3895-219X}
\author{Jun Sakuma}
\email{jun@cs.tsukuba.ac.jp}
\orcid{0000-0001-5015-3812}
\author{Youhei Akimoto}
\email{akimoto@cs.tsukuba.ac.jp}
\orcid{0000-0003-2760-8123}
\affiliation{%
  \institution{Faculty of Engineering, Information and Systems, University of Tsukuba \& RIKEN Center for Advanced Intelligence Project}
  \streetaddress{1-1-1 Tennodai}
  \city{Tsukuba}
  \state{Ibaraki}
  \country{Japan}
  \postcode{305-8573}
}



\begin{abstract}

In this study, we consider a continuous min--max optimization problem $\min_{x \in \mathbb{X} \max_{y \in \mathbb{Y}}}f(x,y)$ whose objective function is a black-box. We propose a novel approach to minimize the worst-case objective function $F(x) = \max_{y} f(x,y)$ directly using a covariance matrix adaptation evolution strategy (CMA-ES) in which the rankings of solution candidates are approximated by our proposed worst-case ranking approximation (WRA) mechanism. 
We develop two variants of WRA combined with CMA-ES and approximate gradient ascent as numerical solvers for the inner maximization problem. 
Numerical experiments show that our proposed approach outperforms several existing approaches when the objective function is a smooth strongly convex--concave function and the interaction between $x$ and $y$ is strong. We investigate the advantages of the proposed approach for problems where the objective function is not limited to smooth strongly convex--concave functions. The effectiveness of the proposed approach is demonstrated in the robust berthing control problem with uncertainty.

\end{abstract}

\maketitle
\sloppy

\section{Introduction} \label{sec:intro}

\paragraph{Background}

Simulation-based optimization is an attractive technique in various industrial fields. 
Given a design vector $x \in \mathbb{X} \subseteq 
\R^{d_x}
$, the objective function $h_{\mathrm{sim}} : \mathbb{X} \to 
\R
$ is evaluated via numerical simulation. 
Simulation-based optimization has been used in several real-world applications, such as berthing control \cite{maki2020, Miyauchi2022}, well placement \cite{miyagi@ghgt, bouzarkouna2012, Onwunalu2010}, and topology design \cite{fujii2018, marsden2004}. 
To perform simulation-based optimization, it is necessary to determine simulation conditions in advance so that the objective function values in the real-world, $h_{\mathrm{real}}(x)$, are accurately computed. 
In other words, a simulator such that $h_{\mathrm{sim}}(x) \approx h_{\mathrm{real}}(x)$ must be developed. 
However, owing to some real-world uncertainties, the predetermined conditions often contain errors and hence $h_{\mathrm{sim}}(x)$ does not approximate $h_{\mathrm{real}}(x)$ well \cite{Oberkampf2002,walker2003,bouzarkouna2012phd,chen2013,Anna2018}. 
In such situations, there is a risk that the optimal solution obtained in simulation-based optimization, $x_{\mathrm{sim}} = \argmin_{x \in \mathbb{X}} h_{\mathrm{sim}}(x)$, does not perform well in the real-world and results in $h_{\mathrm{real}}(x_\mathrm{sim}) \gg h_{\mathrm{sim}}(x_\mathrm{sim})$.

One approach to find a robust solution is to formulate the problem as a min--max optimization problem 
\begin{align}
	&\min_{x \in \mathbb{X}} \max_{y \in \mathbb{Y}} f(x, y) , \label{eq:minmax} 
\end{align}
where $f(x,y)$ denotes the objective function and $y \in \mathbb{Y} \subseteq 
\R^{d_y}
$ is a parameter vector for the simulation conditions, called a 
\emph{scenario vector}
in this study, and is uncertain at the optimization stage. 
This approach aims to find the 
\emph{global min--max solution}
$x^*=\argmin_{x \in \mathbb{X}} F(x)$ that minimizes the worst-case objective function $F(x):=\max_{y \in \mathbb{Y}} f(x, y)$. 
In this formulation, the simulator designed by an expert engineer, $h_{\mathrm{sim}}$, corresponds to $f(\cdot, y_{\mathrm{sim}})$ with a scenario vector $y_\mathrm{sim} \in \mathbb{Y}$, and the real-world objective, $h_\mathrm{real}$, corresponds to $f(\cdot, y_\mathrm{real})$ with a scenario vector $y_\mathrm{real} \in \mathbb{Y}$, which is unknown and may change over time.
Minimizing the worst-case objective function $F$ minimizes the upper bound of the objective function values in the real-world, i.e., $F(x) \geq f(x,y_\mathrm{real})$ provided $y_\mathrm{real} \in \mathbb{Y}$.

In this study, we focus on simulation-based optimization where 
the gradient of the objective function $f$ with respect to $x$ and $y$ is unavailable, and 
the objective function $f$ and worst-case objective function $F$ are nonexplicit (black-box functions). 
We refer to such a problem as a 
\emph{black-box min--max optimization}
. 
In particular, we focus on the following two types of problems, for which existing approaches \cite{akimoto2022berthing,Liu2020} for the black-box min--max optimization fail to converge or exhibit slow convergence.
\begin{description}
\item[(A)] $f$ is smooth and strongly convex--concave around $x^*$, but a strong interaction between $x$ and $y$ exists.
\item[(B)] $f$ is not smooth or strongly convex--concave around $x^*$. 
\end{description}
These difficulties are not well addressed in existing approaches. 
However, it does not necessarily mean these problems are not important to address. 
Indeed, it has been reported in \cite{Razaviyayn2020, Bertsimas2010} that the objective function in real-world applications is often not convex--concave, i.e., falls into problem of Type (B). Moreover, because the strength of the interaction term can not be known in advance, we consider approaches to the black-box min--max optimization should be able to treat such interaction, just like that approaches to black-box optimization should be able to treat highly ill-conditioned problems.


\begin{figure*}[t]
    \centering
    \begin{subfigure}{0.48\hsize}%
    \includegraphics[width=\hsize]{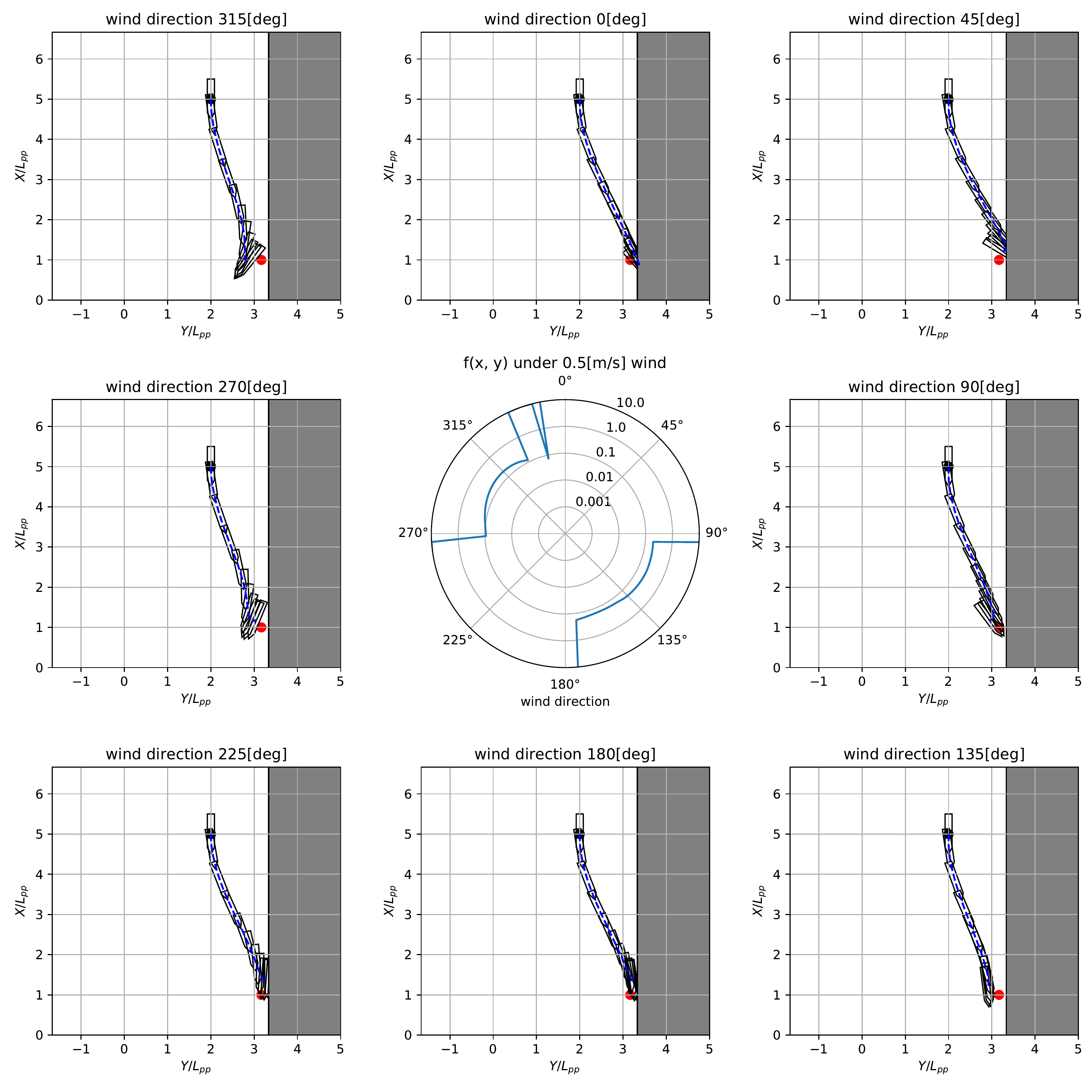}%
    \caption{Controller optimized for no wind condition}%
    \label{fig:berthing:nowind}%
    \end{subfigure}%
    \hfill
    \begin{subfigure}{0.48\hsize}%
    \includegraphics[width=\hsize]{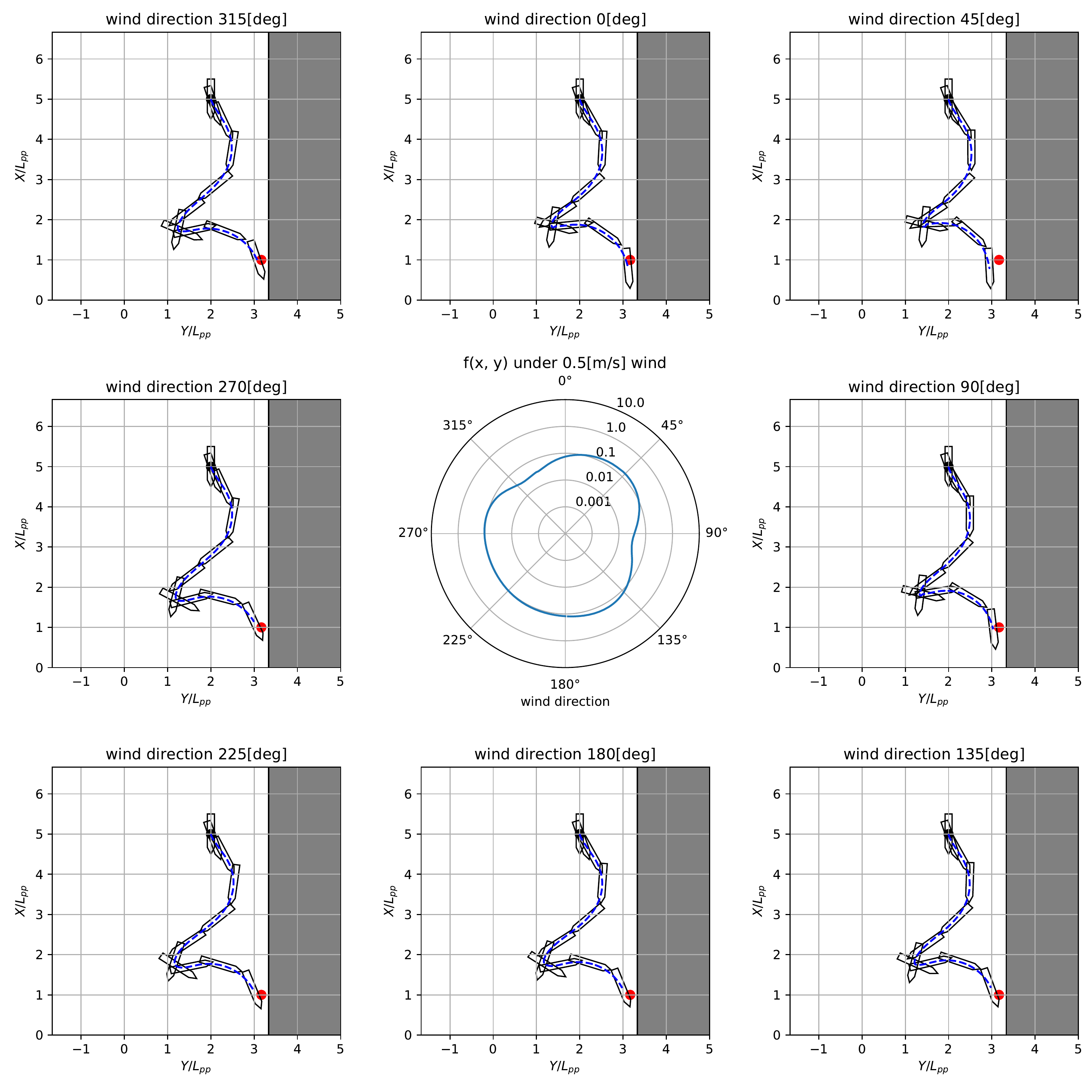}%
    \caption{Controller optimized for the worst wind condition}%
    \label{fig:berthing:advcma}%
    \end{subfigure}%
    \caption{%
    Visualization of trajectories obtained by controllers for (a) no wind condition and (b) the worst wind condition with a $0.5$-[m/s] maximum wind velocity. Center: the objective function values $f(x, y)$ under $y$ representing a wind velocity of $0.5$ [m/s] and varying wind direction presented by the polar axis. An objective function value smaller than $10$ implies that the ship is controlled without a collision with the berth.
    The others: trajectories observed under wind for $0, 45, \dots, 315$ [deg] with a velocity of $0.5$ [m/s]. The red points are the target positions. The controllers were obtained in \cite{akimoto2022berthing}.
    }%
    \label{fig:berthing-trajectory-adv}
\end{figure*}

\paragraph{Robust berthing control problem}

As an example of real-world applications, we consider the automatic berthing control problem \cite{maki2020, Miyauchi2022}. 
The objective is to obtain a controller that realizes a fine control of a ship toward a target state with the least collision risk and minimum elapsed time. 
Given a controller parameterized by $x$, a trajectory of the ship motion is simulated by numerically solving a ship maneuvering model. 
The objective function values are evaluated based on the computed trajectory. 
However, this simulation contains some uncertainties.

In a previous study \cite{akimoto2022berthing}, the problem of finding a robust berthing controller was formulated as a min--max optimization problem. 
\Cref{fig:berthing-trajectory-adv} shows the importance of considering the uncertainties. \Cref{fig:berthing:nowind} and \Cref{fig:berthing:advcma}, respectively, show the trajectories and objective function values under different wind conditions when a controller optimized by the (1+1)-covariance matrix adaptation evolution strategy (CMA-ES) 
\cite{Arnold2010, igel2006}
under no wind disturbance and a controller optimized by Adversarial CMA-ES (\acma{}) \cite{akimoto2022berthing}, where the uncertainty of wind direction ($[0, 360]$ [deg]) and wind speed ($[0, 0.5]$ [m/s]) are considered. 
When the controller optimized under the no wind assumption is used, we often observe the collision of the ship and berth under wind disturbance of a velocity of $0.5$ [m/s]. Meanwhile, the robust controller obtained by \acma{} successfully avoids collision for all wind directions.

The robust berthing control problem is expected to fall into Type (B).
Its vestiges can be seen in  \Cref{fig:berthing-trajectory-adv}. The central figure shows that the objective function $f(x, y)$ is multimodal with respect to the wind direction. Therefore, $f(x, y)$ is nonconcave in terms of $y$. 
Moreover, we observed in our preliminary experiments that the worst-case scenario switches between offshore-to-berth wind $y_{\mathrm{sea}}$ and berth-to-offshore wind $y_{\mathrm{land}}$. 
This is explained as follows. 
In offshore-to-berth wind $y_{\mathrm{sea}}$, the optimum control avoids getting too close to the berth to avoid a collision. Under such control, berth-to-offshore wind $y_{\mathrm{land}}$ becomes the worst-case scenario because the ship stops at a position far from the target position near the berth, resulting in a high objective function value. Conversely, if the optimum control for $y_{\mathrm{land}}$ is operated, the worst-case scenario is $y_{\mathrm{sea}}$, which causes the ship to collide with the berth. 
Therefore, the control that minimizes the objective function at the worst-case scenario is expected to exist on the boundary of the regions where the worst-case scenario changes between $y_{\mathrm{land}}$ and $y_{\mathrm{sea}}$, and it is not the optimal solution under either scenario.

\paragraph{Related works}

Recently, gradient-based min--max optimization has been actively studied.
However, most existing studies investigate the min--max problems of functions that are concave in $y$, although several real-world problems are not necessarily concave in $y$ \cite{Razaviyayn2020}. 
In addition, a general nonconvex--nonconcave min--max problem is theoretically intractable \cite{minmaxcomplexity}. 
Some studies have been conducted to identify the structures of nonconcave min--max problems that make it efficiently solvable \cite{nonconvexpl,nonconcaveweakmvi,nonconcaveweakconcave,nonconcavehidden} or to exploit a small domain for scenario vectors \cite{nonconcavesmall}. 
These studies do not cover Type (B). 
Gradient-based approaches for general nonconcave min--max problems, where both implementation error and parameter uncertainty are considered, have been developed in previous studies \cite{Bertsimas2010, Bertsimas2010b}. 
However, this approach is designed to exploit the existence of implementation error, and it is not trivial to extend it to derivative-free situations via gradient approximation.

Derivative-free approaches for min--max optimization include coevolutionary approaches \cite{Barbosa1999, Herrmann1999,Qiu2018, Abdullah2019}, simultaneous descent--ascent approaches \cite{akimoto2022berthing,Liu2020}, and surrogate-model-based approaches \cite{Bogunovic2018}. 
Particularly, \acma{} \cite{akimoto2022berthing} and \zopgda{} \cite{Liang2019} are theoretically guaranteed to converge to the optimal solution and its neighborhood, respectively, in smooth strongly 
convex--concave
min--max problems. 
Nevertheless, they fail to converge in Type (B) and exhibit slow convergence in Type (A) \cite{akimoto2022berthing}. 
Although some coevolutionary approaches, such as minimax differential evolution  \cite{Qiu2018}, are intended to address the difficulty in Type (B), they fail to converge not only on such problems but also on strongly convex--concave problems \cite{akimoto2022berthing}. STABLEOPT \cite{Bogunovic2018}, a Bayesian optimization approach, is expected to address the difficulty in Type (B). However, because of the high computational time of the Gaussian process, it is impractical for problems where numerous $f$-calls are required to obtain satisfactory solutions, according to \cite{Liu2020}. 

\paragraph{Contributions}

The contributions of this study are as follows.

\emph{Approach (\Cref{sec:proposed})}.
    Aiming at addressing the limitations of existing approaches observed in Types (A) and (B), a novel approach that directly searches for the global min--max solution is proposed. 
    The proposed approach minimizes the worst-case objective function using CMA-ES \cite{Hansen2001,Hansen2014,akimoto2019} wherein the rankings of solution candidates are approximated by our proposed worst-case ranking approximation (WRA) mechanism. 
    The WRA mechanism approximately solves the maximization problem $\max_{y \in \mathbb{Y}}f(x,y)$ for each solution candidate $x$. 
    To save $f$-calls required to solve each maximization problem, we design a 
    warm-starting
    strategy and an early-stopping strategy. 
    We propose two variants of WRA implementations using CMA-ES and approximate gradient ascent (AGA) as inner solvers.
    To consider nonconvex real-world applications, we incorporate a restart strategy and a local search strategy. 

\emph{Evaluation (\Cref{sec:test})}.
    We designed $11$ test problems with different characteristics, including both Types (A) and (B). Numerical experiments on the $11$ test problems reveal the limitations of existing approaches and show that the proposed approach can handle both Types (A) and (B). 
    We experimentally show that the scaling of the runtime on a smooth strongly convex--concave with respect to the interaction term (denoted by $b$ in \Cref{sec:test}) is significantly improved over existing approaches.
    To understand when the proposed approach is effective in Type (B), we investigate the effect of each component of the WRA mechanism on each of the following situations: (S) the global min--max solution $x^*$ is a strict min--max saddle point, (W) $x^*$ is a weak min--max saddle point, and (N) $x^*$ is not a min--max saddle point. 

\emph{Application (\Cref{sec:berthing})}.
    The proposed approach and existing approaches are applied to the robust berthing control problem with three types of scenario vectors.
    In the cases where the wind direction is included in a scenario vector, we confirm that
    the proposed approach often obtains controllers that can avoid collision with the berth in the worst-case scenario, whereas the controllers obtained by the existing approaches tend to fail to avoid collision with the berth in the worst-case scenario.
    In the case where an existing approach can often obtain collision-free controllers, we confirm that the existing approach achieves better worst-case performance than the proposed approach.
    We also demonstrate the effect of a hybrid of the existing and proposed approaches.
    
\emph{Implementation}.
    Our implementations are publicly available.
    \footnote{
    Hidden for blind review.
    }

\begin{figure}[t]
\centering
\includegraphics[width=0.5\hsize]{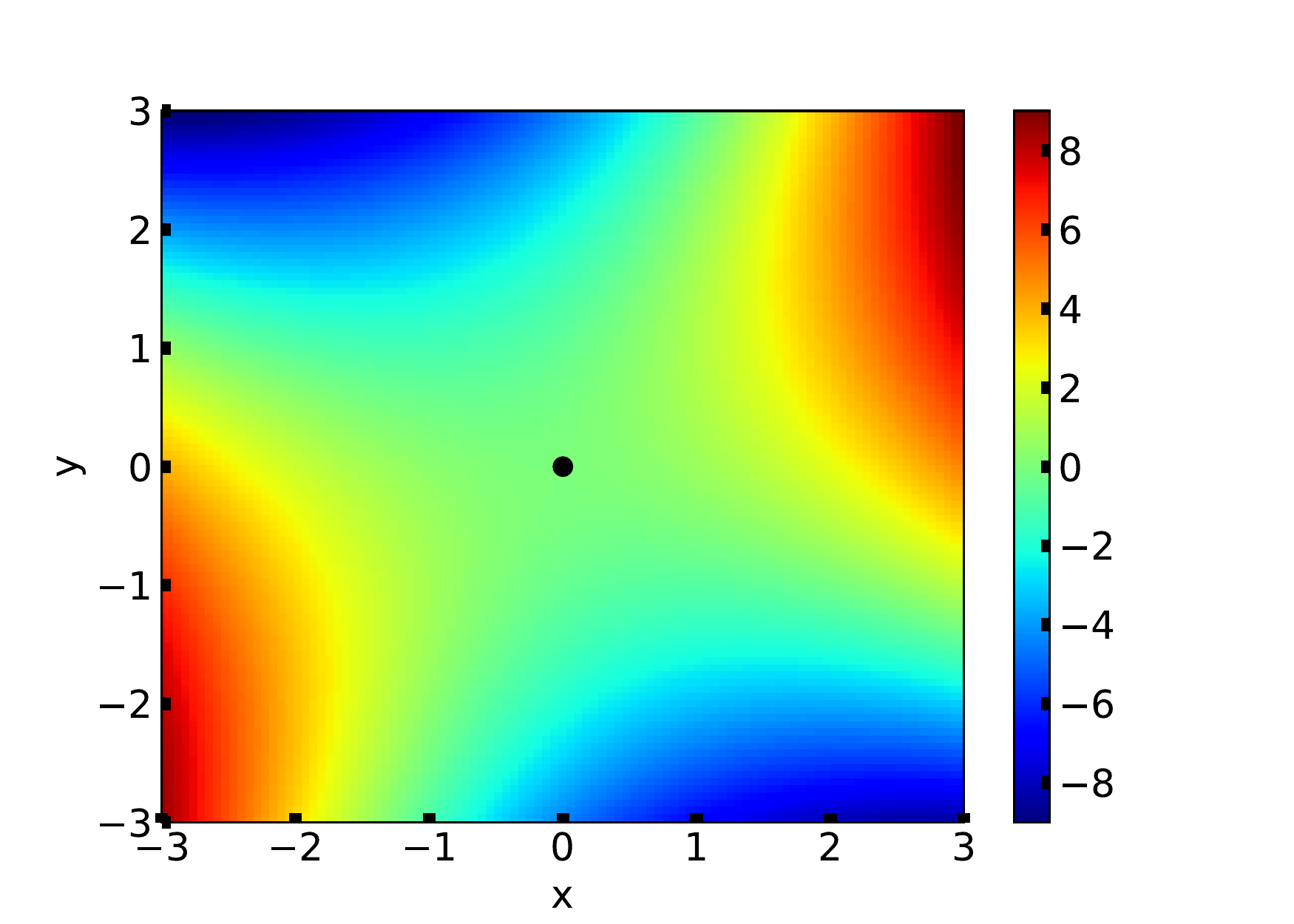}%
\caption{
Landscape of a smooth and strictly convex-concave function $(f(x,y)=\frac{1}{2} \norm{x}_2^2 + x^\T y - \frac{1}{2} \norm{y}_2^2)$ when $d_x=d_y=1$. A black point at $(x,y)=(0,0)$ is the global min--max saddle point. The global min--max saddle point $(x^*, y^*)$ is such that $x^*$ is the global minimum point of $f(x, y^*)$ ($= \frac12 \norm{x}_2^2$ in this case) and $y^*$ is the global maximum point of $f(x^*, y)$ ($= - \frac12 \norm{y}_2^2$ in this case). 
}%
\label{fig:saddle}
\end{figure}

\section{Preliminaries}\label{sec:formulation}

The objective of this study is to find the global minimum solution to the worst-case objective function $F$ defined as follows:
\begin{align}
	F(x) = \max_{y \in \mathbb{Y}} f(x, y) , \label{eq:Fx}
\end{align}
where $f:\mathbb{X} \times \mathbb{Y} \rightarrow 
\R
$ denotes the objective function,
$\mathbb{X} \subseteq 
\R^{d_x}
$ denotes the search domain for \emph{design vector} $x$,
and $\mathbb{Y} \subseteq 
\R^{d_y}
$ denotes the search domain for \emph{scenario vector} $y$.
We assume that $f$ and $F$ are black boxes and their gradient information and higher order information are unknown.
For each $x \in \mathbb{X}$, let $\ywset(x) = \{y \mid F(x)=f(x, y)\} = \argmax_{y \in \mathbb{Y}} f(x, y)$ be the \emph{worst-case scenario set}. 
If $\ywset(x)$ is a singleton, i.e., $\abs{\ywset(x)} = 1$, then the unique element is called the \emph{worst-case scenario} and is denoted by $\yw(x)$.
The global minimum solution of $F$ is called the \emph{global min--max solution} of $f$ and is denoted by $x^* = \argmin_{x \in \mathbb{X}} F(x)$.

One possible approach is to model uncertainty with a finite number of scenarios $S=\{y_1,\dots,y_{s}\}$ by discretizing the space $\mathbb{Y}$.
In this case, the min-max problem can be formulated as $\min_{x \in \mathbb{X}} \max_{y \in S} f(x,y)$, and this formulation is employed in many applications, particularly in geo-science field \cite{bouzarkouna2012phd,Yeten2003spe, minmax, miyagi2023}.
However, in this formulation, the worst-case function $F_s:=\max_{y \in S} f(x,y)$ significantly depends on the discretization method of the space $\mathbb{Y}$ and the number of scenarios $\abs{S}$. Therefore, the performance of the optimal solution on $F_s$ may be significantly degraded on the true worst-case function $F$, as has been demonstrated in a previous study \cite{akimoto2022berthing} for the above-mentioned robust berthing control problem.
Therefore, we focus on solving \eqref{eq:minmax} without discretization in this work.

The \emph{min--max saddle point} (\Cref{def:saddlep}) is an essential concept that characterizes the difficulties in obtaining $x^*$. 
An example of the \emph{min--max saddle point} is visualized in \Cref{fig:saddle}.
In what follows, a neighborhood of design vector $x$ is a set $\mathcal{E}_x$ such that there exists an open ball $\mathbb{B}(x, r)=\{\tilde{x} \in 
\R^{d_x}
\mid \norm{x-\tilde{x}}<r\}$ included in $\mathcal{E}_x$ as a subset.
We analogously define a neighborhood $\mathcal{E}_y$ of scenario vector $y$.
A 
\emph{critical point} 
$(x, y) \in \mathbb{X}\times\mathbb{Y}$ of the objective function $f$ is such that $\nabla f (x,y) = (\nabla_x f (x,y), \nabla_y f (x,y)) = 0$. 
\begin{definition}[min--max saddle point]\label{def:saddlep}
A point $(\tilde{x}, \tilde{y}) \in \mathbb{X} \times \mathbb{Y}$ is a local min--max saddle point of a function $f:\mathbb{X} \times \mathbb{Y} \rightarrow 
\R
$ if there exists a neighborhood $\mathcal{E}_x \times \mathcal{E}_y \subseteq \mathbb{X} \times \mathbb{Y}$ including $(\tilde{x}, \tilde{y})$ such that for any $(x,y) \in \mathcal{E}_x \times \mathcal{E}_y$, the condition $ f(\tilde{x}, y) \leq f(\tilde{x}, \tilde{y}) \leq f(x,\tilde{y})$ holds. If $\mathcal{E}_x = \mathbb{X}$ and $\mathcal{E}_y = \mathbb{Y}$, the point $(\tilde{x}, \tilde{y})$ is called the global min--max saddle point. If the equality holds only if $(x,y) = (\tilde{x}, \tilde{y})$, it is called a strict min--max saddle point. A saddle point that is not a strict min--max saddle point is called a weak min--max saddle point.
\end{definition}

We focus on whether $x^*$ is a strict min--max saddle point. 
If so, the problem of locating $x^*$ turns into the problem of locating the global min--max saddle point. 
In such a situation, locating multiple local min--max saddle points, $\{(x_i, y_i)\}_{i=1}^{K}$, and selecting the best, $\argmin_{1\leq i \leq K} \max_{1\leq j \leq K}f(x_i, y_j)$, can offer the optimal solution $x^*$ provided the global min--max saddle point is included in $\{(x_i, y_i)\}_{i=1}^{K}$. 
Existing approaches \cite{akimoto2022berthing,Liu2020} for locating local min--max saddle points may be used for this purpose.
However, if $x^*$ is not a strict min--max saddle point, the above approach may not provide a reasonable solution; a different approach is required.

\providecommand{\btx}{z}
\begin{table*}[]
    \centering\small
    \caption{%
    Test problem definitions and their worst-case scenarios. 
    The search domains for $x$ and $y$ are $\mathbb{X} = [\ell_x, u_x]^{d_x}$ and $\mathbb{Y} = [-b_y, b_y]^{d_y}$, respectively. 
    The interaction between $x$ and $y$ is controlled by ${d_x} \times {d_y}$ matrix $B$.
    For $f_3$, we assume that $B$ is of full column rank, and let $B^\dagger = [b_1^\dagger, \dots, b_{d_x}^\dagger]$ be the Moore--Penrose inverse of $B$, $\alpha = -\frac{\min(\abs{u_x}, \abs{\ell_x})}{ (30/7) \max_{i=1,\dots,{d_x}} \norm{b_i^\dagger}_1}$, and $\gamma > 0$.
    For $f_9$, we set ${d_y*} = \min\{{d_y}, 3\}$.
    For $f_{10}$, we assume ${d_x} = {d_y}$ and $B = \diag(1, \dots, 1)$.
    The optimal solutions for the worst-case objective functions are $x^* = \bm{0}$, except for $f_1$, $f_3$, and $f_9$. The optimal solution is 
    $[B x^*]_{i} = 0$ for $f_1$, 
    $[B x^*]_{i} = \alpha$ for $f_3$, 
    and $[B^\T x^*]_{i} = - \sinh(1)$ for $i \leq {d_y*}$ and $[B^\T x^*]_{i} = 0$ for $i > {d_y*}$ for $f_9$.
   Here, $[x]_i$ denotes the $i$th coordinate of a vector $x$. 
    }
    \label{tab:testp}
    \begin{tabular}{ll}
    \toprule
    Definition & 
    $[\yw(x)]_i$ (here $\btx = B^{\T} x$ for short)
    \\
    \midrule
    
    $f_1 = x^{\T} B y$ & 
    $\begin{cases} b_y \sign([z]_i ) & [z]_i \neq 0 \\ \text{arbitrary} & [z]_i = 0 \end{cases}$
    \\

    $f_2 = \frac12 \norm{x}_2^2 + x^\T B y$ & 
    $\begin{cases} b_y \sign([z]_i ) & [z]_i \neq 0 \\ \text{arbitrary} & [z]_i = 0 \end{cases}$
    \\

    $f_3 = \frac{1}{2} \norm{B^\T x - (\alpha - \gamma b_y) \bm{1}_n}_2^2 + \gamma x^\T B y$ & 
    $\begin{cases} b_y \sign([z]_i ) & [z]_i \neq 0 \\ \text{arbitrary} & [z]_i = 0 \end{cases}$
    \\        
        
    $f_4 = \frac{1}{2} \norm{x}_2^2 + x^\T B y + \frac{1}{2} \norm{y}_2^2$ & 
    $\begin{cases} b_y \sign([z]_i ) & [z]_i \neq 0 \\ \pm b_y & [z]_i = 0 \end{cases}$    \\

    $f_5 = \frac{1}{2} \norm{x}_2^2 + x^\T B y - \frac{1}{2} \norm{y}_2^2$ & 
    $\begin{cases}
    [z]_i & \abs{[z]_i} \leq b_y \\
    b_y \sign([z]_i) & \abs{[z]_i} > b_y
    \end{cases}$ 
    \\

    $\textstyle f_6 = \frac12 \norm{x}_2^2 + \norm{x}_1 + x^\T B y - \norm{y}_1 - \frac12 \norm{y}_2^2$ & 
    $\begin{cases}
    0 & \abs{[z]_i} \leq 1 \\
    [z]_i - \sign([z]_i)  & 1 < \abs{[z]_i} \leq b_y + 1 \\
    b_y \sign([z]_i) & b_y + 1 < \abs{[z]_i}
    \end{cases}$ 
    \\
    
    $f_7 = \frac14 \norm{x}_2^4 + x^\T B y - \frac14\norm{y}_2^4$ & 
    $\begin{cases}
    \frac{[z]_i}{\norm{z}_2^{2/3}}  & \frac{[z]_i}{\norm{z}_2^{2/3}} \leq b_y \\
    b_y \sign([z]_i) & \frac{[z]_i}{\norm{z}_2^{2/3}} > b_y
    \end{cases}$
    \\

    $f_8 = \norm{x}_1 + x^\T B y - \norm{y}_1$ & 
    $
    \begin{cases}
    0 & \abs{[z]_i} \leq 1 \\
    b_y \sign([z]_i) & \abs{[z]_i} > 1 
    \end{cases}
    $ 
    \\

    $\begin{aligned}
    f_9 &= \textstyle \sum_{i=1}^{d_y*} \left([B^\T x]_i + \exp\left(\sign([y]_i)\right) \cdot \sin\left( \frac{ \pi [y]_i}{b_y}\right) \right)^2  \\
    &\quad\textstyle + \sum_{i={d_y*}+1}^{d_y} \left([B^\T x]_i^2 - [y]_i^2 \right)
    \end{aligned}$ & 
    $
    \begin{cases}
    (b_y / 2)  & [z]_i \geq - \sinh(1) \ \&\ i\leq {d_y*}\\
    - (b_y / 2)  & [z]_i \leq - \sinh(1) \ \&\ i\leq {d_y*}\\
    0 &  i > {d_y*}
    \end{cases}
    $
    \\

    $f_{10} = \norm{x^\T B}_2^2 - 2 \norm{y - x^\T B}_2^{2} $ & 
    $
    \begin{cases}
    [z]_i & \abs{[z]_i} \leq b_y \\
    b_y \sign([z]_i) & \abs{[z]_i} > b_y
    \end{cases} 
    $
    \\

    $f_{11} = \sum_{i=1}^{d_y} \left(\frac{1}{2} [x]_i^2 + 10^{-3i/{d_y}} [x^\T B]_i [y]_i - \frac{10^{-6i/{d_y}}}{2} [y]_i^2 \right)$ & 
    $
    \begin{cases}
    10^{3i/{d_y}}[z]_i & \abs{10^{3i/{d_y}}[z]_i} \leq b_y \\
    b_y \sign([z]_i) & \abs{10^{3i/{d_y}}[z]_i} > b_y
    \end{cases} 
    $
    \\

    \bottomrule
    \end{tabular}
\end{table*}

\section{Test Problems}\label{sec:testp}
\Cref{tab:testp} lists the test problems used in our experiments. 
Although it is difficult to formally frame our target problems as our approach is a heuristic, this list provides examples of problems in the scope of this study. We describe the characteristics of these problems in the following.

The functions $f_3$, $f_5$, $f_6$, $f_7$, $f_8$, and $f_{11}$ are strictly convex--concave.
On such problems, the worst-case scenario set $\ywset(x)$ is a singleton for each $x \in \mathbb{X}$.
The global min--max solution $x^*$ is the strict global min--max saddle point $(x^*, \yw(x^*))$. 
Particularly, $f_5$, $f_6$, and $f_{11}$ are strongly convex--concave, and $f_5$, $f_7$, and $f_{11}$ are smooth.
The functions $f_5$ and $f_{11}$ are both smooth and strongly convex--concave, where the convergence of the existing approaches is investigated. 
Different from $f_5$ and the other functions, $f_{11}$ is designed to be highly ill-conditioned in $y$ to demonstrate the impact of ill-conditioning.
Although $x^*$ is a strict global min--max saddle point, based on our experiments, the existing approaches fail to converge if the objective function is nonsmooth ($f_{6}$ and $f_{8}$) or exhibit slow convergence if the objective function is not strongly convex--concave ($f_{7}$). 

The functions $f_1$, $f_2$, and $f_3$ are convex--linear. 
On such problems, the worst-case scenario is typically located at the boundary of the scenario domain $\mathbb{Y}$. 
On $f_1$ and $f_2$, where the former is bilinear and the latter is strongly convex in $x$, the global min--max solution $x^*$ forms a weak min--max saddle point $(x^*, y)$ for any $y \in \mathbb{Y}$. 
Hence, the worst-case scenario set at $x^*$ is $\ywset(x^*) = \mathbb{Y}$, but the worst-case scenario $\yw(x)$ in a neighborhood of $x^*$ is one of the $2^{d_y}$ vertices $\bar{Y}$ of $\mathbb{Y}$. 
For $f_3$, $\abs{\ywset(x^*)} = 1$ and $(x^*, \yw(x^*))$ is a strict global min--max saddle point.

The functions $f_4$, $f_9$, and $f_{10}$ are not convex--concave.
On these problems, the global min--max solution $x^*$ does not form a min--max saddle point. 
Similar to $f_1$ and $f_2$, the worst-case scenarios of $f_4$ are located at the vertices $\bar{Y}$ of $\mathbb{Y}$.
However, different from $f_1$ and $f_2$, $\ywset(x^*) = \bar{Y}$ and $x^*$ does not form a min--max saddle point in $f_4$. 
For $f_9$, the worst-case scenarios are not at the vertices of $\mathbb{Y}$ but at some specific points inside $\mathbb{Y}$, and $\abs{\ywset(x^*)} > 1$.
These two functions are multimodal in $y$ for each $x \in \mathbb{X}$, and the global maximum (i.e., the worst-case scenario) changes depending on $x$. 
Different from $f_4$ and $f_9$, $f_{10}$ is concave in both $y$ and $x$. 
Because of the concavity in $y$, we have $\abs{\ywset(x)} = 1$ for all $x \in \mathbb{X}$.
Moreover, $\yw(x)$ is continuous. 
The worst-case objective function $F$ is convex around $x^*$. 
However, $x^*$ is not a min--max saddle point.

We focus on some characteristics related to the difficulty in approximating the local landscape of the worst-case objective function $F$. 
A characteristic common to $f_1$--$f_4$ and $f_9$ is that the worst-case scenario changes discontinuously. 
Particularly for $f_1$, $f_2$, $f_4$, and $f_9$, the worst-case scenarios spread over multiple distant points in a neighborhood of the global min--max solution $x^*$.
The landscape of $F$ cannot be approximated well around such a discontinuous point if we only have a single candidate $\tilde{y}$ of the corresponding worst-case scenario.
We expect from \Cref{fig:berthing-trajectory-adv} that the robust berthing control problem discussed in \Cref{sec:intro} has the above difficulty.
The landscape of $F$ cannot be approximated well with a single candidate $\tilde{y}$ on $f_{10}$ as well because of the concavity of $f_{10}$ in $x$.
The nonsmoothness of $f_6$ and $f_8$ in $y$ can also cause a difficulty in approximating $F(x)$ in a neighborhood of $x^*$ by $f(x, \tilde{y})$ with a single candidate $\tilde{y}$. 
We expect that the landscape of $F$ is easier to approximate for smooth convex--concave functions such as $f_5$, $f_7$, and $f_{11}$. 
However, if the worst-case scenario $\yw: x \mapsto \yw(x)$ is continuous yet very sensitive, then approximating the landscape of $F$ with a single candidate $\tilde{y}$ will be unreasonable.
Such sensitivity is controlled by $B$ in the test problem definition.
The greater the greatest singular value of $B$ is, the more sensitive the worst-case scenario is.
In these situations, approximating the landscape of $F(x)$ locally around some point $\bar{x}$ by $f(x, \tilde{y})$ with a single candidate $\tilde{y} \approx \yw(\bar{x})$ is inadequate. 

\section{Limitations of existing approaches} \label{sec:limit}

As mentioned in \Cref{sec:intro}, \zopgda{} \cite{Liu2020} and \acma{} \cite{akimoto2022berthing} are promising approaches for the black-box min--max optimization. 
Both approaches are designed to converge to a strict local min--max saddle point $(\tilde{x}, \tilde{y})$. Let $(x^{t}, y^{t})$ be a pair of the solution candidate and the scenario candidate at iteration $t$. These approaches update it as 
\begin{align}
(x^{t+1}, y^{t+1}) = (x^t, y^t) + (\eta_x \cdot B_x, \eta_y \cdot B_y) , \label{eq:xyupdate}
\end{align}
where $\eta_x$ and $\eta_y$ denote the learning rates, and $B_x$ and $B_y$ denote the update vectors for $x$ and $y$, respectively. 
In \zopgda{}, $(B_x, B_y)$ comprises approximate gradients of the objective function, $(-\widehat{\nabla_{x} f}(x^t, y^t), \widehat{\nabla_{y} f}(x^t, y^t))$. The learning rates need to be tuned for each problem.
In \acma{}, $(B_x, B_y)$ comprises $(\bar{x}^t - x^t, \bar{y}^t - y^t)$, where $\bar{x}^t$ and $\bar{y}^t$ are approximations of $\argmin_{x \in \mathbb{X}}f(x, y^t)$ and $\argmax_{y \in \mathbb{Y}}f(x^t, y)$, respectively, obtained using (1+1)-CMA-ES 
\cite{Arnold2010, igel2006}
.
The learning rates are adapted during the optimization to alleviate tedious parameter tuning.

The above two existing approaches are theoretically guaranteed to converge to the global min--max saddle point \cite{akimoto2022berthing} or its neighborhood \cite{Liu2020} when the objective function is twice continuously differentiable and globally strongly convex--concave. 
Because the global min--max solution $x^*$ is the global min--max saddle point of $f$ in such problems, there is convergence to $x^*$ or its neighborhood.
In particular, the authors of \cite{akimoto2022berthing} showed sufficient conditions for linear convergence. 
Although the global convergence is not theoretically guaranteed, updating $x$ and $y$ alternately as in \eqref{eq:xyupdate} is expected to converge to a local min--max saddle point if the objective function is a locally smooth and strongly convex--concave around the local min--max saddle point.

In addition, the authors of \cite{akimoto2022berthing} reported several limitations of the above two existing approaches. 
Among them, the limitations for problems of Type (A) and (B) described in \Cref{sec:intro} are described below.  

\paragraph{Difficulty (I): slow convergence on smooth strongly convex--concave problems}

First, we discuss the slow convergence issue on smooth strongly convex--concave problems highlighted in \cite{akimoto2022berthing}. 
For instance, consider a convex--concave quadratic problem $f_{ex}(x,y)=(a/2)x^2+bxy-(c/2)y^2$. 
The worst-case scenario is $\yw(x)= (b/c) x$ for each $x$ and the optimal solution is $\hat{x}(y) = - (b/a) y$ for each $y$. 
It is intuitive that both $\yw(x)$ and $\hat{x}(y)$ should not be too sensitive to follow their change by \eqref{eq:xyupdate}. 
In fact, it has been theoretically derived that, for linear convergence, the learning rate must be set as $\eta_x, \eta_y \in O(ac / (ac + b^2))$ and the required number of iterations to find near-optimal solution is $\Omega(1 + b^2 / (ac))$; refer to \cite{akimoto2022berthing} for details. 
A similar limitation has been reported for the simultaneous gradient descent--ascent (SGDA) approach \cite{Liang2019}. 
The same limitation is expected to exist in \zopgda{} because it is regarded as an approximation of the SGDA approach. 
The adaptation of the learning rates in \acma{} can mitigate the difficulty in tuning learning rates.
However, it cannot avoid the slow convergence problem. 

The situation is worse if the objective function is convex--concave but not strongly convex--concave. 
For example, consider $f_7$ with ${d_x} = {d_y} = 1$ and $B = b$. 
This objective function is similar to $f_{ex}$, but the coefficients are regarded as $a=(1/2)x^2$ and $c=(1/2)y^2$, i.e., decreasing as the solution approaches the global min--max saddle point $(x^*, y^* = \yw(x^*))$.
In this problem, the learning rate must converge to zero as the solution approaches $(x^*, y^*)$.
This jeopardizes the advantage of the existing approaches, i.e., linear convergence to the min--max saddle point.
In fact, the authors of \cite{akimoto2022berthing} reported such an issue empirically.

\paragraph{Difficulty (II): nonconvergence to a min--max solution that is not a strict min--max saddle point}

Next, we discuss the nonconvergence issue on problems where $x^*$ is not a strict min--max saddle point. 
The existing approaches fail to converge to $x^*$. 
Such a situation occurs when the objective function is not strictly convex--concave. 
The situations can be categorized into two:
(W) $x^*$ is a weak min--max saddle point and (N) $x^*$ is not a min--max saddle point. 
Among the test problems in \Cref{tab:testp}, $f_1$ and $f_2$ fall into Category (W), and $f_4$, $f_9$, and $f_{10}$ fall into Category (N).
A numerical experiment in \cite{akimoto2022berthing} has shown that \acma{} fails to converge to $x^*$ on such problems. 
A theoretical investigation in \cite{Liang2019} has shown that SGDA fails to converge as well. 
Therefore, \zopgda{} is also expected to fail. 
The authors of \cite{Liang2019} reported that with some modifications, SGDA can converge to the weak global min--max saddle point on bilinear functions.
The existing approaches may tackle problems of Category (W) by incorporating such a modification. 
However, problems of Category (N) cannot be solved. 

In our experiments, we also confirmed that there exists a situation where the existing approaches fail to converge even if $x^*$ is a strict global min--max saddle point. Example functions are $f_3$, $f_6$,  and $f_8$, which are strictly convex--concave but nonsmooth. 
The situation where $x^*$ is a strict global min--max saddle point but $f$ is nonsmooth is denoted as Category (S).

\paragraph{Direction to address Difficulties (I) and (II)}

One approach to avoid Difficulty (II) is to approximate the worst-case objective function $F$ by solving $\max_{y \in \mathbb{Y}} f(x, y)$ numerically and optimize it directly.
If $F$ can be approximated well for each $x \in \mathbb{X}$, i.e., $\max_{y \in \mathbb{Y}} f(x, y)$ can be solved efficiently for each $x$, and $F$ can be globally optimized efficiently by a numerical solver, it does not matter whether $x^*$ is a min--max saddle point or not. Therefore, Difficulty (II) can be addressed naturally. 

We also expect that there can be a solution to Difficulty (I).
Because any smooth strongly convex--concave function can be approximated by a quadratic convex--concave function around the global min--max saddle point, we focus on $f_{ex}$ for simplicity.
Its worst-case objective function is $F(x) = \frac12 (a + b^2/c) x^2$. 
Because it is a convex quadratic function, a reasonable solver converges linearly to its global minimum point $x^*$.
For ${d_y} > 1$ and ${d_x} > 1$, the worst-case objective function can be ill-conditioned. 
However, if we employ a solver that uses second-order information, such as CMA-ES  \cite{Hansen2001, Hansen2014, akimoto2019}, we expect that it can be solved efficiently. Therefore, the number of $f$-calls spent by the approach that directly optimizes $F$ is expected to be less sensitive to the interaction term.
If the objective function is smooth and weakly convex--concave, this argument does not hold. 
However, considering the aforementioned example $f_{7}$, we have $F(x) = (1/4) x^4 + (3/4)(b x)^{4/3}$, which is smooth at $x^*=0$ and strictly convex. 
Therefore, we expect that a comparison-based approach, invariant to any increasing transformation of the objective function, can solve it efficiently.

\section{Proposed approach}\label{sec:proposed}

We propose a novel approach to address Difficulties (I) and (II). 
The main idea is to directly minimize the worst-case objective function $F$.
The bottleneck of directly minimizing $F$ in the black-box min--max optimization setting is the computational time for each $F(x)$ evaluation, which 
requires solving 
maximization problem $\max_{y \in \mathbb{Y}} f(x, y)$ approximately. 
To tackle this bottleneck, we propose to employ CMA-ES to minimize $F$ (\Cref{sec:cmaes4F}), and propose the WRA mechanism that approximates the ranking of $\{F(x_i)\}_{i=1}^{\lambda_x}$ for the given solution candidates $\{x_i\}_{i=1}^{\lambda_x}$ (\Cref{sec:wra}).

For the proposed approach to work effectively, we suppose (a) $\abs{\ywset(x)} = 1$ and $\yw(x)$ is continuous almost everywhere in $\mathbb{X}$, (b) the solver for the inner maximization problem can globally maximize $f(x, y)$ with respect to $y$ efficiently for each $x \in \mathbb{X}$, and (c) the solver for the outer minimization problem, CMA-ES in this study, can minimize $F$ efficiently. 
Unfortunately, one can not confirm these assumptions in advance as our target problems are black-box. 
However, (a) is very natural to assume if the objective function is continuous almost everywhere, and (b) and (c) are more like our hope to justify our choice of the baseline optimizer.

\begin{wrapfigure}[33]{R}{0.55\hsize}
\begin{minipage}{\hsize}
\begin{algorithm}[H]
	\caption{WRA}\label{alg:wra}
    \begin{algorithmic}[1]
    \Require $x_1, \dots, x_{\lambda_x}$ 
    \Require $\{({y}_k, \omega_k, p_k)\}_{k=1}^{\ny}$
    \Require $\tau_{\mathrm{threshold}}$, $p_{\mathrm{threshold}}$, $\bar{p}_{+}$, $\bar{p}_{-}$
    
    \State\label{line:wra-initstart} \texttt{// 
    Warm-starting
    }
    \For{$i = 1, \dots, \lambda_x$}
    \State evaluate $f(x_{i}, y_k)$ for all $k = 1,\dots,\ny$
    \State $k^\mathrm{worst}_{i} = \argmax_{k \in \{ 1, \dots,\ny\}} f(x_i, y_k)$
    \State $\hat{y}_i=y_{k^\mathrm{worst}_{i}}$, $\tilde{\omega}_i = \omega_{k^\mathrm{worst}_{i}}$, and $F^0_i = f(x_i, y_{k^\mathrm{worst}_{i}})$
    \EndFor\label{line:wra-initend}
    
    \State\label{line:wra-worststart} \texttt{// Early-stopping}
    \State initialize $\tilde{\theta}_1, \dots, \tilde{\theta}_{\lambda_x}$
    \For{$\mathrm{rd} = 1,2,\dots$}
    \For{$i = 1, \dots, \lambda_x$}
    \State $F_{i}^{\mathrm{rd}}, \hat{y}_i, \tilde{\omega}_i, \tilde{\theta}_i \gets \mathcal{M}(F_{i}^{\mathrm{rd}-1}, \hat{y}_i, \tilde{\omega}_i, \tilde{\theta}_i)$ 
    \EndFor
    \State $\tau = \text{Kendall}(\{F^{\mathrm{rd}-1}_{i}\}_{i=1}^{\lambda_x}, \{F^{\mathrm{rd}}_{i}\}_{i=1}^{\lambda_x})$
    \State \textbf{break if} $\tau > \tau_{\mathrm{threshold}}$
    \EndFor\label{line:wra-worstend}
    
    \State \texttt{// Postprocessing}\label{line:wra-poststart}
    \State $S^\mathrm{worst} = \{{k^\mathrm{worst}_{i}} \text{ for } i = 1,\dots,\lambda_x\}$ 
    \For{$\tilde{k} \in S^\mathrm{worst}$}
    \State $\ell = \argmin_{i=1,\dots,\ny} \{F^{\mathrm{rd}}_i \mid k^\mathrm{worst}_i = \tilde{k}\}$ 
    \State $y_{\tilde{k}} = \hat{y}_\ell$, $\omega_{\tilde{k}} = \tilde{\omega}_\ell$  
    \State $p_{\tilde{k}} = \min(p_{\tilde{k}} + \bar{p}_{+}, 1)$
    \EndFor
    \State $p_k = p_k -  \bar{p}_{-} \cdot \ind{k \notin S^\mathrm{worst}}$ for all $k=1,...,\ny$
    \For{$k = 1,\dots,\ny$}
    \State refresh $(y_k, \omega_k, p_k)$ \textbf{if} $p_k < p_{\mathrm{threshold}}$
    \EndFor
    \label{line:wra-postend}
    \State \Return $\{F_i^{\mathrm{rd}}\}_{i=1}^{\lambda_x}$ and $\{({y}_k, \omega_k, p_k)\}_{k=1}^{\ny}$ for the next call
  \end{algorithmic}
\end{algorithm}
\end{minipage}
\end{wrapfigure}

\subsection{CMA-ES for outer minimization} \label{sec:cmaes4F}

The proposed approach tries to solve the outer minimization problem of \eqref{eq:minmax} using the CMA-ES.
The CMA-ES is a state-of-the-art derivative-free optimization approach for continuous black-box optimization problems \cite{hansen2009, Hansen2010, rios2013derivative} and has been used in several real-world applications \cite{miyagi@ghgt, maki2020, urieli2011, fujii2018, tanabe2021}. 
There are two essential characteristics of the CMA-ES that attract attention.
One is that it is a quasiparameter-free approach, i.e., one does not need any hyperparameter tuning except for a population size $\lambda_x$, which is desired to be increased if the problem is multimodal or noisy or if several CPU cores are available.
Because the worst-case objective function $F$ is a black-box and it is difficult to understand the characteristics of $F$ in advance, the parameter-free nature is essential.
The second is that it is parallel-implementation friendly. 
The objective function values ($F$ in our case) of multiple solution candidates generated at an iteration can be evaluated in parallel. 
It is desired when the computational cost of the objective function evaluation is high. 
Because each evaluation of $F$ is expensive as it requires solving maximization problem $\max_{y \in \mathbb{Y}} f(x, y)$ approximately, this is practically essential.

The CMA-ES repeats the sampling, evaluation, and update steps until a termination condition is satisfied. 
Let $t \geq 0$ be the iteration counter.
First, $\lambda_x$ solution candidates $\{x_i\}_{i=1}^{\lambda_x}$ are generated independently from a Gaussian distribution $\mathcal{N}(m_x^t, \Sigma_x^t)$ with mean vector $m_x^t \in \mathbb{X}$ and covariance matrix $\Sigma_x^t \in \R^{d_x \times d_x}$.
Next, the worst-case objective function values of the $\lambda_x$ solution candidates, $\{F(x_{i})\}_{i=1}^{\lambda_x}$, are evaluated, and 
their 
rankings $\mathrm{Rank}_{F}(\{x_{i}\}_{i=1}^{\lambda_x})$ are computed, where the $i$th ranked solution candidate has the $i$th smallest $F$ value. 
Finally, the CMA-ES updates the distribution parameters, $m^t_x$ and $\Sigma^t_x$, and other dynamic parameters using the solution candidates and their rankings. 
An important aspect of the update of the CMA-ES is that it is comparison-based. 
That is, provided the rankings of the solution candidates, $\mathrm{Rank}_{F}(\{x_{i}\}_{i=1}^{\lambda_x})$, are computed, the worst-case objective function values, $\{F(x_i)\}_{i=1}^{\lambda_x}$, do not need to be accurately computed.

In this study, we implemented the version of the CMA-ES proposed in \cite{akimoto2019}, namely, dd-CMA-ES, as the default solver.\footnote{
The code for DD-CMA-ES is downloaded from \url{https://gist.github.com/youheiakimoto/1180b67b5a0b1265c204cba991fa8518}
.}%
The configuration of the CMA-ES follows the default proposed procedure in \cite{akimoto2019}.
If the search domain has a box constraint, we employ the mirroring technique along with upper-bounding the coordinate-wise standard deviation $\sqrt{[\Sigma_x^t]_{\ell,\ell}}$ for $\ell = 1, \dots, d_x$ \cite{yamaguchi2018}, where $[\Sigma_x^t]_{\ell,\ell}$ denotes the $(\ell, \ell)$-th element of $\Sigma_x^t$.
The initial distribution parameters, $m_x^0$ and $\Sigma_x^0$, should be set problem-dependently. 
We terminate the CMA-ES when $\max_{\ell \in \{1,\dots,d_x\}} \sqrt{[\Sigma_x^t]_{\ell,\ell}} < V_{\min}^{x}$ is satisfied, where $V_{\min}^{x}$ is a problem-dependent threshold, or $\Cond(\Sigma_x) > \Cond_{\max}^{x}=10^{14}$ is satisfied, where $\Cond(\Sigma_x)$ is the condition number, i.e., the ratio of the greatest to smallest eigenvalues, of $\Sigma_x^t$.

\subsection{
Worst-case Ranking Approximation
}\label{sec:wra}

The proposed WRA mechanism approximates the rankings of solution candidates by roughly solving maximization problems $\max_{y \in \mathbb{Y}} f(x_i, y)$ for each solution candidate $\{x_i\}_{i=1}^{\lambda_x}$. 
To save the inner $f$-calls to approximate the rankings $\mathrm{Rank}_{F}(\{x_{i}\}_{i=1}^{\lambda_x})$, we incorporate a 
warm-starting 
strategy, where we try to start each maximization $\max_{y \in \mathbb{Y}} f(x_i, y)$ with a good initial solution candidate and a good configuration of the inner solver (\Cref{sec:warmstart}), and an early-stopping strategy, where we try to stop each maximization $\max_{y \in \mathbb{Y}} f(x_i, y)$ once $\mathrm{Rank}_{F}(\{x_{i}\}_{i=1}^{\lambda_x})$ are considered well-approximated (\Cref{sec:earlystop}).
The overall framework is summarized in \Cref{alg:wra}.

Hereinafter, let $\mathcal{M}$ be a solver used to approximately solve $\max_{y \in \mathbb{Y}} f(x_i, y)$. 
Let $\omega$ represent the configurations of the solver $\mathcal{M}$ inherited over the WRA calls.
Let $\theta$ represent the other configurations that are not inherited.

\subsubsection{
Warm-starting
strategy}\label{sec:warmstart}

Two key ideas behind the design of our 
warm-starting
strategy are as follows. 

First, we inherit the worst-case scenario candidates and the configurations from the last WRA call. 
The Gaussian distribution $\mathcal{N}(m^t, \Sigma^t)$ of the CMA-ES for the outer minimization does not significantly change in one iteration. 
Then, the distribution of the worst-case scenarios for the solution candidates generated at iteration $t$ is considered to be similar to that at iteration $t+1$.
Therefore, we expect that using the solver configurations used at the last iteration will contribute to reduce the number of $f$-calls. 
This idea is expected to be effective for the problem where $\abs{\ywset(x)} = 1$ and $\yw(x)$ is continuous almost everywhere 
in~$\mathbb{X}$.  

Second, we maintain $\ny$ ($\geq 1$) configurations.
Consider situations (W) and (N) described in \Cref{sec:testp}.
The worst-case scenarios corresponding to solution candidates $\{x_i\}_{i=1}^{\lambda_x}$ generated in a single iteration may not be concentrated at one point but may be distributed around $\abs{\ywset(x^*)}$ distinct points even if $\{x_i\}_{i=1}^{\lambda_x}$ are concentrated around $x^* = \argmin_{x \in \mathbb{X}} F(x)$. 
If we only maintain one configuration, it may be a good initial configuration only for a small portion of $\{x_i\}_{i=1}^{\lambda_x}$.
There is a high risk that $F$ values are accurately estimated only for these candidates and they are underestimated for the others due to insufficient maximization.
To address this difficulty, we maintain multiple configurations and try to keep them diverse. 

These two ideas are implemented in our 
warm-starting
strategy.
It comprises (1) selecting a good initial worst-case scenario candidate $\tilde{y}$ and configuration $\tilde{\omega}$ of solver $\mathcal{M}$ for each solution candidate $x_i$ among $\ny$ pairs $\{(y_k, \omega_k)\}_{k=1}^{\ny}$ (Lines~\ref{line:wra-initstart}--\ref{line:wra-initend} in \Cref{alg:wra}) and (2) preparing $\ny$ pairs $\{(y_k, \omega_k)\}_{k=1}^{\ny}$ for the next WRA call (Lines~\ref{line:wra-poststart}--\ref{line:wra-postend} in \Cref{alg:wra}).
For each $x_i$, we evaluate $f(x_i, y_k)$ for $k = 1, \dots, \ny$ and select the worst-case scenario candidate. 
Let $k_i^{\mathrm{worst}}=\argmax_{k\in \{1, \dots, \ny\}} f(x_i, y_k)$ be the index of the worst-case scenario candidate among $\{y_k\}_{k=1}^{\ny}$. 
Then, we select the configuration $\omega_{k_i^{\mathrm{worst}}}$ of the solver that generated $y_{k_i^{\mathrm{worst}}}$ as the initial configuration $\tilde{\omega}_i$ to search for the worst-case scenario for $x_i$. 
After approximating $\{F(x_i)\}_{i=1}^{\lambda_x}$, we update the set of configurations of the solver. 
Basically, we replace the selected configurations with the configurations obtained after the solver execution. 
If the same configuration is selected for different solution candidates, we replace the configuration with the one used for the solution candidate with the optimal approximated worst-case value. 

Moreover, to avoid keeping unused configurations, we refresh such configurations and try to have diverse configurations. For this purpose, we maintain a parameter $p_k \in (0,1]$ for $k = 1,\dots,\ny$ and initialize the parameter as $1$. 
The parameter $p_k$ is increased by $\bar{p}_{+}$ if the $k$th configuration is selected. It is decreased by $\bar{p}_{-}$ otherwise. 
Once we have $p_k \leq p_{\mathrm{threshold}}$, the $k$th configuration and the corresponding worst-case scenario candidate are refreshed in the same manner as their initialization, and $p_k$ is reset to $1$.

\subsubsection{Early-stopping strategy}\label{sec:earlystop}

Our early-stopping strategy is to save $f$-calls by terminating $\lambda_x$ solvers once the rankings of the worst-case objective function values of the given solution candidates, $\mathrm{Rank}_{F}(\{x_{i}\}_{i=1}^{\lambda_x})$, are regarded as well-approximated. The early-stopping strategy is described at Lines \ref{line:wra-worststart}--\ref{line:wra-worstend} in \Cref{alg:wra}.

The main idea is as follows.
As aforementioned, the CMA-ES is a comparison-based approach. 
Therefore, the worst-case objective function values are not needed to be accurately estimated provided their rankings are computed. 
We further hypothesize that the CMA-ES behaves similarly on the approximated rankings if the rankings of solution candidates are approximated with a high correlation to the true rankings, according to Kendall \cite{kendall}. This hypothesis is often imposed in surrogate-assisted approaches and related approaches \cite{lqcmaes,multifidelity,constrainedmultifidelity,miyagi2021,evolutioncontrol, miyagi2023} and is partly validated in theory \cite{surrogatetheory}.
Because the true rankings of the worst-case objective function values are unknown, instead of trying to check the rank correlation between the true and approximate rankings, we keep track of changes in the rankings and stop if the change is regarded as sufficiently small.

To compute the rankings of the worst-case objective function values, $\lambda_x$ solvers are run in parallel, and we periodically compute the rankings of the solution candidates using the approximated worst-case objective function values, $\{F_i^{\mathrm{rd}} = f(x_i, \hat{y}_i)\}_{i=1}^{\lambda_x}$, where $\mathrm{rd} \geq 0$ is the number of ranking computations so far and is called the round. 
After each round, we compute the Kendall's rank correlation between the current and last approximations of the rankings, $\tau(\{F_i^{\mathrm{rd}-1}\}_{i=1}^{\lambda_x}, \{F_i^{\mathrm{rd}}\}_{i=1}^{\lambda_x})$. 
If it is greater than the predefined threshold $\tau_\mathrm{threshold} \geq 0$, we regard the rankings are well-approximated and terminate the solvers. A reasonable definition of a round of a solver call depends on the choice of the solver. We discuss the solver choice and round definition in the next section.

\subsubsection{Hyperparameters} \label{sec:wraparameters}

The hyperparameters for WRA are the threshold for Kendall's rank correlation $\tau_{\mathrm{threshold}}$, number of configurations $\ny$, threshold $p_{\mathrm{threshold}}$, and parameters $\bar{p}_{+}$ and $\bar{p}_{-}$ for the refresh strategy. 
The initial configurations $\{\omega_k\}_{k=1}^{\ny}$ and initial worst-case scenario candidates $\{y_k\}_{k=1}^{\ny}$ must be set problem-dependently. 
We describe the expected effect of these hyperparameters in this section. 
The sensitivities of $\tau_{\mathrm{threshold}}$, $\bar{p}_{+}$ and $\bar{p}_{-}$ are empirically investigated in \Cref{app:wrasensitivity}.

Threshold $\tau_{\mathrm{threshold}}$ should be set to a relatively high value to approximate $\mathrm{Rank}_{F}(\{x_{i}\}_{i=1}^{\lambda_x})$ with high accuracy.
However, setting a high value of $\tau_{\mathrm{threshold}}$ (e.g., $\tau_{\mathrm{threshold}} = 1$) has a risk of spending too many $f$-calls. 
Based on our sensitivity analysis in \Cref{app:wrasensitivity}
, 
we set its default value as $\tau_{\mathrm{threshold}} = 0.7$ and used this value throughout our experiments. 

The number of configurations, $\ny$, is desired to be set no smaller than the number $\abs{\ywset(x^*)}$ of worst-case scenarios around $x^*$ to maintain good configurations and good initial scenarios for each solution. 
In addition, because $\ny$ $f$-calls are required to select the initial configuration for each $x$, $\ny$ is desired to be as small as possible. 
However, $\abs{\ywset(x^*)}$ is unknown in advance and is problem-dependent.
We suggest setting $\ny$ to be a few times greater than $\lambda_x$ to allow $\lambda_x$ solution candidates a chance to use $\lambda_x$ distinct worst-case scenario candidates. The effect is further discussed in \Cref{sec:test}.

The parameters $p_{\mathrm{threshold}}$, $\bar{p}_{+}$, and $\bar{p}_{-}$ affect the frequency of each configuration to be refreshed.
If the configurations are frequently refreshed, our 
warm-starting
strategy may be less effective. 
In our sensitivity analysis described in \Cref{app:wrasensitivity},
we confirmed that the performance of the proposed approach was not very sensitive to the change of the frequency of refreshing configurations on the test problems.
Therefore, we set $p_{\mathrm{threshold}} = 0.1$, $\bar{p}_{+} = 0.4$ and $\bar{p}_{-} = 0.05$ as the default values and these values were used in all experiments in this paper. 
In this case, the configurations $\{\tilde{\omega_i}\}_{i=1}^{\lambda_x}$ are kept for at least $6 = (\bar{p}_{+} - p_{\mathrm{threshold}}) / \bar{p}_{-}$ outer loop iterations after the last use or $18 = (1 - p_{\mathrm{threshold}}) / \bar{p}_{-}$ iterations after the initialization or last refresh.

\subsection{Implementation of WRA}

We implement two variants of WRA with the CMA-ES (\Cref{sec:wracma}) and AGA (\Cref{sec:wraslsqp}) as solvers $\mathcal{M}$.

\subsubsection{WRA using CMA-ES} \label{sec:wracma}

The first variant, summarized in \Cref{alg:CMAESinwra}, uses dd-CMA-ES \cite{akimoto2019} as a solver $\mathcal{M}$. 
If the search domain has a box constraint, we employ the mirroring technique along with upper-bounding the coordinate-wise standard deviation \cite{yamaguchi2018}.
The configuration $\tilde{\omega}$ includes the mean vector $\tilde{m}$ and covariance matrix $\tilde{\Sigma}$, and $\tilde{\theta}$ includes other parameters such as evolution paths, iteration  counter $t' \geq 0$ (initialized as $t' = 0$), and termination flag $h$ (initialized as $h = \textsc{False}$).

Because the proposed approach is a double-loop approach, setting the termination conditions for the inner loop is crucial. 
\Cref{alg:CMAESinwra} runs the CMA-ES until the worst-case scenario candidate is improved $c_{\max}$ times. 
If the worst-case scenario candidate is improved for $c_{\max}$ times, we regard that it is significantly improved.
Similar to the CMA-ES for outer minimization, we terminate the maximization process if all coordinate-wise standard deviations, $\sqrt{[\tilde{\Sigma}]_{\ell,\ell}}$, become smaller than $V_{\min}^{y}$.
In this situation, we expect that the distribution is sufficiently concentrated and no more significant improvement will be obtained.
We stop the CMA-ES if the condition number, $\Cond(\tilde{\Sigma})$, becomes greater than $\Cond_{\max}^{y}$.
If one of the latter two conditions is satisfied, we set $h = \textsc{True}$, and the CMA-ES will not be executed in the following rounds in the current WRA call.

The distribution parameters are inherited over WRA calls. 
Once the condition $\max_{\ell} \sqrt{[\tilde{\Sigma}]_{\ell,\ell}} < V_{\min}^{y}$ is satisfied for some configurations, it is expected to be immediately satisfied in the next WRA call if these configurations are selected. 
However, because the objective function with respect to $y$, i.e., $f(x_i, y)$, differs as solution candidates $x_i$ differ in each WRA call, there is a chance that the distribution will be enlarged due to the step-size adaptation mechanism of the CMA-ES, and a significant improvement will be realized. 
Therefore, we force all coordinate-wise standard deviations to be no smaller than $V_{\min}^y$ once the greatest one becomes smaller than $V_{\min}^y$ (Lines \ref{l:sig_correct_1}--\ref{l:sig_correct_2}) and the CMA-ES to run at least $T_{\min}$ iterations for each WRA call.

The hyperparameters includes the initial configurations for inner CMA-ES $(\{m_k\}_{k=1}^{\ny}, \{\Sigma_k\}_{k=1}^{\ny}, \theta)$, initial scenarios $\{y_k\}_{k=1}^{\ny}$, and termination conditions for \Cref{alg:CMAESinwra}, $c_{\max}$, $V_{\min}^{y}$, and $T_{\min}$. 
The configuration and initialization of $\theta$, including the initialization of evolution paths and population size $\lambda_y$, follow the values proposed in \cite{akimoto2019}. 
The parameter $c_{\max}$ impacts the approximation accuracy of the rankings on the worst-case objective function values $\mathrm{Rank}_{F}(\{x_i\}_{i=1}^{\lambda_x})$ and $f$-calls to approximate the rankings.  
If $c_{\max}$ is set to a greater value, WRA will require more $f$-calls. Meanwhile, setting $c_{\max}$ to a smaller value has a risk to terminate the scenario improvement before the ranking on the worst-case objective function $\mathrm{Rank}_{F}(\{x_{i}\}_{i=1}^{\lambda_x})$ is estimated with sufficient accuracy. 
The parameter $T_{\min}$ can be set to a constant value, as the CMA-ES can increase the standard deviation rapidly if it is desired. 
We set $T_{\min} = 10$ as the default value.
The parameter $V_{\min}^{y}$ and initial distributions $\{(m_k, \Sigma_k)\}_{k=1}^{\ny}$ must be set problem-dependently. 
The initial scenarios $\{y_k\}_{k=1}^{\ny}$ are drawn from the initial distributions, i.e., $y_k \sim \mathcal{N}(m_k, \Sigma_k)$.

\noindent\begin{minipage}{0.54\hsize}%
\begin{algorithm}[H]
	\caption{CMA-ES as $\mathcal{M}$}\label{alg:CMAESinwra}
    \begin{algorithmic}[1]
    \Require $x, \hat{y}, F_y, \tilde{\omega} = (\tilde{m}, \tilde{\Sigma}), \tilde{\theta} = (h, t', \dots)$ 
    \Require $V_{\min}>0$, $c_{\max} \geq 1$, $\lambda_y = \lfloor 4 + 3\log(d_y) \rfloor$
    
    \State $\tilde{\Sigma}_{\mathrm{init}}=\tilde{\Sigma}$, $c=0$
    \While{$c < c_{\max}$ \textbf{and} $h = \textsc{False}$}
    
    \State Sample $ \{\hat{y}'\}_{k=1}^{\lambda_y} \sim \mathcal{N}(\tilde{m}, \tilde{\Sigma})$
    \State Evaluate $f_k = f(x, \hat{y}_{k}^{'})$ for all $k=1,\dots,\lambda_y$
    
    \State Select the worst index $\tilde{k}^\mathrm{worst}=\argmax_{k=1,\dots,\lambda_y} f_k$
    \If {$f(x, \hat{y}_{\tilde{k}^\mathrm{worst}}^{'}) > F_{y}$}
    \State $F_{y} = \max_{k=1,\dots,\lambda_y} f_k$, $\hat{y} = \hat{y}'_{\tilde{k}^\mathrm{worst}}$, and $c = c + 1$
    \EndIf
    
    \State Perform CMA-ES update using $\{\hat{y}_{k}^{'}, f_k\}_{k=1}^{\lambda_y}$
    \If {$\max_{\ell}\Big\{\sqrt{[\tilde{\Sigma}]_{\ell,\ell}}\Big\} < V_{\min}^{y}$ \text{and} $t' \geq T_{\min}$} 
    \State\label{l:sig_correct_1} $D = \diag\bigg(\max\bigg(1, \frac{V_{\min}^y}{\sqrt{[\tilde{\Sigma}]_{1,1}}}\bigg), \dots, \max\bigg(1, \frac{V_{\min}^y}{\sqrt{[\tilde{\Sigma}]_{d_y,d_y}}}\bigg)\bigg)$
    \State\label{l:sig_correct_2} $\tilde{\Sigma} = D \tilde{\Sigma} D$ and $h = \textsc{True}$
    \EndIf
    \State $h = \textsc{True}$ and set $\tilde{\Sigma}=\tilde{\Sigma}_{\mathrm{init}}$ \textbf{ if } $\Cond(\tilde{\Sigma}) > \Cond_{\max}^{y}$
    \State $t'=t'+1$
    \EndWhile
    \State \Return $\hat{y}$, $F_y$, $\tilde{\omega} = (\tilde{m}, \tilde{\Sigma})$, $\tilde{\theta} = (h, t', \dots)$
  \end{algorithmic}
\end{algorithm}
\end{minipage}
\hfill
\begin{minipage}{0.42\hsize}%
\begin{algorithm}[H]
	\caption{AGA as $\mathcal{M}$}\label{alg:SLSQPinwra}
    \begin{algorithmic}[1]
    \Require $x, \hat{y}, F_y, \tilde{\omega} = \tilde{\eta}, \tilde{\theta} = (h, \dots)$
    \Require $U_{\min}>0$, $c_{\max} \geq 1$, $\beta \in (0,1)$
    \State $c=0$
    \While{$c < c_{\max}$ \textbf{and} $h = \textsc{False}$}
    \State Obtain approximated gradient $\bar{\nabla}_y f$ at $\hat{y}$    
    \State $\hat{y}' = \hat{y} + \tilde{\eta} \bar{\nabla}_y f $
    \If {$f(x,\hat{y}')>F_{y}$}
    \State $\tilde{\eta} = \tilde{\eta} / \beta$
    \Else
    \While{$f(x,\hat{y}') \leq F_y $}
    \State $\tilde{\eta} = \tilde{\eta} \times \beta$
    \State $\hat{y}' = \hat{y} + \tilde{\eta} \bar{\nabla}_y f$
    \State $h = \textsc{True}$ \textbf{ if } $\norm{\tilde{\eta} \bar{\nabla}_y f}_{\infty} \leq U_{\min}$
    \EndWhile
    \EndIf

    \If {$f(x, \hat{y}') > F_y $}
    \State $F_y = f(x, \hat{y}')$, $\hat{y} = \hat{y}'$, and $c = c + 1$
    \EndIf
    \EndWhile
    
    \State \Return $\hat{y}$, $F_y$, $\tilde{\omega} = \tilde{\eta}$, $\tilde{\theta} = (h, \dots)$
  \end{algorithmic}
\end{algorithm}
\end{minipage}

\subsubsection{WRA using AGA} \label{sec:wraslsqp}

The second variant, summarized in \Cref{alg:SLSQPinwra}, uses AGA as a solver $\mathcal{M}$. The AGA solver uses the numerical gradient $\bar{\nabla}_y$ at the worst-case scenario candidate $\hat{y}$ obtained by \slsqp{} function in the SciPy module in Python.\footnote{
Note that \slsqp{} is not used for maximizing the objective function value but obtaining the numerical gradient.
}
If the search domain for the scenario vector has a box constraint, the projected gradient idea is used to force the worst-case scenario candidate to be feasible.
The configuration $\omega$ for \Cref{alg:SLSQPinwra} includes the learning rate $\{\eta_k\}_{k=1}^{\ny}$, and the other parameters are included in $\theta$. 

We use a simple adaptation mechanism for the learning rate $\tilde{\eta}$ in \Cref{alg:SLSQPinwra}, similar to the backtracking line search.
The learning rate $\tilde{\eta}$ is decreased by $\beta \in (0,1)$ until the worst-case scenario candidate is improved.
If the worst-case scenario candidate is improved for the first trial, the learning rate is increased by $1/\beta$.
This is because a significant improvement of the worst-case scenario candidate is expected by a large learning rate in the next iteration.

The termination criteria of \Cref{alg:SLSQPinwra} are described as follows.
\Cref{alg:SLSQPinwra} is terminated when the scenario is improved for $c_{\max}$ times.
If the infinity norm of the update vector is smaller than $U_{\min}$, i.e.,
$\norm{\tilde{\eta} \bar{\nabla}_y f}_{\infty} \leq U_{\min}$, 
we consider that a significant increase of the objective function value is not expected and terminate the solver.
When \Cref{alg:SLSQPinwra} is terminated by the latter condition, we set $h=\textsc{True}$ and $\mathcal{M}$ is not called with the current configuration in the current WRA call.

The hyperparameters include the initial learning rate $\{\eta_k\}_{k=1}^{\ny}$, parameter for updating the learning rate $\beta$, termination threshold $U_{\min}$, and maximum number of improvements, $c_{\max}$. 
They should be set problem-dependently.

\subsection{Restart and Local Search Strategy} \label{sec:restart}

We implement two devices for practical use to enhance exploration (by restart) and exploitation (by local search).

A restart strategy is implemented to obtain good local optimal solutions when $F$ is multimodal. 
When a termination condition is satisfied before an $f$-call budget or a wall clock time budget is exhausted, the $\lambda_x$ solution candidates and $\ny$ worst-case scenario candidates at the last iteration are stored in $\mathcal{X}^*$ and $\mathcal{Y}^*$, respectively. 
We restart the search without inheriting any information from previous restarts.
Once the budgets are exhausted, the last solution candidates and worst-case scenario candidates are stored as well.
The final output of the algorithm, i.e., the candidate of the global min--max solution, is $\argmin_{x \in \mathcal{X}^*} \max_{y \in \mathcal{Y}^*} f(x,y)$.
One can also include randomly sampled scenario vectors to $\mathcal{Y}^*$ when deciding the final output for a good estimate of $F(x)$.
The resulting algorithms using CMA-ES and AGA with this restart strategy are denoted as \wracma{} and \wraslsqp{}, respectively.

We implement an optional local search strategy using \acma{}.
If the problem is locally smooth and strongly convex--concave, \acma{} exhibits significantly faster convergence than WRA. 
Therefore, by stopping each run of WRA early and performing \acma{}, we expect that the solution candidate obtained by WRA will be more locally improved by \acma{} than by spending the same $f$-calls by WRA.
This is implemented as follows.
When a termination condition is satisfied, let $\mathcal{Y} = \{y_k\}_{k=1}^{\ny}$ be the set of $\ny$ worst-case scenario candidates, $i_\mathrm{Adv} = \argmin_{i=1, \dots,\lambda_x} \max_{y \in \mathcal{Y}} f(x_i, y)$ be the best-case solution candidate, and $k_{\mathrm{Adv}} =  \argmax_{k=1,\dots,\ny} f(x_{i_\mathrm{Adv}}, y_k)$ be the corresponding worst-case scenario index obtained at the last iteration.
Then, \acma{} is applied to optimize $f_{\mathcal{Y}}(x, y) = \max_{\tilde{y} \in \{y\} \cup \mathcal{Y}}f(x, \tilde{y})$, with distributions initialized around $(x_{i_\mathrm{Adv}}, y_{k_{\mathrm{Adv}}})$ to exhibit local search.
The search distribution for $x$ in \acma{} is initialized by the distribution for $x$ in WRA at the last iteration. 
The distribution parameters for search in $y$ is initialized by those of $\omega_{k_\mathrm{Adv}}$ if \Cref{alg:CMAESinwra} is used.
When \Cref{alg:SLSQPinwra} is used, the mean vector is initialized by $y_{k_\mathrm{Adv}}$, and 
a relatively small initial covariance matrix, $10^{-2} \times \left(\frac{b_y}{2}\right)^2 I_{d_y}$, is used as \acma{} is used for local search.
The other parameters of \acma{} are set to the default values proposed in \cite{akimoto2022berthing}. 
Once \acma{} is terminated, we perform a restart as in \wracma{} and \wraslsqp{}. 
The approaches using \Cref{alg:CMAESinwra} and \Cref{alg:SLSQPinwra} with the local search and restart strategies are denoted as \wracmaadv{} and \wraslsqpadv{}, respectively.

\section{Numerical experiments on test problems} \label{sec:test}

We performed numerical experiments to confirm that existing approaches \acma{} and \zopgda{} face Difficulties (I) and (II), whereas the proposed approach can cope with them.
\footnote{
The code for \acma{} is downloaded from \url{https://gist.github.com/youheiakimoto/ab51e88c73baf68effd95b750100aad0}.
The code for \zopgda{} is downloaded from \url{https://github.com/KaidiXu/ZO-minmax}.
}%

\subsection{Common settings} \label{sec:common}

We used test problems listed in \Cref{tab:testp}.
Unless otherwise specified, the dimensions were ${d_x} = {d_y} = 20$, and the search domains were $\mathbb{X}=[-3, 3]^{d_x}$ and $\mathbb{Y}=[-3, 3]^{d_y}$.
The coefficient matrix $B$ was set to $B = \diag(b,.... ,b)$.

The proposed and existing approaches were configured as follows. 
The initial mean vector (\wracma{} and \wraslsqp{}) and initial solution candidate (\acma{} and \zopgda{}) for outer minimization were drawn from $\mathcal{U}(\mathbb{X})$.
The initial covariance matrices for the outer minimization were set to $\left(\frac{u_x - \ell_x}{4}\right)^2 I_{d_x}$ in \wracma, \wraslsqp, and \acma{}. 
The initial mean vectors (\wracma{}) and initial worst-case scenario candidates (\wraslsqp{}, \acma{}, and \zopgda{}) were drawn independently from $\mathcal{U}(\mathbb{Y})$. 
The initial covariance matrices for the inner maximization were set to $\left(\frac{b_y}{2}\right)^2 I_{d_y}$ in \wracma{} and \acma{}.
For \wracma{} and \wraslsqp{}, we set $V_{\min}^x=10^{-12}$, $c_{\max}=1$, and $\ny=36 (=3 \times \lambda_x)$.
For \wracma{}, we set $V_{\min}^y=10^{-4}$ and $T_{\min} =10$. For \wraslsqp{}, we set $U_{\min}=10^{-5}$, $\beta=0.5$, and the initial learning rate $\{\eta_k\}_{k=1}^{\ny} = 1$.  
For \zopgda{}, referencing \cite{Liu2020}, we set the learning rates as $\eta_x = 0.02$ and $\eta_y =0.05$, the number of random direction vectors as $q=5$,  and the smoothing parameter for gradient estimation as $\mu=10^{-3}$.
For \acma, referencing \cite{akimoto2022berthing}, we set  the threshold parameter for restart as $G_{\mathrm{tol}} = 10^{-6}$, the minimal learning rate as $\eta_{\min}=10^{-4}$, and the minimal standard deviation as $\bar{\sigma}_{\min} = 10^{-8}$. 
 For simplicity of the analysis, the restart strategy of \wracma{} and \wraslsqp{} was not used in these experiments.\footnote{
 The worst-case functions for our test problems are all single peak functions. On such problems, \wracma{} and \wraslsqp{} (i.e., CMA-ES) are expected to converge toward the optimal solution as long as the WRA mechanism approximates $\mathrm{Rank}_{F}(\{x_{i}\}_{i=1}^{\lambda_x})$ properly. Therefore, we omitted the restart strategy to investigate the goodness of WRA solely.
 }
\acma{} performed restart because it is implemented by default. 
We also turned off the diagonal acceleration mechanism both in CMA-ES for the outer minimization and inner maximization in \wracma{} 
for fair comparison of efficiency (to avoid the speed-up effect of the diagonal acceleration) in \Cref{fig:ex1} below.
\footnote{
We recommend to use the diagonal acceleration mechanism both in the outer and inner minimization in practice. The performance of the proposed approach on the test problems will not degrade with diagonal acceleration.
}

The performance of each algorithm is evaluated by $20$ independent trials. 
We regarded a trial as successful if $\abs{F(z)-F(x^*)} \leq 10^{-6}$ was satisfied for $z = m_x^t$ in case of \acma{}, \wracma{}, and \wraslsqp{} and for $z = x^t$ in case of 
\zopgda{}
before $10^7$ $f$-calls were spent. If the maximum $f$-calls were spent or some internal termination conditions were satisfied, we regard the trial as failed.

\subsection{Experiment 1} \label{sec:ex1}

To confirm that the proposed approach overcomes Difficulty (I), four approaches were applied to smooth convex--concave problems $f_5$, $f_{7}$ and $f_{11}$ for varying $b$ with and without bounds for the search domains. 
Note that the strength of the interaction between $x$ and $y$ is controlled by $b$ as the interaction term is $x^\T By$ and we set $B = \diag(b,.... ,b)$ in this experiment.

\providecommand{\exss}{0.5}
\begin{figure}[t]
\centering%
\begin{subfigure}{0.33\hsize}%
\centering%
    \includegraphics[width=\hsize]{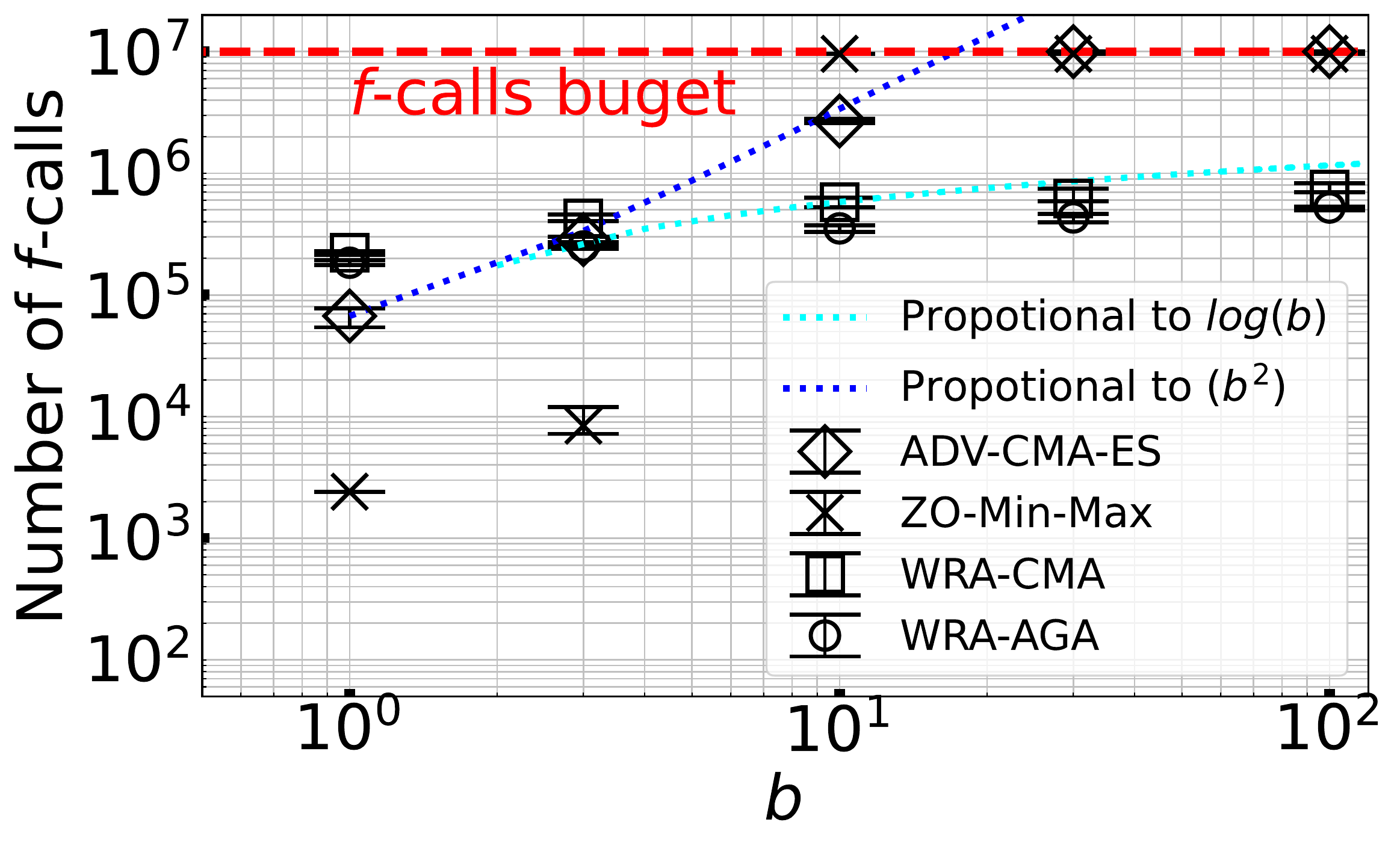}%
    \caption{$f_{5}$ (unbounded)}%
\end{subfigure}%
\begin{subfigure}{0.33\hsize}%
    \includegraphics[width=\hsize]{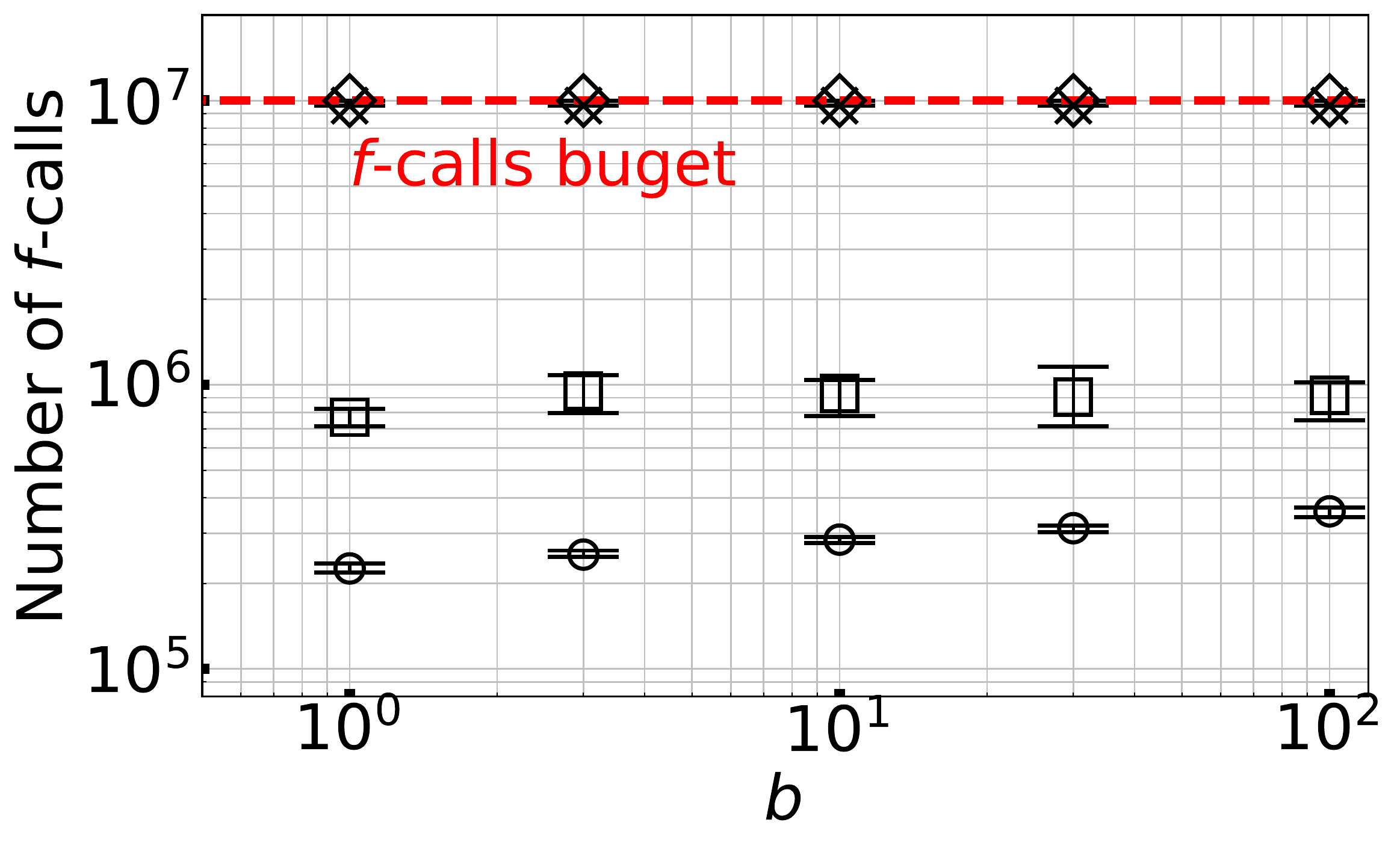}%
  \caption{$f_7$ (unbounded)}%
\end{subfigure}%
\begin{subfigure}{0.33\hsize}%
\centering%
    \includegraphics[width=\hsize]{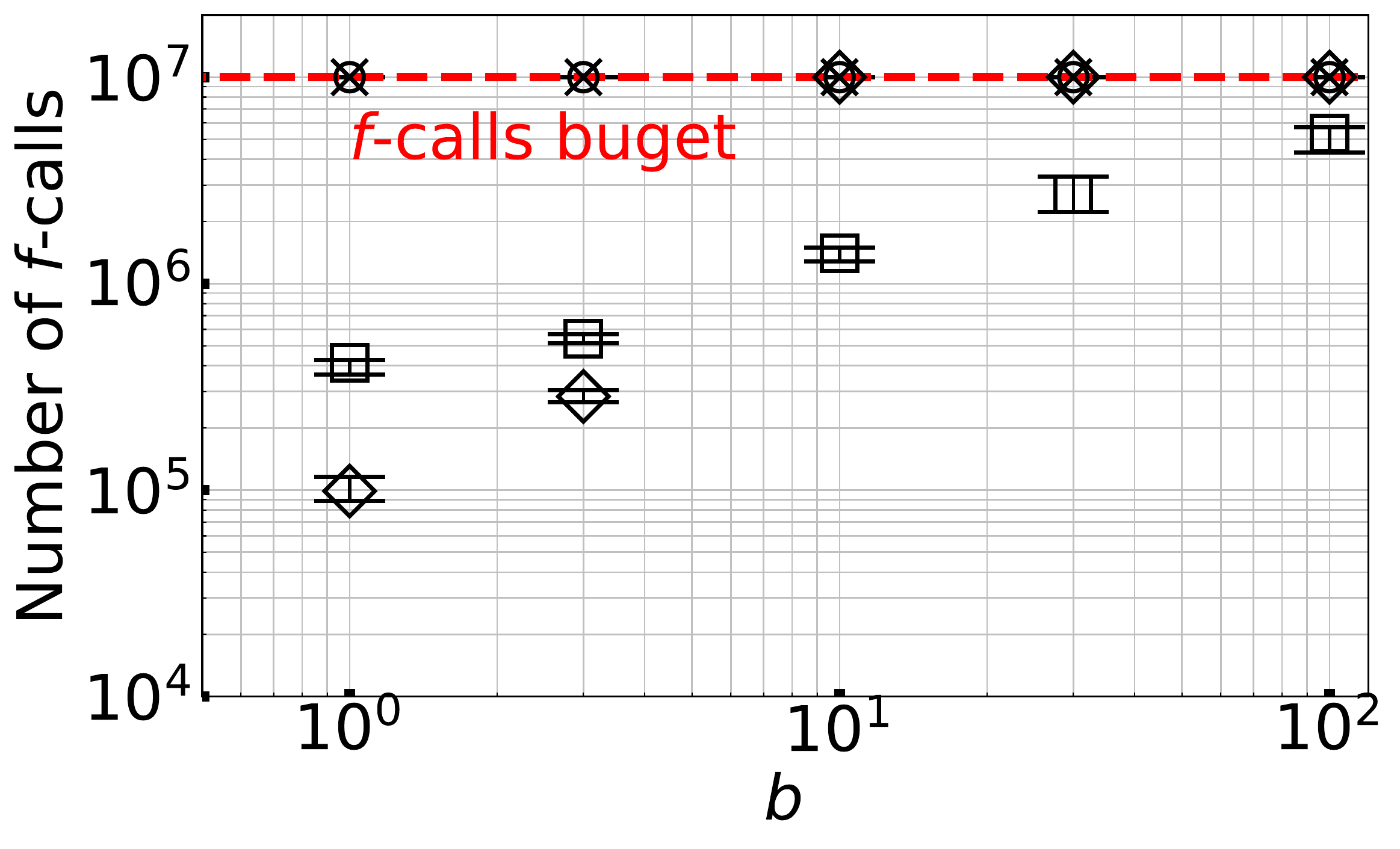}%
    \caption{$f_{11}$ (unbounded)}%
\end{subfigure}%
\\
\begin{subfigure}{0.33\hsize}%
\centering%
    \includegraphics[width=\hsize]{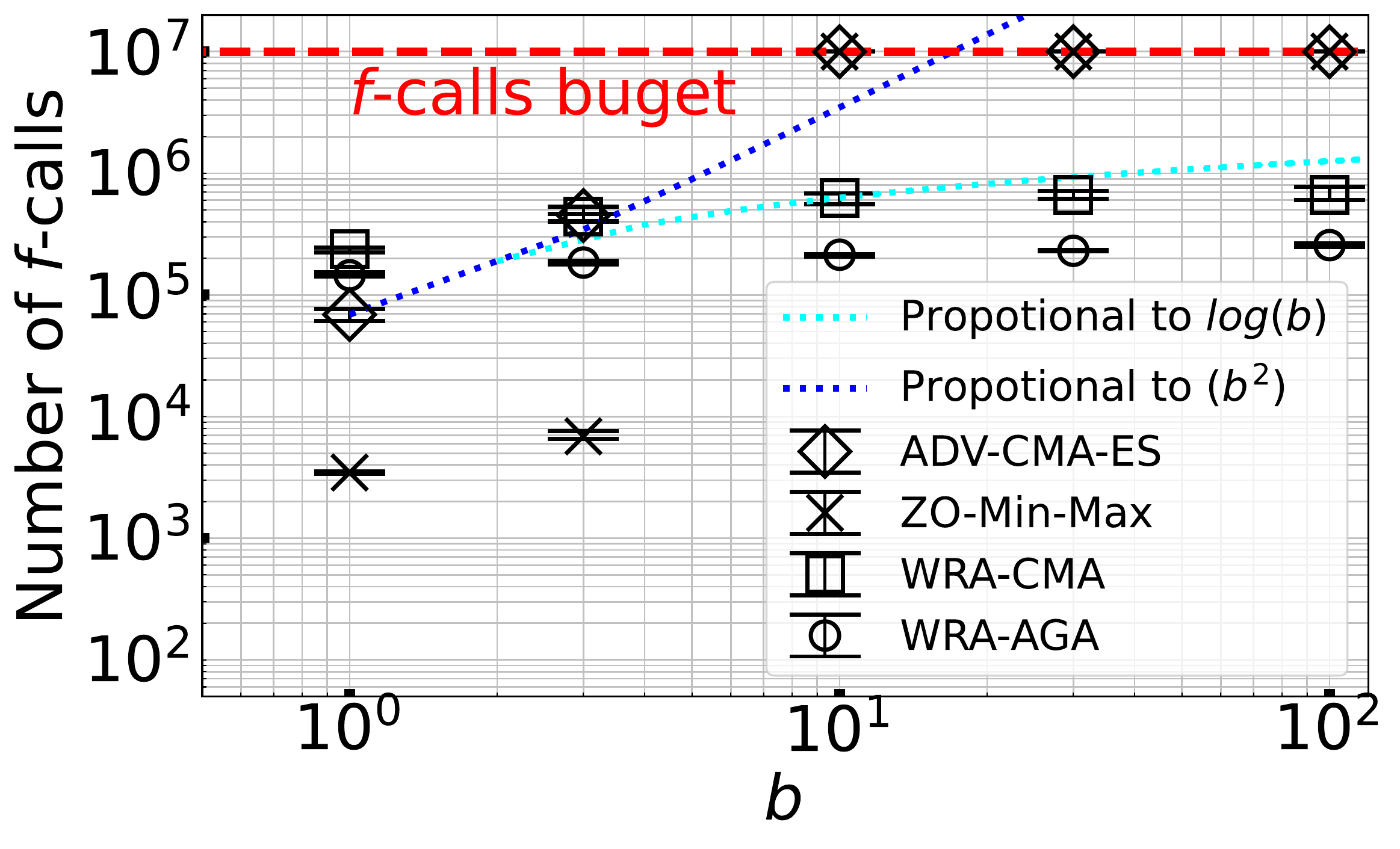}%
    \caption{$f_{5}$}%
\end{subfigure}%
\begin{subfigure}{0.33\hsize}%
    \includegraphics[width=\hsize]{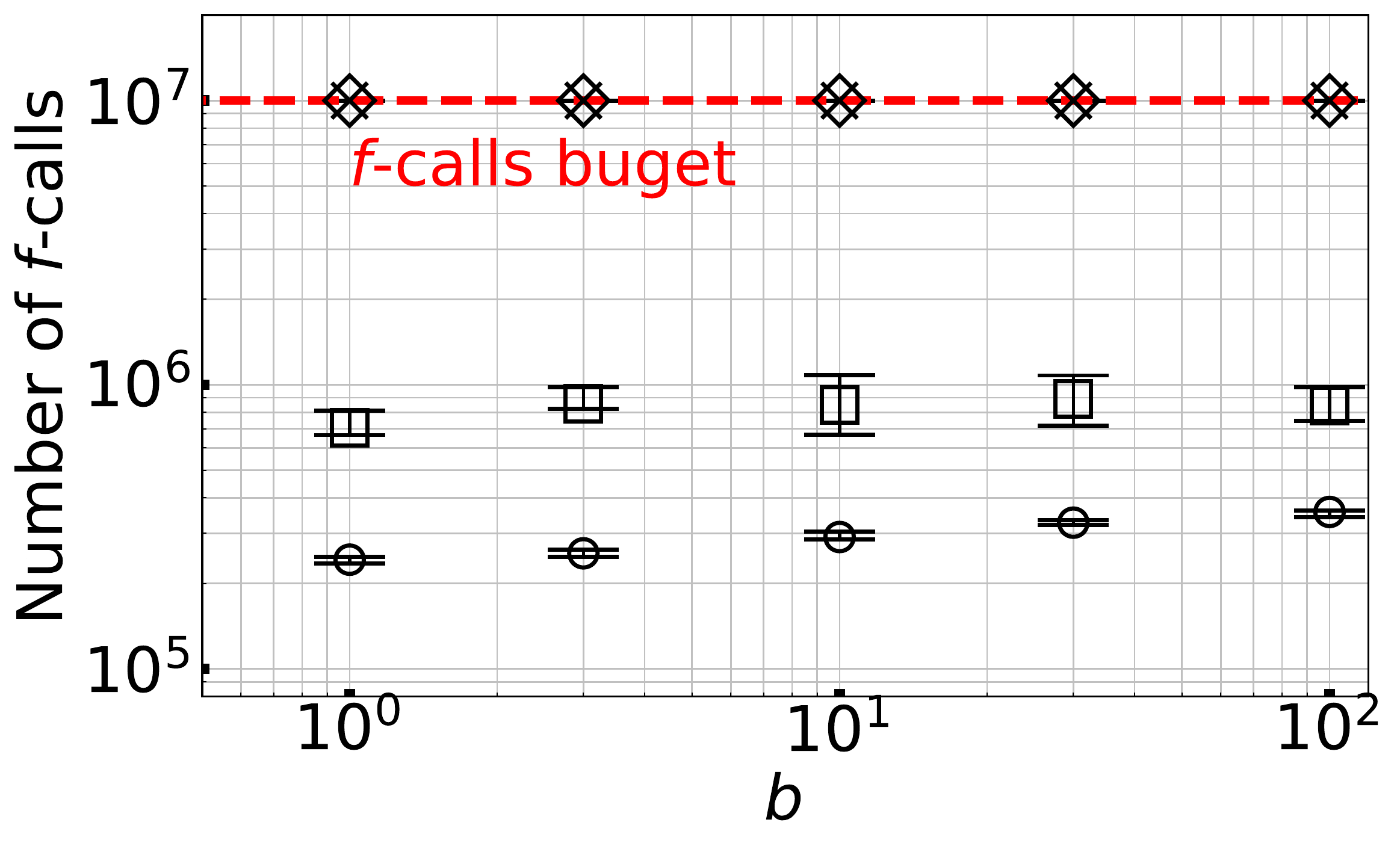}%
  \caption{$f_7$}%
\end{subfigure}%
\begin{subfigure}{0.33\hsize}%
\centering%
    \includegraphics[width=\hsize]{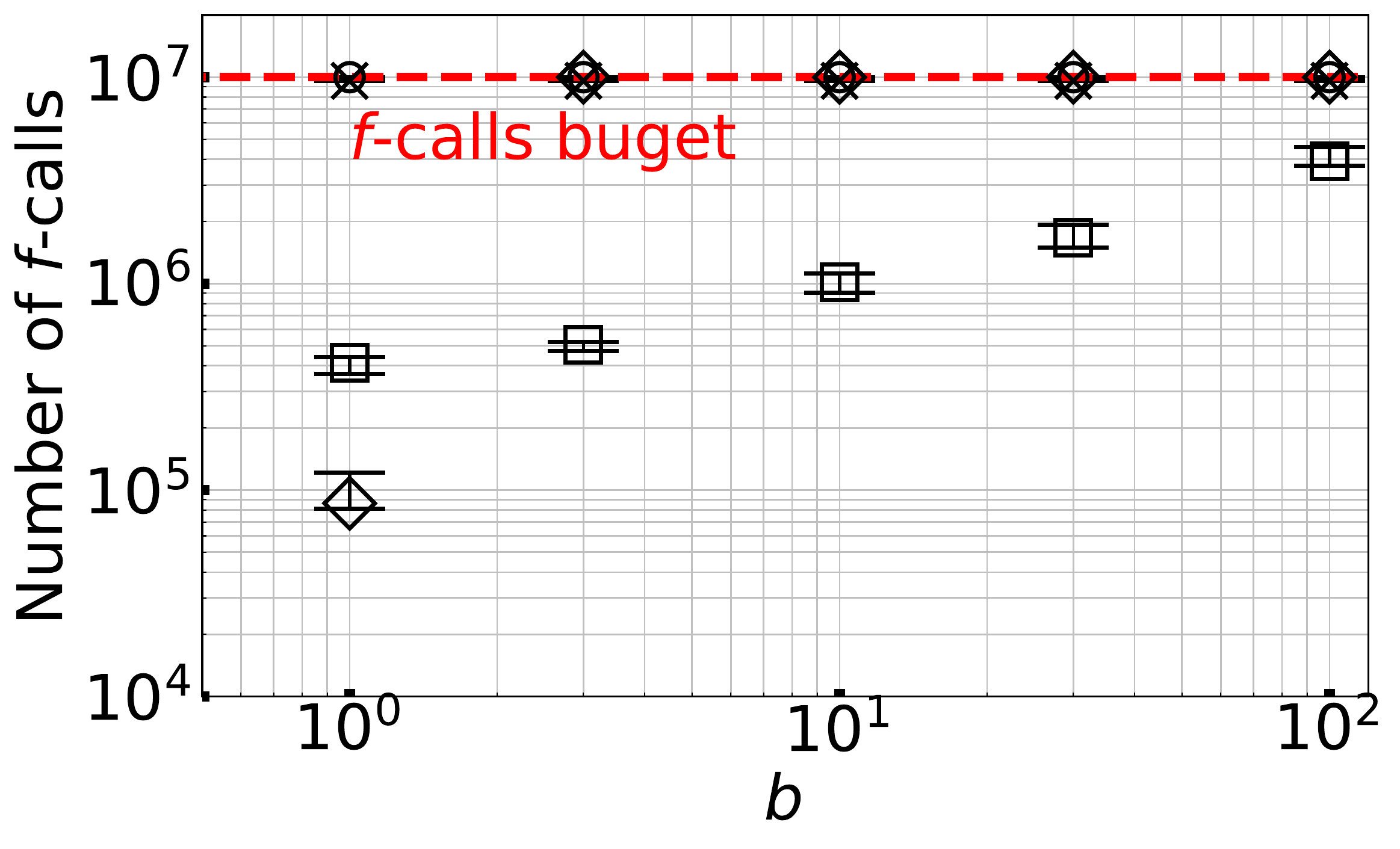}%
    \caption{$f_{11}$}%
\end{subfigure}%
\caption{
Median and interquartile range of the number of $f$-calls
spent by \wracma{}, \wraslsqp{}, \zopgda{}, and \acma{} over $20$ trials on $f_5$, $f_{7}$, and $f_{11}$ with $b \in \{1, 3, 10, 30, 100 \}$.
Note that the interquartile ranges were so small that the gaps between bars are barely visible in some cases.
Top: unbounded search domains ($\mathbb{X} = \R^{d_x}$ and $\mathbb{Y} = \R^{d_y}$). Bottom: bounded search domains.
}
\label{fig:ex1}
\end{figure}

\begin{figure}[t]
\centering%
\begin{subfigure}{0.48\hsize}%
\centering%
    \includegraphics[width=\hsize]{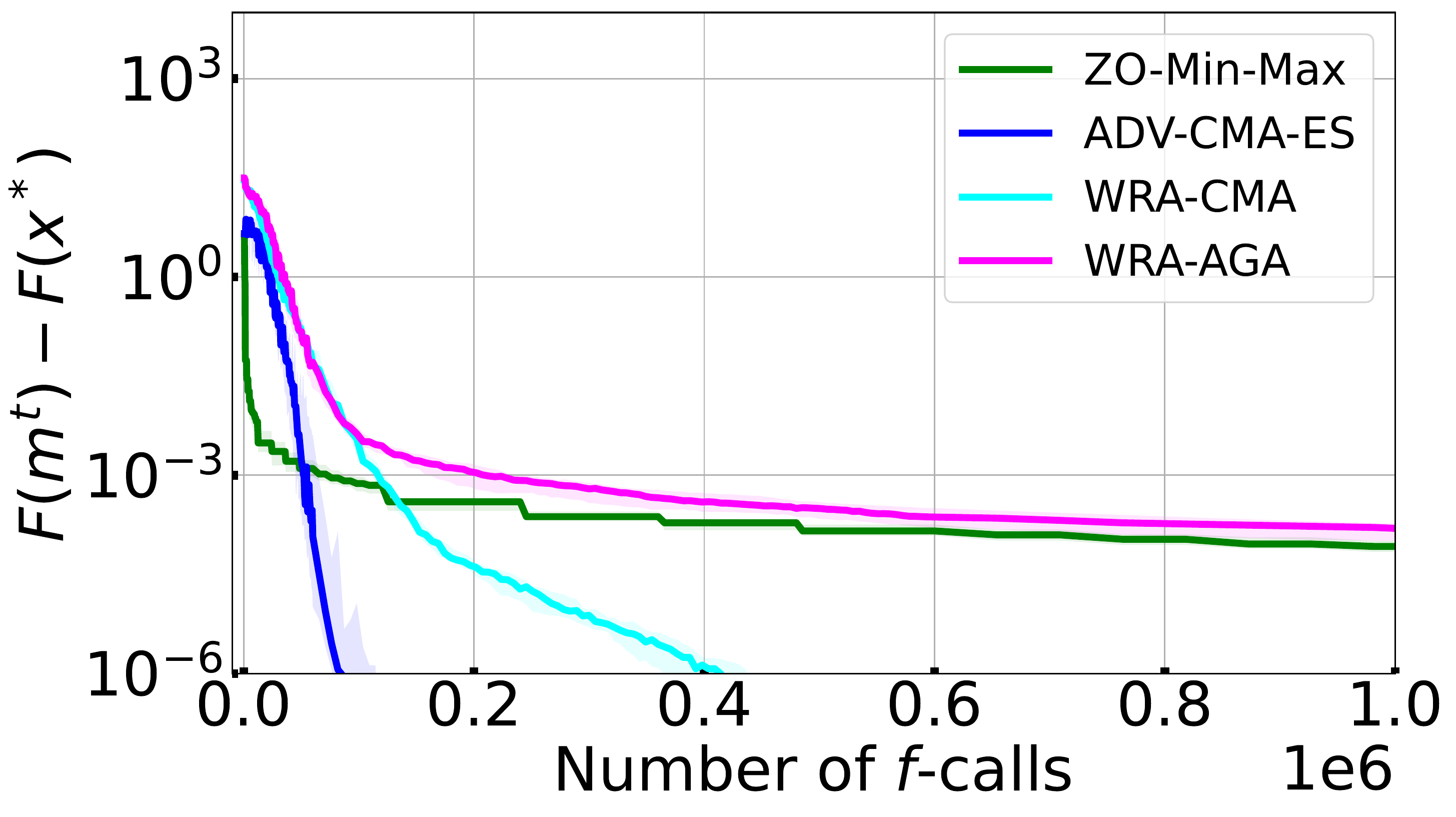}%
    \caption{$f_{11}$}%
\end{subfigure}%
\hfill
\begin{subfigure}{0.48\hsize}%
\centering%
    \includegraphics[width=\hsize]{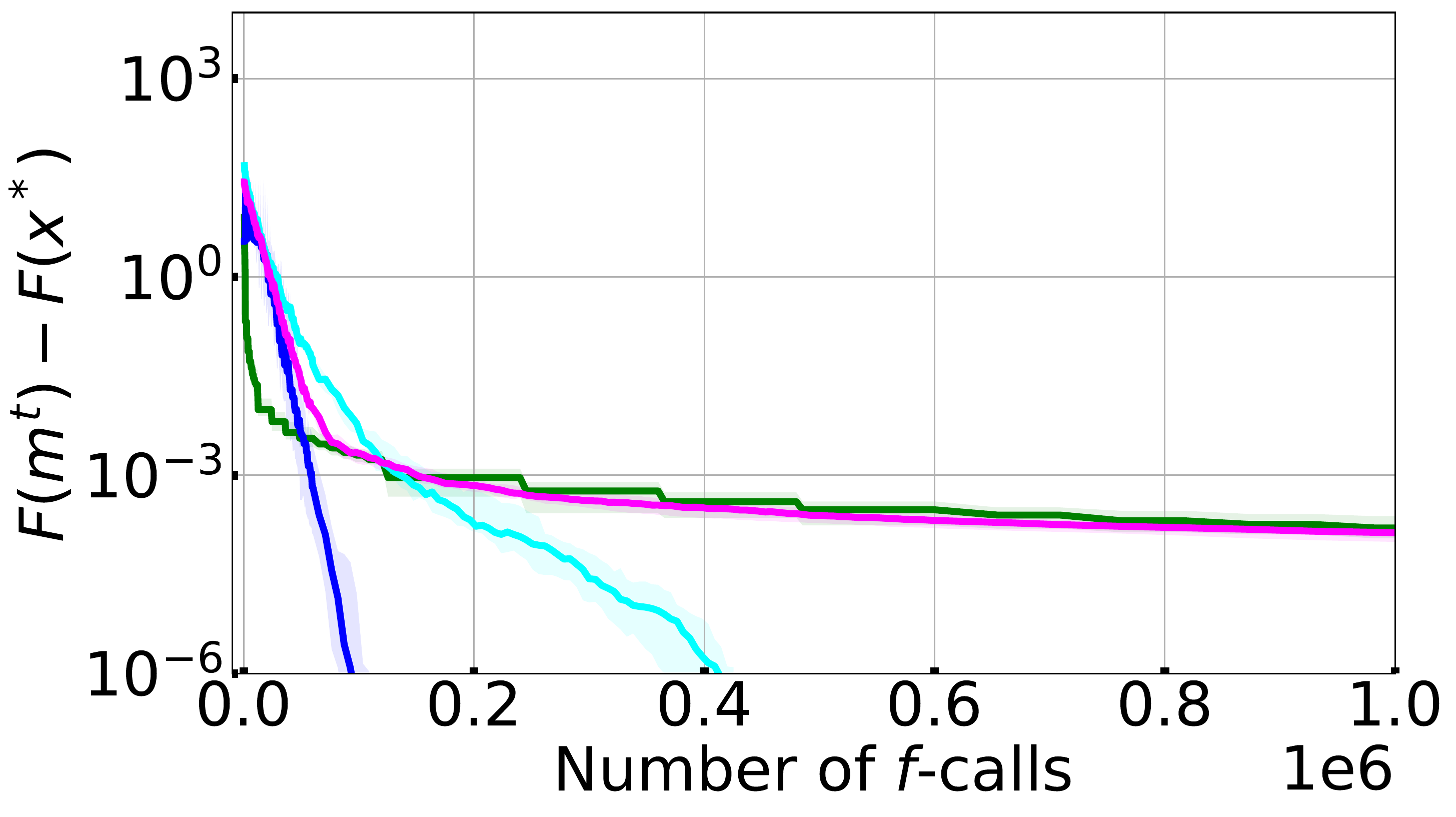}%
    \caption{$f_{11}$ (unbounded)}%
\end{subfigure}%
\caption{Gap $\abs{F(m^t) - F(x^*)}$ with the number of $f$-calls at $b=1$ on $f_{11}$.
Solid line: median (50 percentile) over 20 runs. Shaded area: interquartile range ($25$--$75$ percentile) over 20 runs. 
Note that the interquartile ranges were so small that the shaded areas are barely visible in some cases.
}
\label{fig:ex1_fcalls}
\end{figure}

\subsubsection{Results}

\Cref{fig:ex1} shows that \wracma{} and \wraslsqp{} could successfully optimize $f_5$ and $f_7$ with all $b \in \{1, 3, 10, 30, 100 \}$ in all trials, whereas \acma{} and \zopgda{} failed to optimize them except for $f_5$ with $b \leq 3$. 
\wracma{} was the only approach that successfully optimized $f_{11}$ with all $b$ values, whereas \acma{} could optimize $f_{11}$ with $b = 1$ with and without boundary and $b = 3$ without boundary. 
From these results, we confirm that both \zopgda{} and \acma{} fail in problems where the min--max solution is a global min--max saddle point but is not locally smooth and strongly convex--concave, and our approaches can solve such problems. 

When the search domain is unbounded on $f_5$, both \zopgda{} and \acma{} successfully locate near-optimal solutions for $b \leq 3$ with smaller $f$-calls than our approaches. However, for $b \geq 30$, they failed to converge, although $f_5$ is smooth and strongly convex--concave.
For $f_{11}$ with the unbounded search domain, \zopgda{} failed to converge at every trials, and \acma{} could not obtain successful convergence for $b \geq 10$, although $f_{11}$ is also smooth and strongly convex--concave.
For \zopgda{}, an inadequate learning rate may be a possible reason. 
For convergence, it must be tuned problem-dependently.
However, even if an appropriate value is set, the slow convergence issue discussed in \Cref{sec:limit} occurs.
For \acma{}, when $b \geq 30$ in $f_5$ and $f_{11}$ with the unbounded search domain, slow convergence issue is the main reason, as the expected $f$-calls (blue dash line in \Cref{fig:ex1}) in $f_5$ exceeded $f$-call budget. 
When the search domain is unbounded, we expect \acma{} to obtain the successful convergence for $f_{11}$ until $b=10$ similarly as $f_5$. However, \acma{} failed to converge in $f_{11}$ with $b=10$. We observed that \acma{} suffered to approach $x^*$ because the learning rate reached to lower bound $\eta_{\min}$. Therefore, to obtain successful convergence for $f_{11}$ with $b=10$, lower bound $\eta_{\min}$ for \acma{} should be properly set.
When there was a bound for search domain, \acma{} failed to converge with $b=10$ for $f_{5}$ and $b=3$ for $f_{11}$.  

The difference between \wracma{} and \wraslsqp{} is in the speed of convergence for $f_5$ and $f_7$ as well as the performance for $f_{11}$. 
For $f_5$ and $f_7$, \wraslsqp{} converged faster than \wracma{}. 
Meanwhile, \wraslsqp{} failed to optimize $f_{11}$ within a given $f$-call budget.
\Cref{fig:ex1_fcalls} shows gap $F(m^t) - F(x^*)$ on $f_{11}$ with $b=1$.
From \Cref{fig:ex1_fcalls}, we expect that \wraslsqp{} eventually converges, but the convergence speed is very slow. 
Preliminary, we confirmed that \Cref{alg:SLSQPinwra} converges slowly on ill-conditioned function.  
Therefore, on $f_{11}$ which is ill-conditioned in $y$, \Cref{alg:SLSQPinwra} with small $c_{\max}$ cannot significantly improve the worst-case scenario candidate $\hat{y}$ and the early-stopping strategy may terminate the inner maximization process before approximating the worst-case scenario in adequate accuracy. Because of the underestimation of the  rankings on the worst-case objective function$\mathrm{Rank}_F\{(x_i)_{i=1}^{\lambda_x})\}$ by WRA, 
the outer minimization
failed to converge at the global min--max solution, indicating the relevance of the choice of the inner solver for the WRA mechanism. Because the CMA-ES is a variable metric approach and the covariance matrices are inherited over WRA calls, \wracma{} could optimize $f_{11}$ efficiently. 

\subsubsection{Discussion on the effect of the interaction term} \label{sec:dis}

We discuss the effect of the worst-case scenario sensitivity (coefficient matrix $B$ of the interaction term $x^T B y$) on $f$-calls spent by our approaches when the objective function is convex--concave.
\Cref{fig:ex1} shows that the numbers of $f$-calls were in $O(\log(b))$ or even in $O(1)$ in terms of the coefficient of the interaction term, $b$. We provide a brief but not rigorous explanation of these results.

For simplicity, we focus on $f_5$ with ${d_x} = {d_y}$ and $B = \diag(b, \dots, b)$. 
The worst-case objective function is $F(x)=\frac{(1+b^2)}{2} \norm{x}_2^2$ and the worst-case scenario is $\yw(x) = b x$ in this case.
Moreover, we focus on \wracma{}.

To proceed, we assume that the CMA-ES converges linearly for such spherical functions. That is, a point in  $\{x : \norm{x-x^*} \leq \epsilon \cdot \norm{m_x^0 - x^*}\}$ around the optimal solution $x^*$ can be found in $O(\log(\norm{m_x^0 - x^*}/\epsilon))$ $f$-calls. Although no rigorous runtime analysis has been performed for the CMA-ES, we have ample empirical evidence. Moreover, (1+1)-ES, which is a simplified version of the CMA-ES, converges linearly on Lipschitz smooth and strongly convex objective functions \cite{aagm2022siam}.

First, we consider how many iterations the CMA-ES for the outer minimization spends to reach a point $m_x^{T_{\epsilon}}$ such that $\norm{m_x^{T_\epsilon} - x^*} \leq \epsilon \norm{m_x^0 - x^*}$. We call $T_\epsilon$ the runtime.
WRA returns approximate rankings of given solution candidates, and they highly correlate with the true rankings.
Then, the CMA-ES is expected to behave similarly in these two rankings. 
Therefore, if the CMA-ES converges linearly for $F$, we expect that the CMA-ES converges linearly for the rankings given by WRA as well, which is partly supported by a theoretical investigation \cite{surrogatetheory}. 
Because the worst-case objective function $F(x)$ is spherical, the CMA-ES is expected to converge linearly, i.e., the runtime $T_\epsilon$ is in $O(\log(\norm{m_x^0 - x^*} / \epsilon))$. 

Next, we consider how many $f$-calls \wracma{} spends in each call. 
Let the current search distribution of the CMA-ES for the outer minimization be $\mathcal{N}(m_x^t, \Sigma_x^t)$. 
Because $F$ is spherical, $\Sigma_x^t$ is expected to be proportional to the identity matrix $I_{d_x}$. 
Let us assume that $\Sigma_x^t \approx \sigma_t^2 I_{d_x}$.
Then, the solution candidates $x_1, \dots, x_{\lambda_x}$ given to WRA are independently $\mathcal{N}(m_x^t, \sigma_t^2 I_{d_x})$-distributed.
For the two solution candidates $x_i$ and $x_j$, the expected difference in the worst-case objective function values is as follows:
\begin{align}
 \E[(F(x_i)-F(x_j))^2]^{1/2} 
  =(1+b^2)\Tr((\Sigma_x^t)^2)^{1/2} \approx {d_x}^{1/2} (1+b^2) \sigma_t^2. \label{eq:Fdistance}
\end{align}

The early-stopping strategy is expected to stop the maximization process once the rankings of the given candidate solutions are well-approximated in terms of Kendall's rank correlation. To have a high value of the Kendall's rank correlation, the orders of $F(x_i)$ and $F(x_j)$ and their approximate values, $f(x_i, \tilde{y}_i)$ and $f(x_j, \tilde{y}_j)$ must be concordant with high probability for each pair $(x_i, x_j)$ among $\lambda_x$ solution candidates $x_1, \dots, x_{\lambda_x}$, where $\tilde{y}_i$ ($i=1,\dots,\lambda_x$) is the approximate worst-case scenario for $x_i$ obtained in WRA. 
It suffices to obtain $\tilde{y}_i$ and $\tilde{y}_j$ such that $\abs{F(x_i)-f(x_i, \tilde{y}_i)}$ and $\abs{F(x_j)-f(x_j, \tilde{y}_j)}$ are both less than $\abs{F(x_i)-F(x_j)}$.
With a simple derivation, we obtain
\begin{align}
    F(x) - f(x, \tilde{y}) 
    &= \frac{1}{2}\norm{\hat{y}(x) - \tilde{y}}^2. \label{eq:Fprecise}
\end{align}
That is, if $\norm{\hat{y}(x_i) - \tilde{y}_i} \leq c \cdot (1 + b^2)^{1/2} \sigma_t$ is satisfied for some $c > 0$, the true order of the two points among $\lambda_x$ solution candidates will be correctly identified with high probability. 
Because the objective function of the inner maximization problem is spherical in $y$, to obtain such approximate worst-case scenarios, the required $f$-calls is $O\left(\log \left(\frac{\norm{y^{(0)} - \hat{y}(x_i)}}{c \cdot (1 + b^2)^{1/2} \sigma_t}\right)\right)$, where $\tilde{y}^{(0)}$ denotes the initial scenario for $x_i$. 

Assuming that the Gaussian distribution of the CMA-ES for the outer loop does not change significantly from the previous iteration, the worst-case scenario for the solution candidate in the current iteration, $\hat{y}(x)$, and that in the previous iteration are expected to follow the distribution $\mathcal{N}(bm_x^t, b^2\Sigma_x^t)$. 
Because the 
warm-starting
strategy selects the worst-case scenario among the set of scenarios including the ones obtained in the previous WRA call, the distance between $\yw{(x)}$ and $\tilde{y}$ is expected to be no greater than $\E[\norm{\hat{y}(x)-\tilde{y}^{(0)}}] = b^2 \Tr(\Sigma_x^t)={d_x} b^2 \sigma_t^2$. From this, we estimate $\norm{y^{(0)}-\hat{y}(x)} \in O(b\sigma_t)$. 
As a result, 
the number of $f$-calls
required to approximate the worst-case objective function values for each solution candidate is $O\left(\log \left(\frac{b}{c \cdot (1 + b^2)^{1/2}}\right)\right)$.

Altogether, the proposed approach is expected to locate a near-optimal solution with 
\begin{equation}
    O\left( \lambda_y \log \left(\frac{b}{c \cdot (1 + b^2)^{1/2}}\right) \cdot \log\left( \frac{\norm{m_x^{(0)} - x^*}}{\epsilon}\right)\right)  \label{eq:fcallorder}
\end{equation}
$f$-calls.
It scales as $\log{(b)}$ for $b\leq 1$ and is constant for $b \to \infty$, which well-estimates the behavior observed in \Cref{fig:ex1}.

\begin{figure}[t]
\centering%
\begin{subfigure}{0.33\hsize}%
    \includegraphics[width=\hsize]{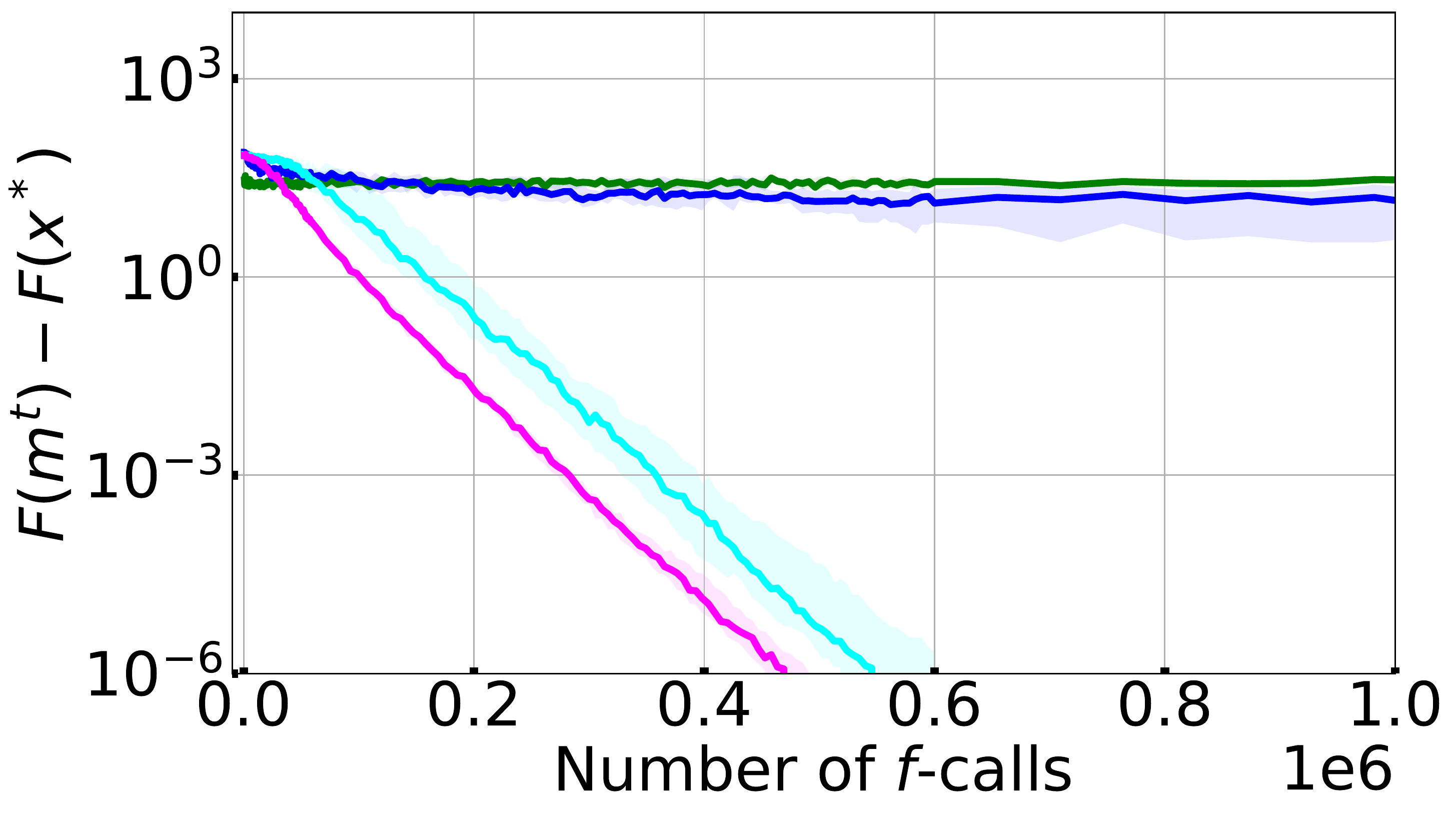}%
  \caption{$f_1$}%
\end{subfigure}%
\begin{subfigure}{0.33\hsize}%
    \includegraphics[width=\hsize]{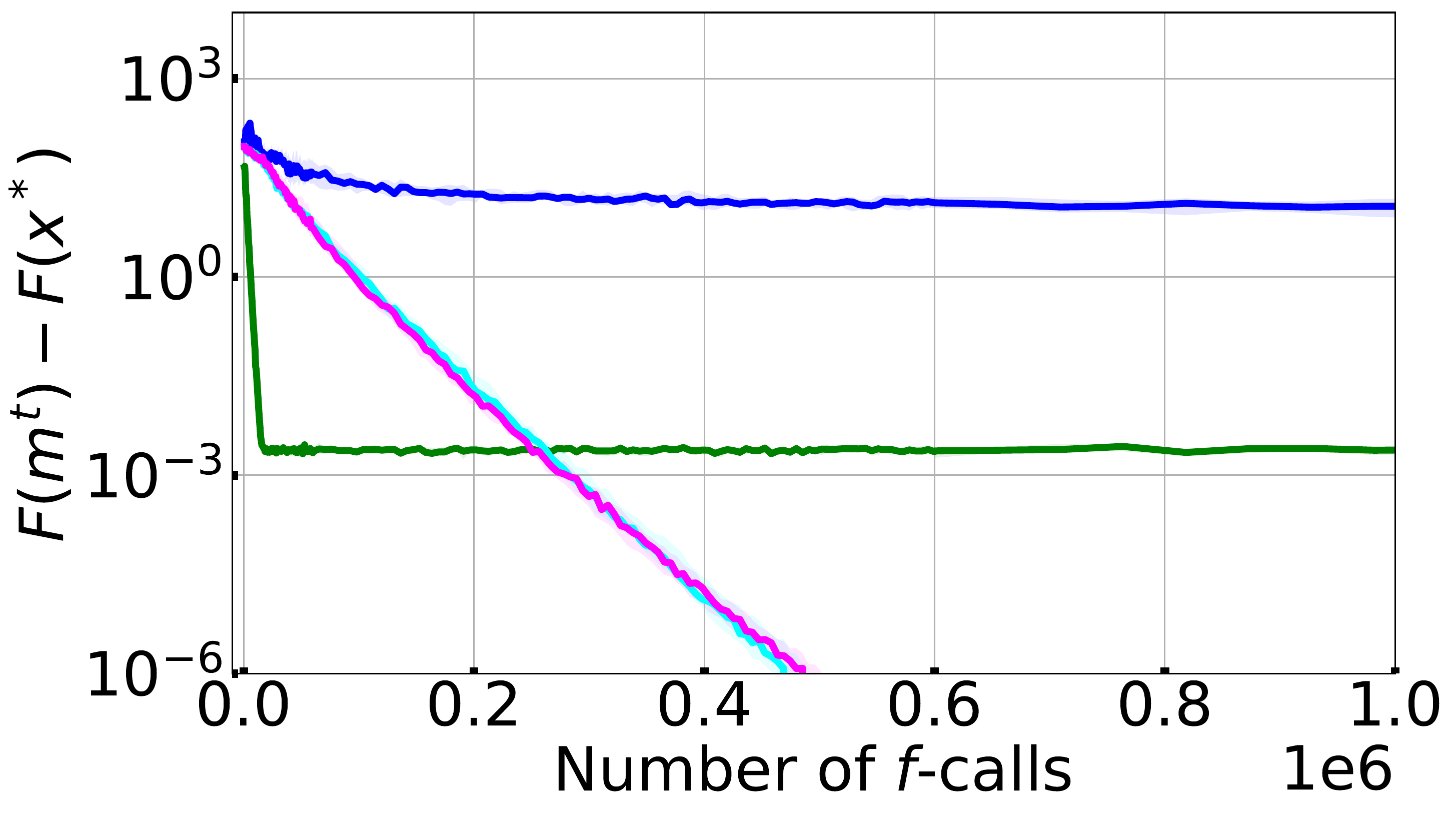}%
  \caption{$f_2$}%
\end{subfigure}%
\begin{subfigure}{0.33\hsize}%
    \includegraphics[width=\hsize]{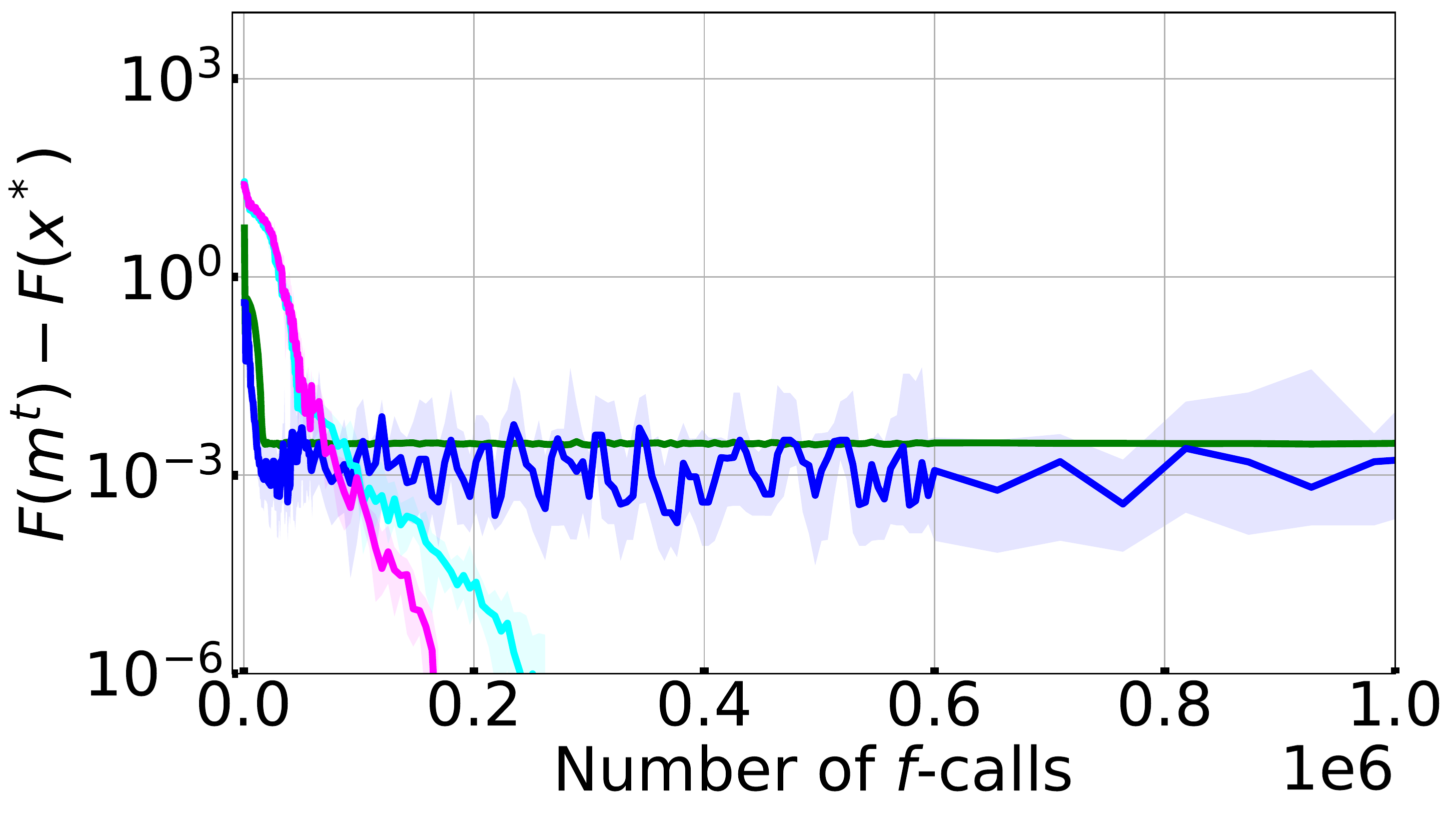}%
  \caption{$f_3$}%
\end{subfigure}%
\\
\begin{subfigure}{0.33\hsize}%
    \includegraphics[width=\hsize]{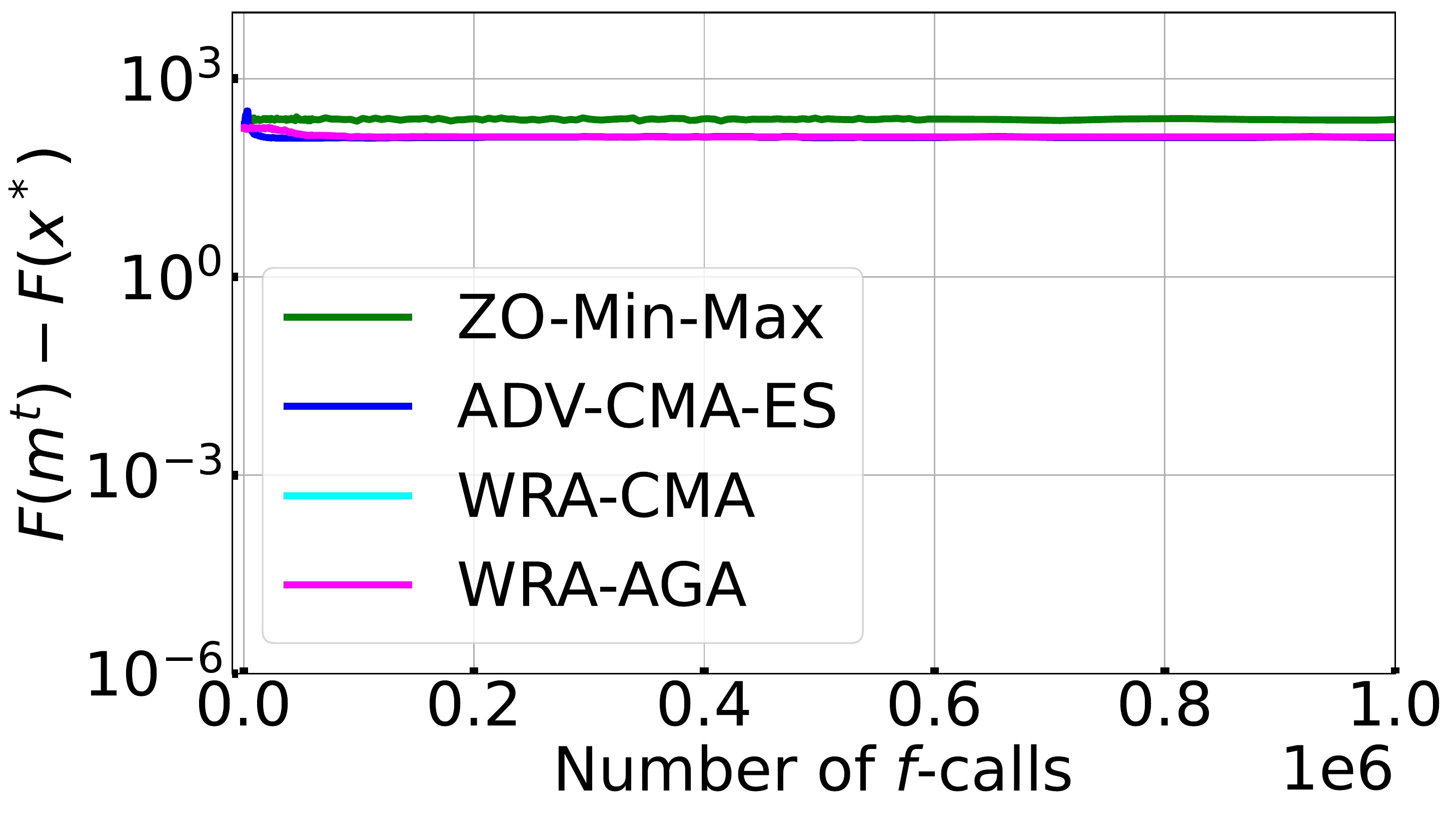}%
  \caption{$f_4$}%
\end{subfigure}%
\begin{subfigure}{0.33\hsize}%
    \includegraphics[width=\hsize]{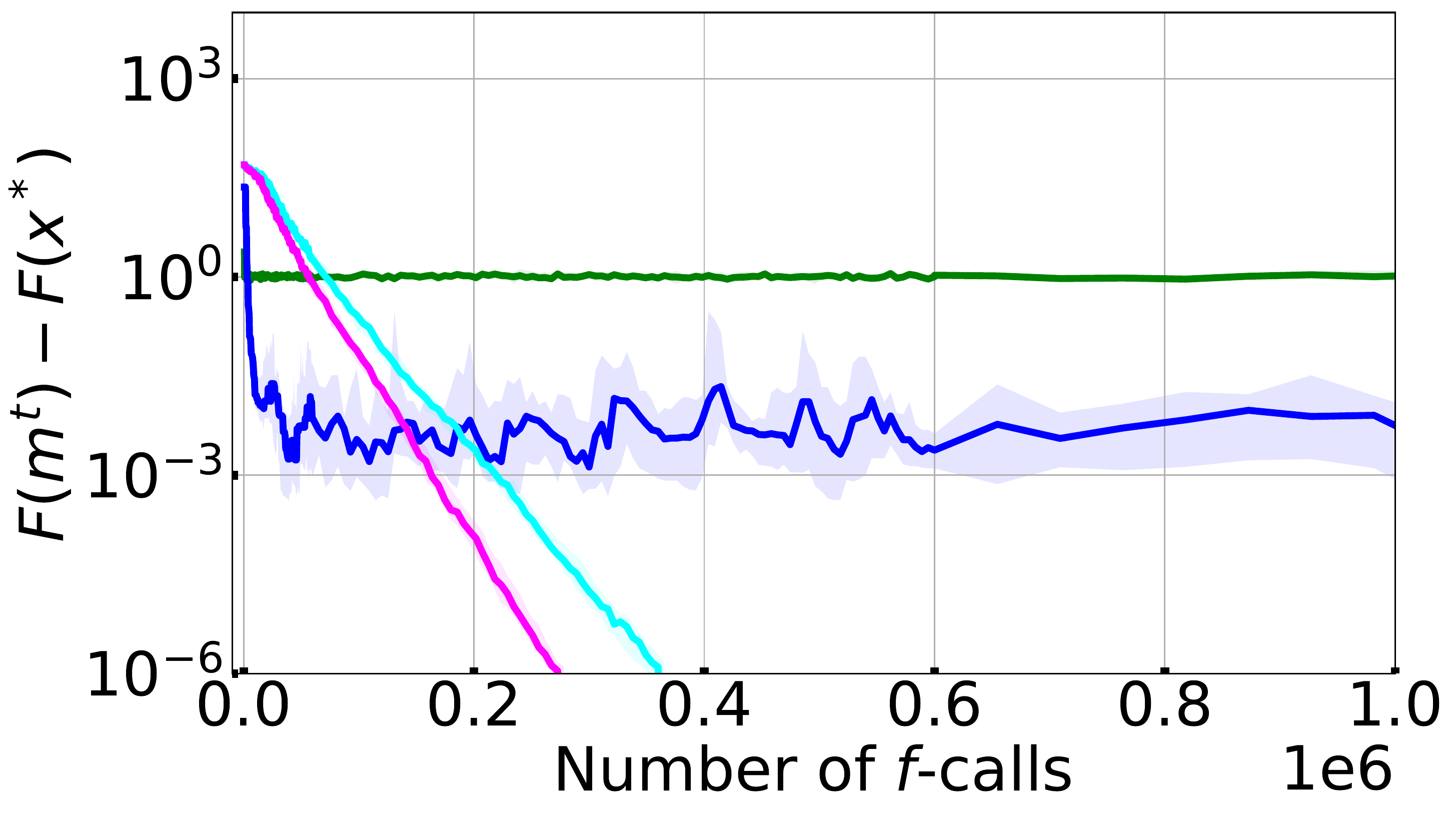}%
  \caption{$f_6$}%
\end{subfigure}%
\begin{subfigure}{0.33\hsize}%
    \includegraphics[width=\hsize]{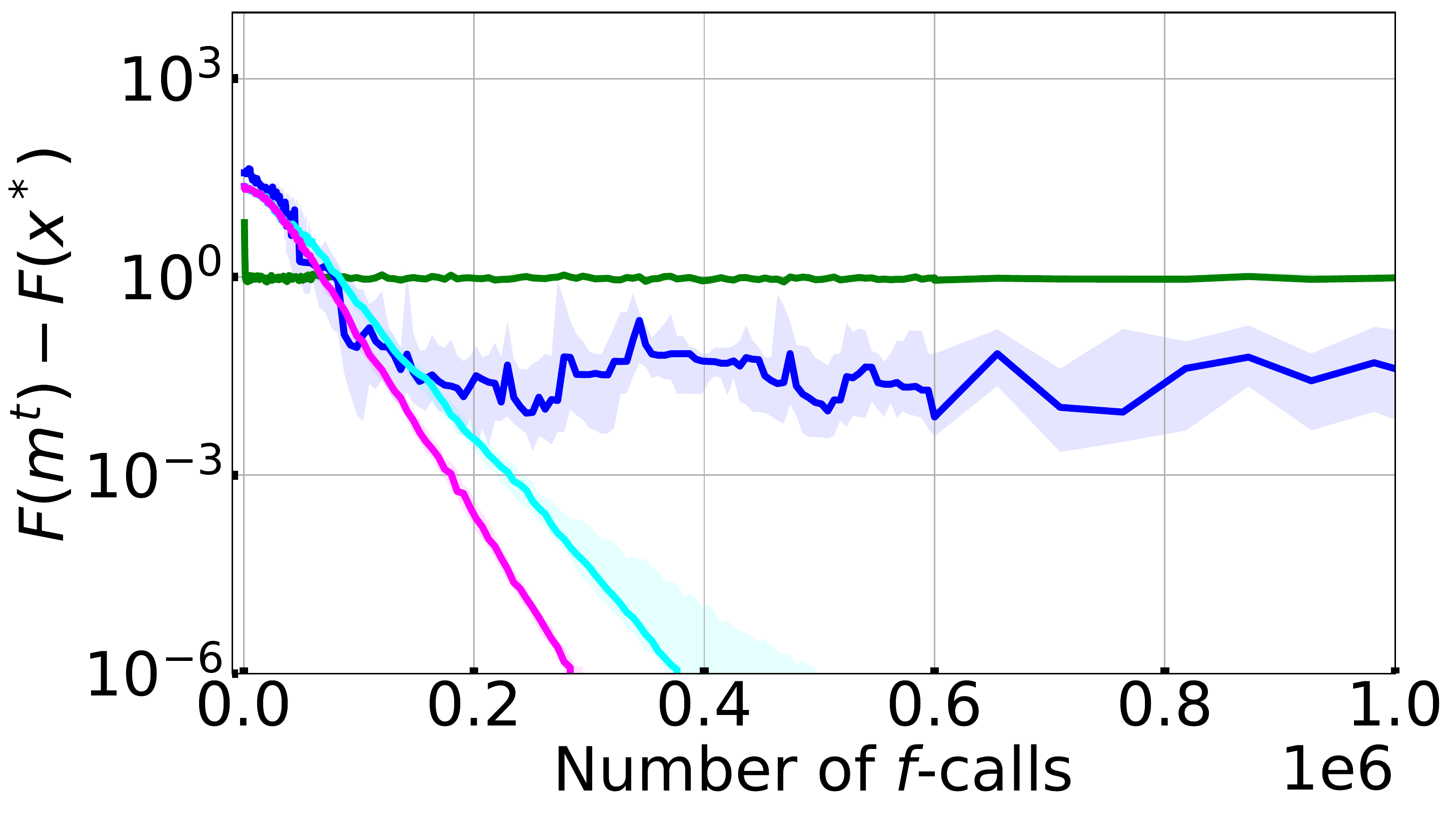}%
  \caption{$f_8$}%
\end{subfigure}%
\\
\begin{subfigure}{0.33\hsize}%
    \includegraphics[width=\hsize]{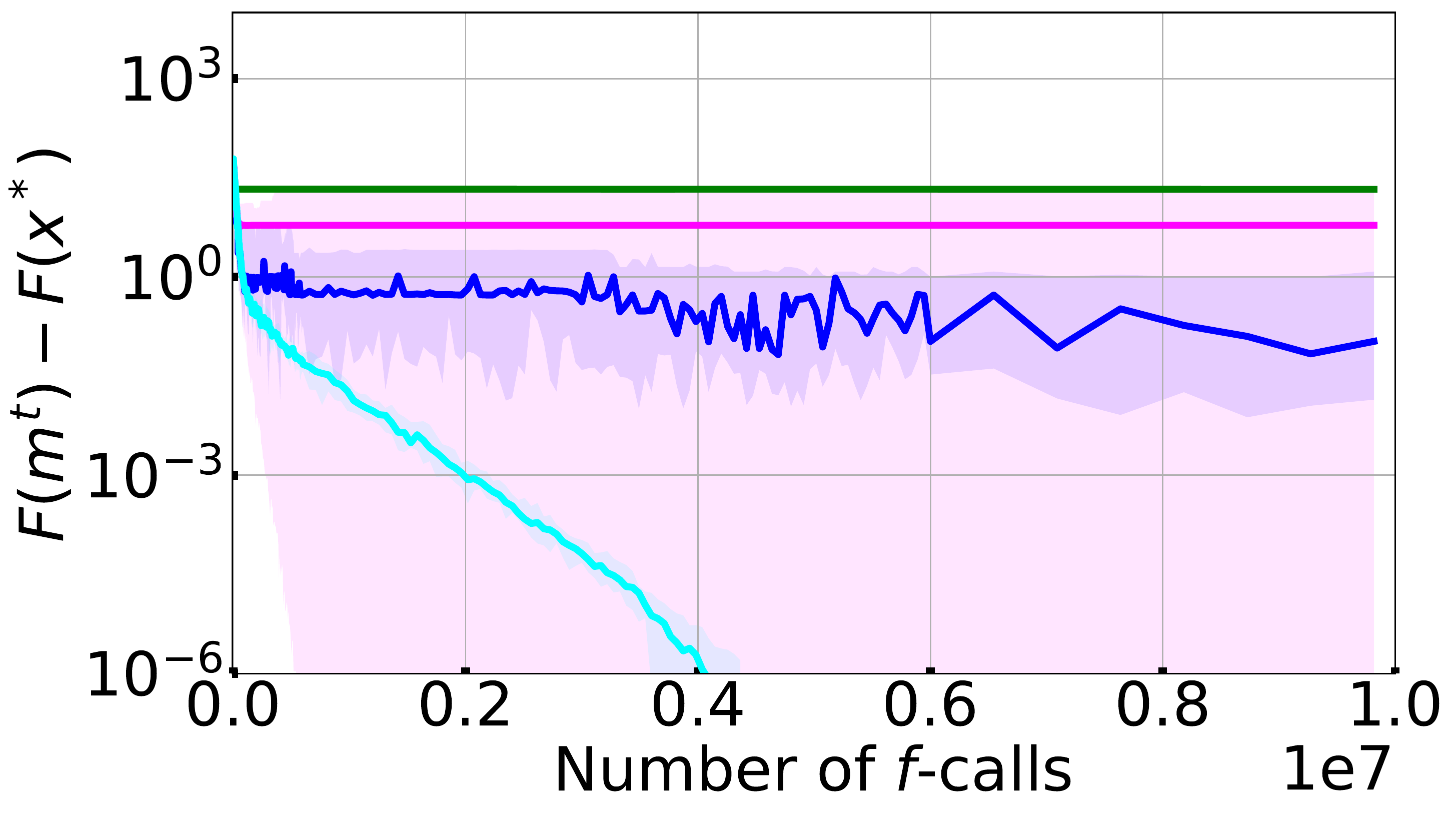}%
  \caption{$f_9$}%
\end{subfigure}%
\begin{subfigure}{0.33\hsize}
    \includegraphics[width=\hsize]{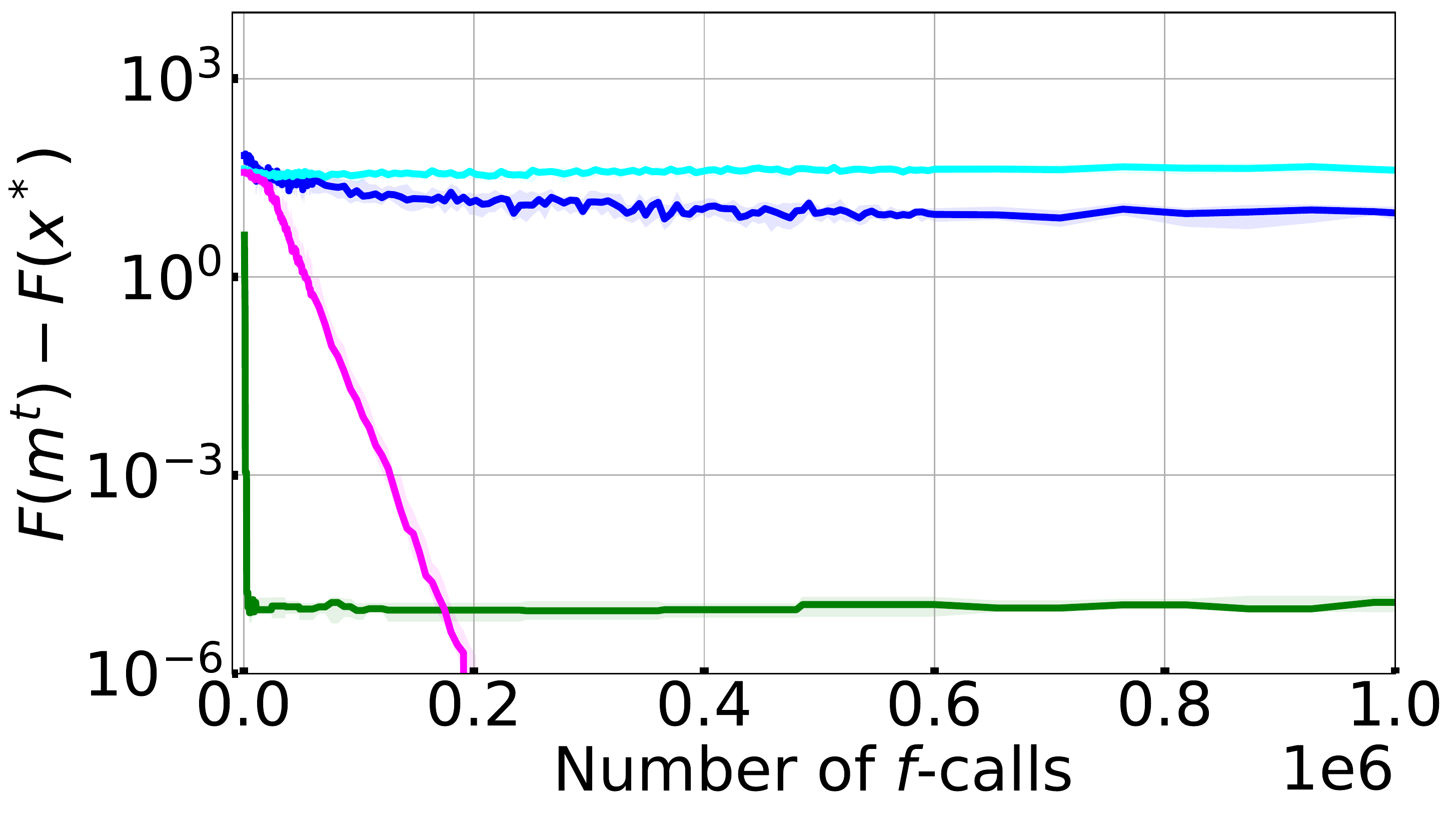}%
  \caption{$f_{10}$}%
  \label{fig:f10}
\end{subfigure}
\begin{subfigure}{0.33\hsize}%
    \includegraphics[width=\hsize]{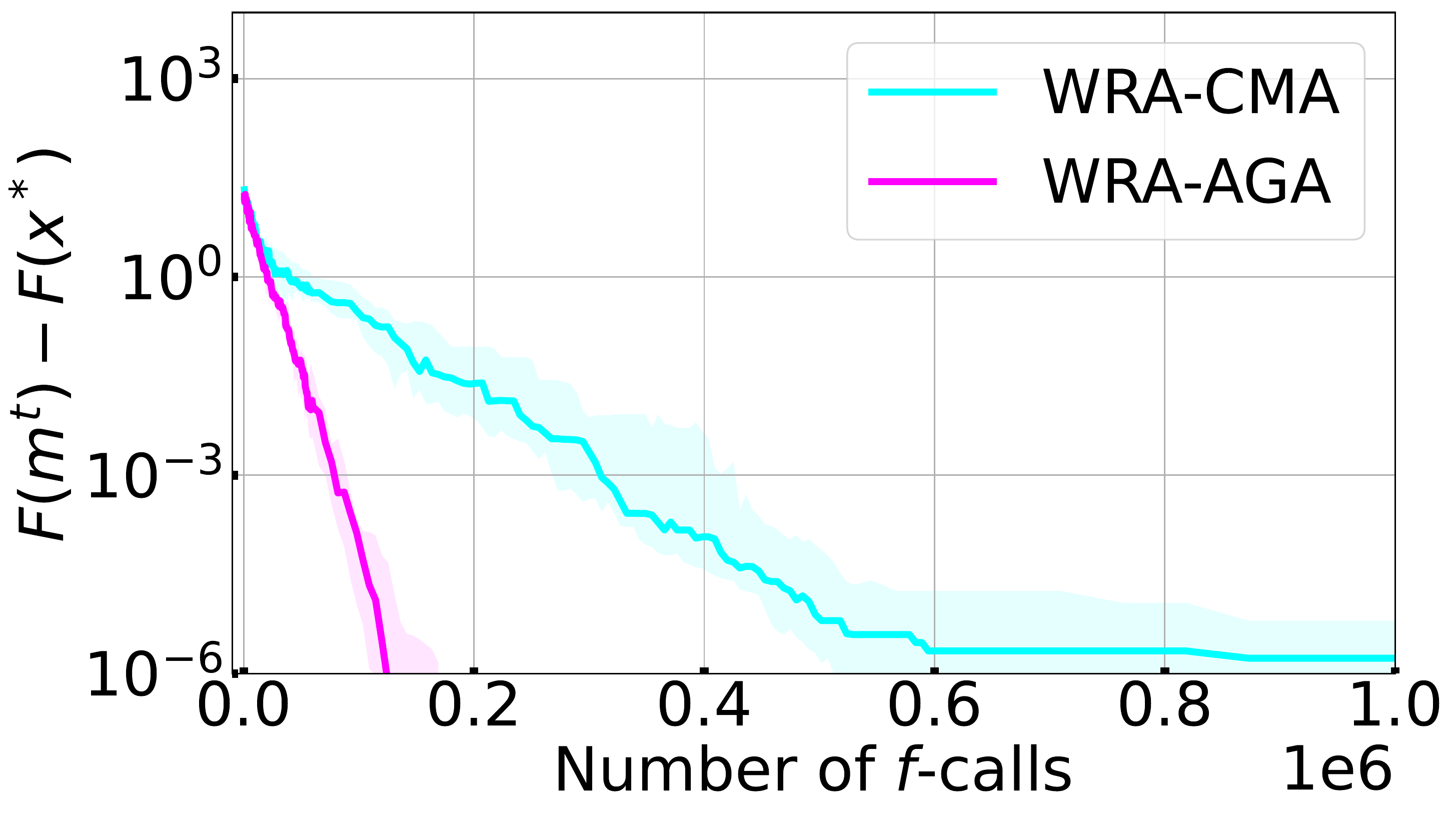}%
   \caption{$f_4$ (${d_x}={d_y}=5$)}%
\label{fig:ex2_f4}%
\end{subfigure}%
\caption{Gap $\abs{F(m^t) - F(x^*)}$ with the number of $f$-calls at $b=1$ on $f_1$--$f_{4}$, $f_6$, and $f_{8}$--$f_{10}$.
Solid line: median (50 percentile) over 20 runs. Shaded area: interquartile range ($25$--$75$ percentile) over 20 runs.
[R1C17] Note that the interquartile ranges were so small that the shaded areas are barely visible in some cases.
}%
\label{fig:ex2}
\end{figure}

\subsection{Experiment 2} \label{sec:ex2}

We applied four approaches to the problems $f_1$--$f_4$, $f_6$, and $f_8$--$f_{10}$ to investigate their performance on the functions that are not smooth and strongly convex--concave.

The results are shown in \Cref{fig:ex2}. We confirm that near-optimal solutions were obtained by \wracma{} for $f_1$--$f_3$, $f_6$, $f_8$, and $f_9$ and by \wraslsqp{} for $f_1$--$f_3$, $f_6$, $f_8$, and $f_{10}$. Moreover, the existing approaches failed to locate near-optimal solutions in all trials.

\subsubsection{Category (S) ($f_3$, $f_6$, and $f_8$)}

\begin{figure}[t]
\centering
\begin{subfigure}{0.48\hsize}%
    \includegraphics[width=\hsize]{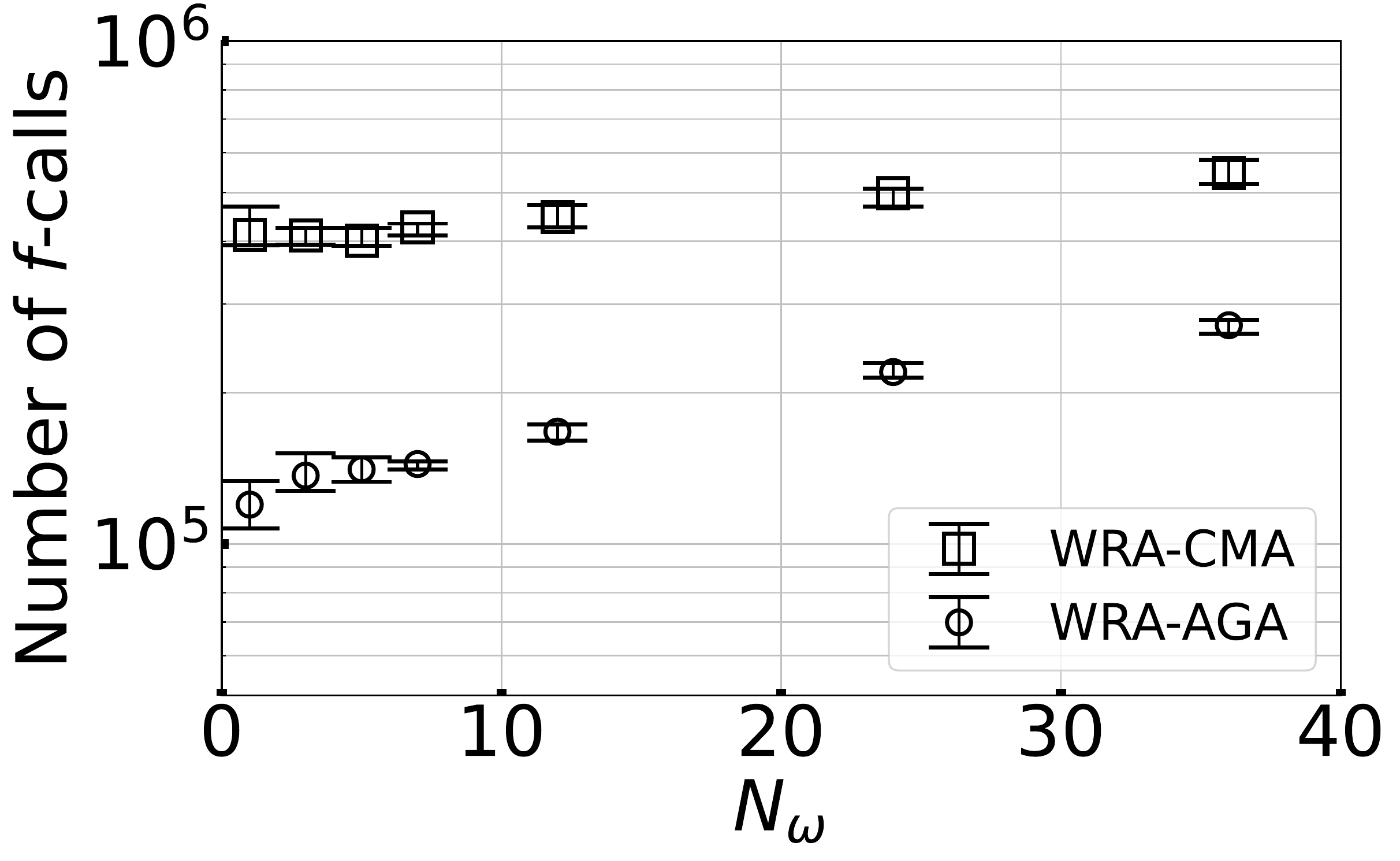}%
    \caption{$f_8$}%
    \label{fig:ex2_f8}%
\end{subfigure}%
\hfill
\begin{subfigure}{0.48\hsize}%
    \includegraphics[width=\hsize]{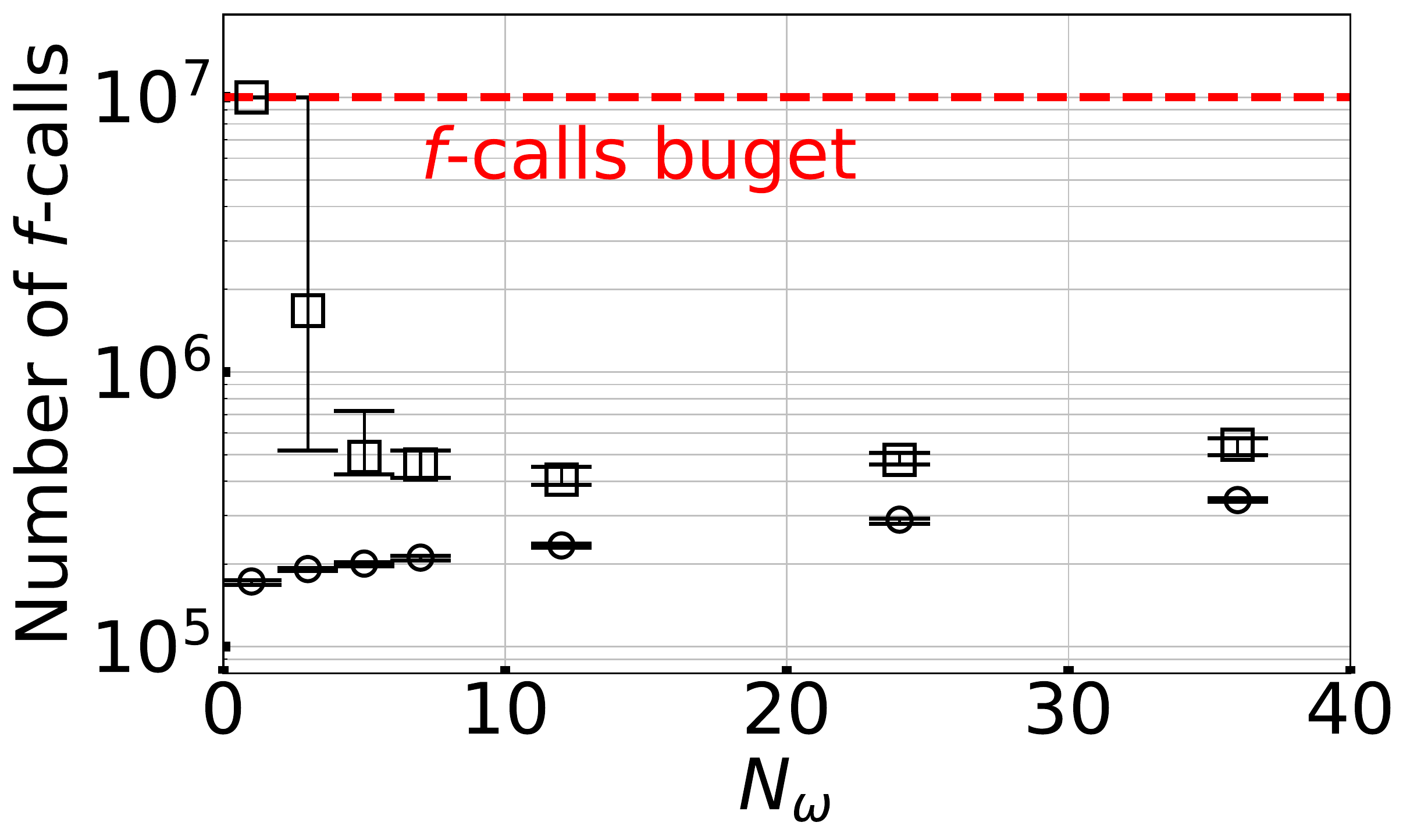}%
   \caption{$f_1$}%
\label{fig:ex2_f1}%
\end{subfigure}%
\caption{
Median and interquartile range of the number of $f$-calls
spent by \wracma{} and \wraslsqp{}  over $20$ trials with $\ny \in \{1, 3, 5, 7, 12, 24, 36\}$.
Note that the interquartile ranges were so small that the gaps between bars are barely visible in most cases.
}%
\end{figure}

Our approaches can solve $f_3$, $f_6$, and $f_8$ even with $\ny = 1$. 
\Cref{fig:ex2_f8} demonstrates the results of \wracma{} and \wraslsqp{} with $\ny=1$ for $f_8$.
The worst-case scenario for $f_3$, $f_6$, and $f_8$ is a singleton $\abs{\ywset(x^*)}=1$ and a constant around the global min--max solution $x^*$. Therefore, maintaining a single configuration ($\ny = 1$) was sufficient for the 
warm-starting
strategy in WRA to work efficiently on these problems. 
\Cref{fig:ex2_f8} shows \wracma{} and \wraslsqp{} with a smaller $\ny$ could converge to near global min--max solution with fewer f-calls. This may be because $f$-calls spent by the 
warm-starting
strategy are saved by setting smaller $\ny$. The reduction of $f$-calls by a small $\ny$ was not significant; therefore, we do not consider $\ny$ should be daringly small. 

\subsubsection{Category (W) ($f_1$, $f_2$)}

Maintaining multiple configurations, i.e., $\ny > 1$, is crucial for the proposed approach to successfully converge to the near global min--max solution $x^*$ for functions in Category (W) as we discussed in \Cref{sec:warmstart}. 
\Cref{fig:ex2_f1} shows the results of \wracma{} and \wraslsqp{} for $f_1$ with $\ny \in \{1,3,5,7, 12, 24, 36\}$. 
We confirmed that \wracma{} with a small $\ny$ failed to converge to $x^*$. Meanwhile, \wraslsqp{} could converge to $x^*$ with $\ny = 1$. 
This may be because AGA can rapidly maximize $f_1$ for $y$ from any starting point in $\mathbb{Y}$ and the 
warm-starting
strategy is unnecessary for \wraslsqp{} in $f_1$.

\subsubsection{Category (N) ($f_4$, $f_9$, and $f_{10}$)}

Multimodality in $y$, particularly with a weak global structure, seems to make it difficult to obtain the global min--max solution. 
As we see in \Cref{fig:ex2} for $f_4$, \wracma{} and \wraslsqp{} could not successfully converge.
The objective function $f(x,\cdot)$ for $f_4$ has $2^{d_y}$ local solutions and is a multimodal function with a weak structure. Such an objective function is difficult to efficiently optimize with any of the currently proposed algorithms \cite{hansen2009}. Therefore, we consider that the proposed approach failed to approximate the worst-case objective function values $\{F(x)\}_{i=1}^{\lambda_x}$ at many iterations; consequently, the outer CMA-ES could not converge to $x^*$.

Setting $\ny$ greater than the number of local maxima in $f(x, \cdot)$ is crucial to obtain successful convergence. 
As we see in \Cref{fig:ex2} for $f_9$, \wracma{} could successfully converge.
The objective function $f(x,\cdot)$ for $f_9$ has $8$ local maxima. When $\{y_k\}_{k=1}^{\ny}$ can include every local solution because of $\ny=36>8$, the solver explores the worst-case scenario using a good initial configuration in any case, i.e., the 
warm-starting
strategy works effectively.
Meanwhile, \wraslsqp{} could not converge to $x^*$ in most trials. AGA failed to even locally maximize $f$, possibly due to the ill-condition, more precisely, the Hessian matrix is not necessarily negative definite at some $x$. As a result of approximating the worst-case rankings $\mathrm{Rank}_F(\{x_i\}_{i=1}^{\lambda_x})$ in several iterations, the outer CMA-ES failed to converge to $x^*$. 
Further, for $f_4$, we confirmed the benefit of setting $\ny$ greater than the number of local maxima in $f(x, \cdot)$. For ${d_y}=5$, $\ny=36$ is greater than the number of local maxima, which is $2^5=32$.
\Cref{fig:ex2_f4} shows the experimental result from \wracma{} and \wraslsqp{} for $f_4$ with ${d_x}={d_y}=5$. As shown in \Cref{fig:ex2_f4}, \wraslsqp{} converged successfully and \wracma{} converged to a near-optimal solution.

\begin{figure}[t]
\centering%
\begin{subfigure}{0.33\hsize}%
\centering%
    \includegraphics[width=\hsize]{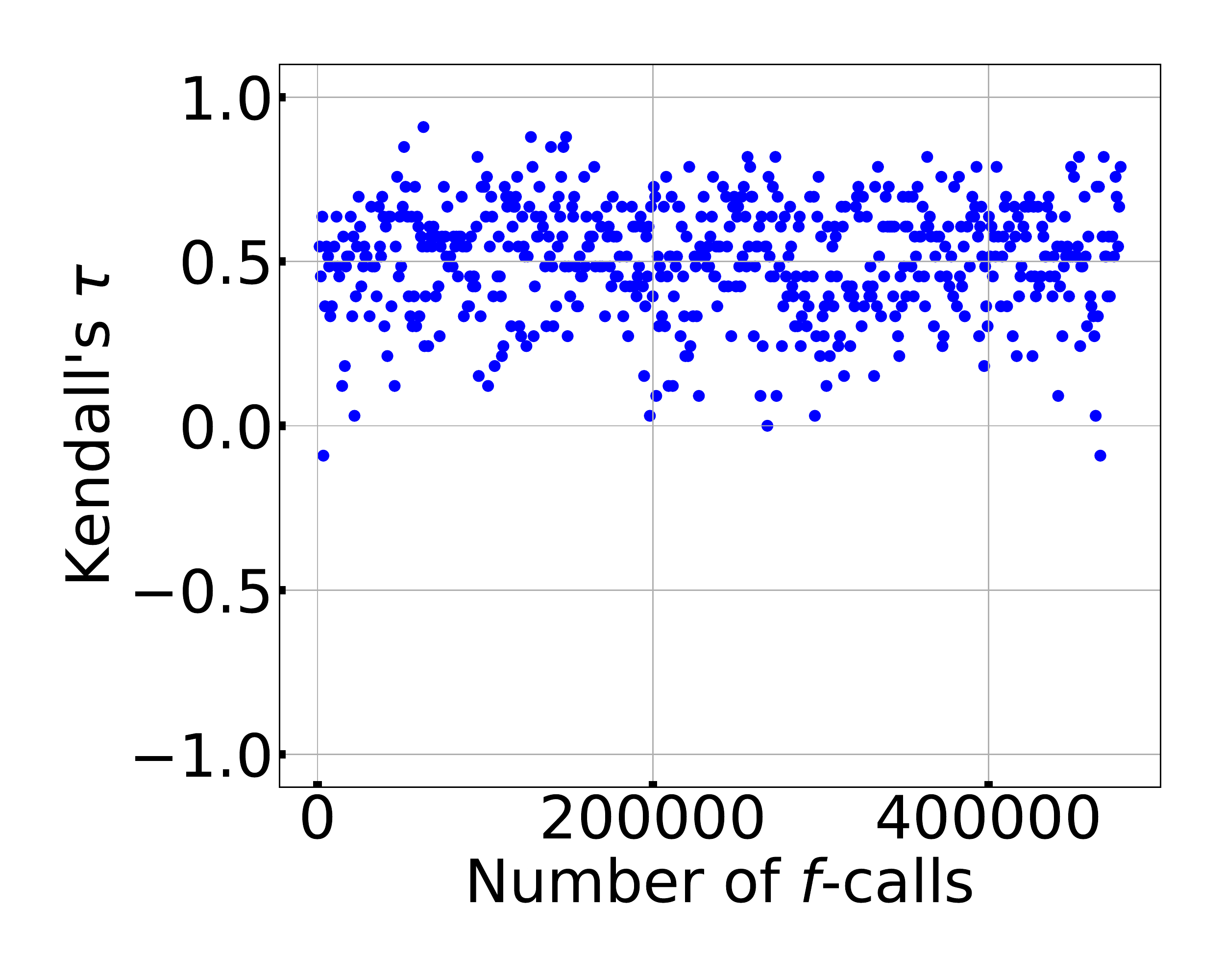}%
    \caption{$f_{1}$}%
\end{subfigure}%
\begin{subfigure}{0.33\hsize}%
\centering%
    \includegraphics[width=\hsize]{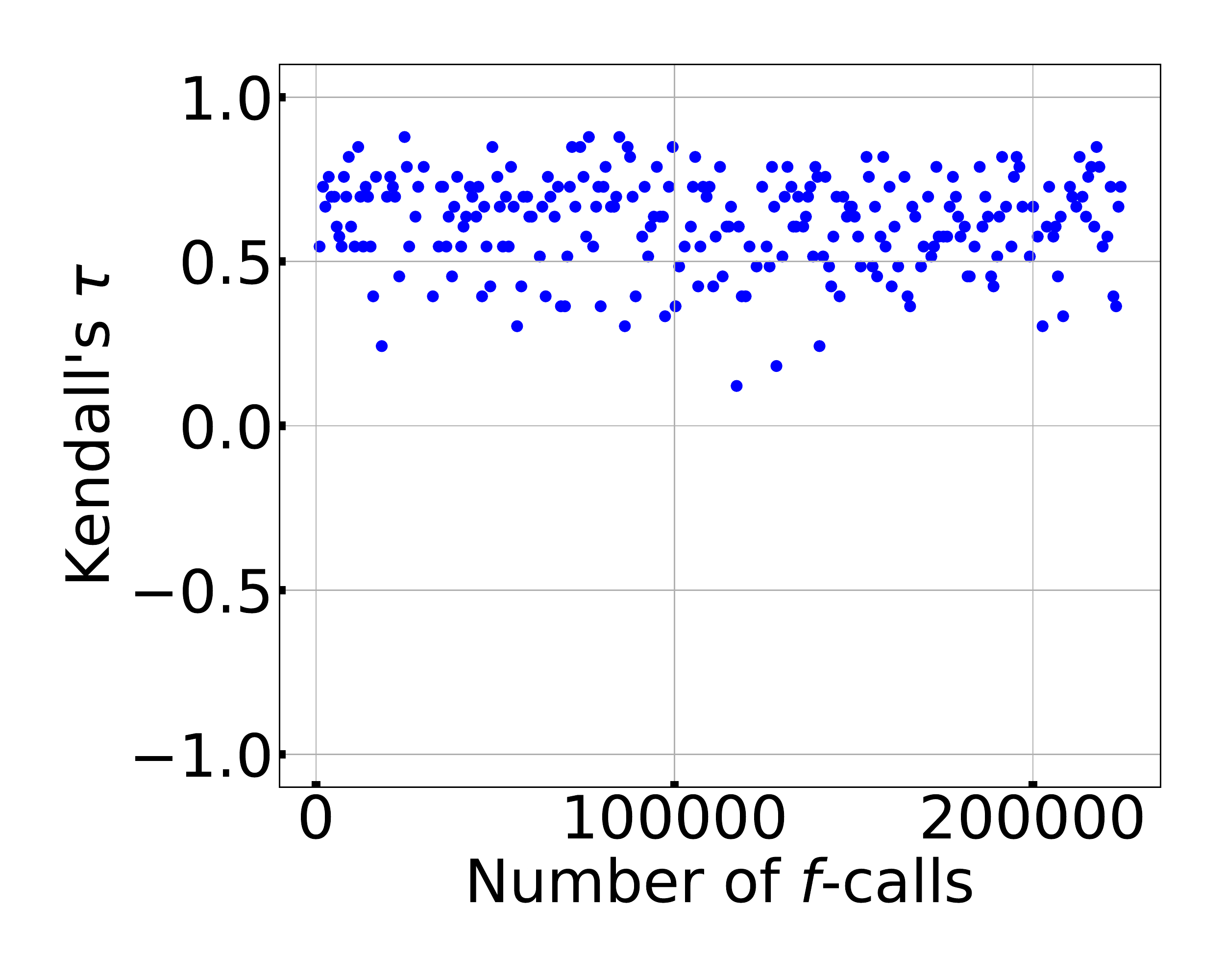}%
    \caption{$f_{5}$}%
\end{subfigure}%
\begin{subfigure}{0.33\hsize}%
\centering%
    \includegraphics[width=\hsize]{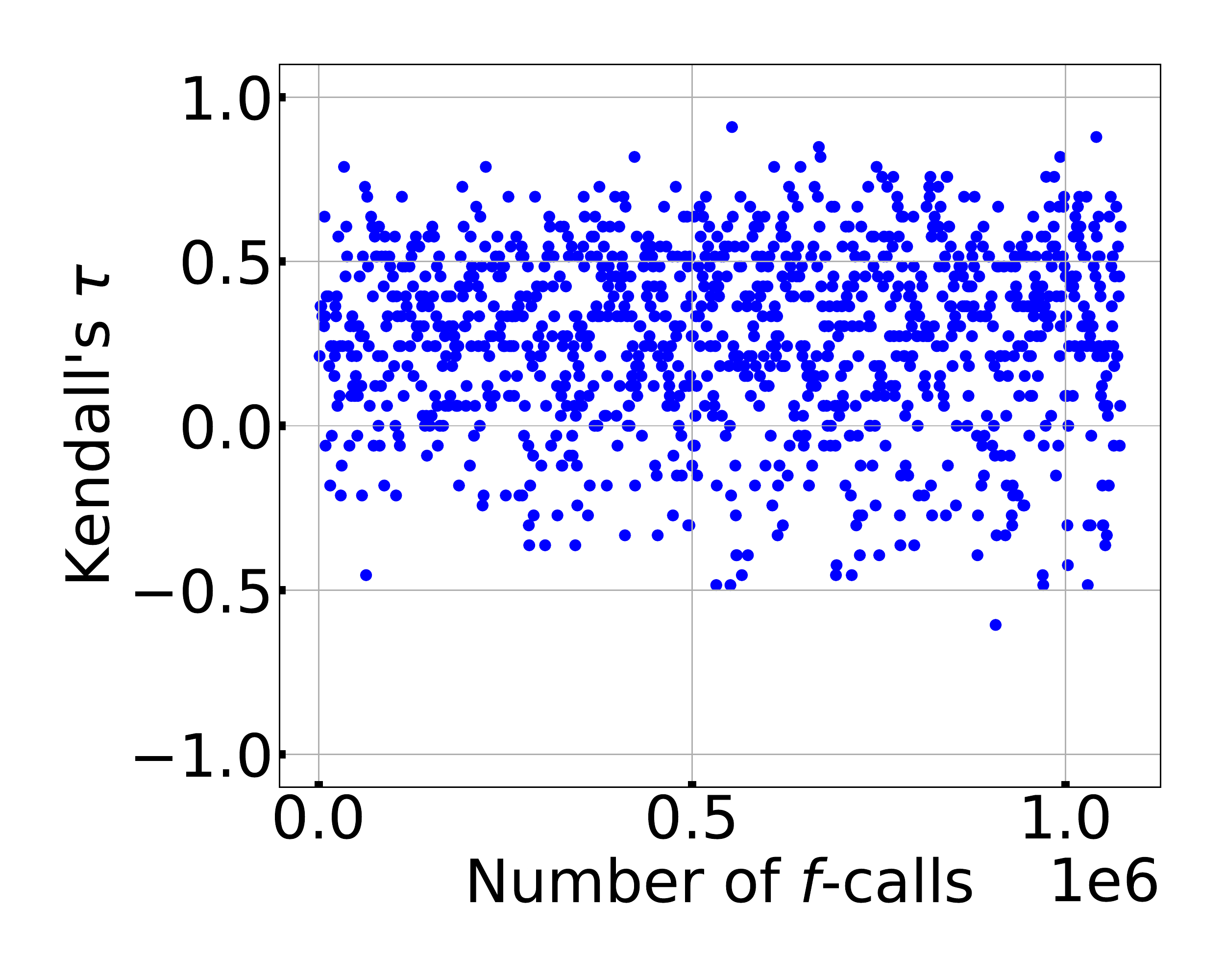}%
    \caption{$f_{10}$}%
\end{subfigure}%
\caption{Kendall's $\tau$ between the rankings of the worst-case objective function values and rankings obtained using the WRA mechanism at each iteration in a typical run of \wracma{}. In $f_{10}$, \wracma{} was terminated due to $\Cond(\Sigma_x) > \Cond_{\max}^x$ before $f$-calls reached $10^7$.}
\label{fig:ex2_kendall}
\end{figure}

\begin{figure}[t]
\centering%
\begin{subfigure}{0.33\hsize}%
\centering%
    \includegraphics[width=\hsize]{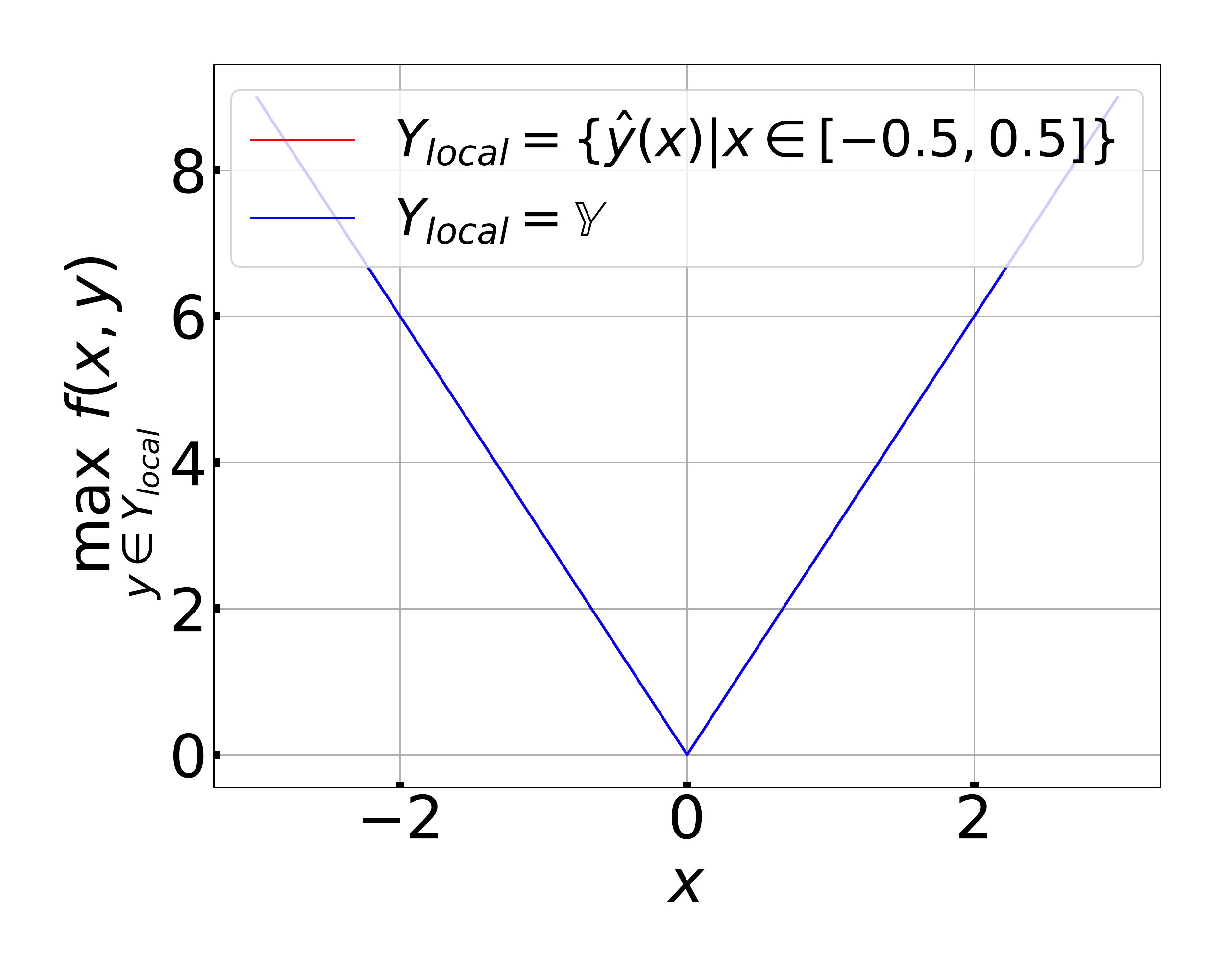}%
    \caption{$f_{1}$}%
\end{subfigure}%
\begin{subfigure}{0.33\hsize}%
\centering%
    \includegraphics[width=\hsize]{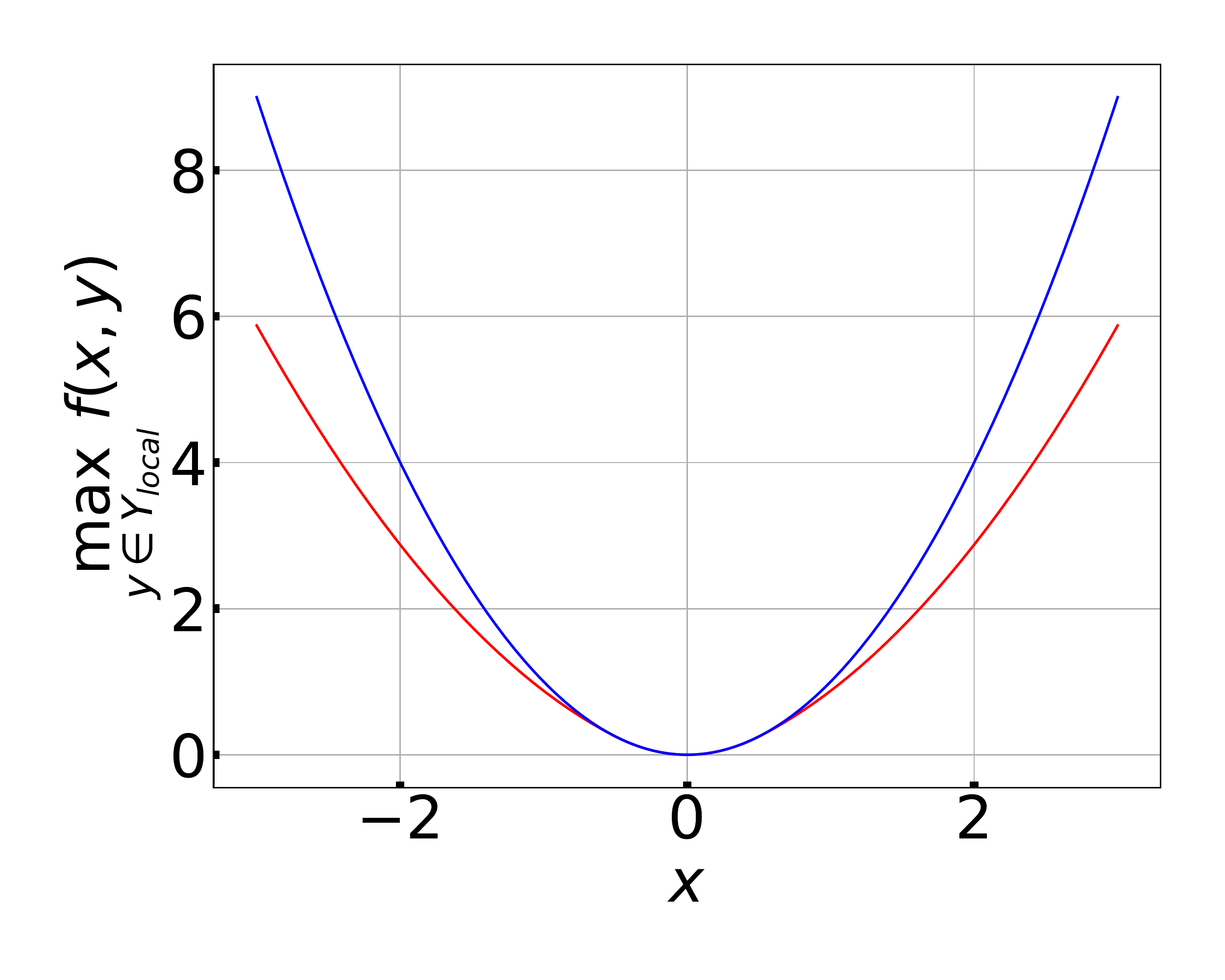}%
    \caption{$f_{5}$}%
\end{subfigure}%
\begin{subfigure}{0.33\hsize}%
\centering%
    \includegraphics[width=\hsize]{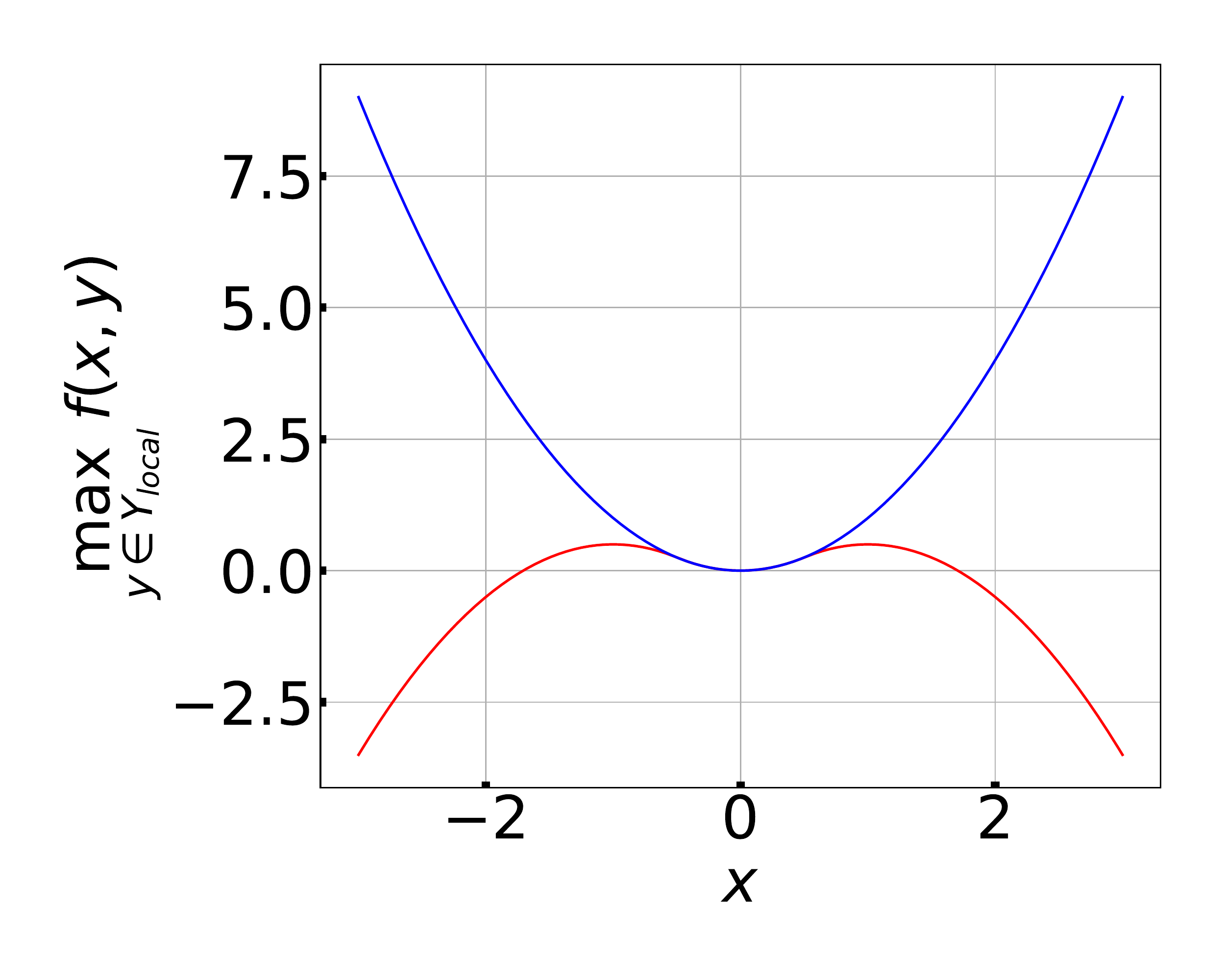}%
    \caption{$f_{10}$}%
\end{subfigure}%
\caption{Landscape of $\max_{y \in Y_{local}} f(x,y)$ with $Y_{local} = \mathbb{Y}$ and $Y_{local} = \{\yw(x) | x \in [-0.5, 0.5]\}$ on $f_1$, $f_5$, and $f_{10}$ with ${d_x} = {d_y} = 1$.}
\label{fig:ex2_landscape}
\end{figure}

\begin{figure}[t]
\centering
\includegraphics[width=0.5\hsize]{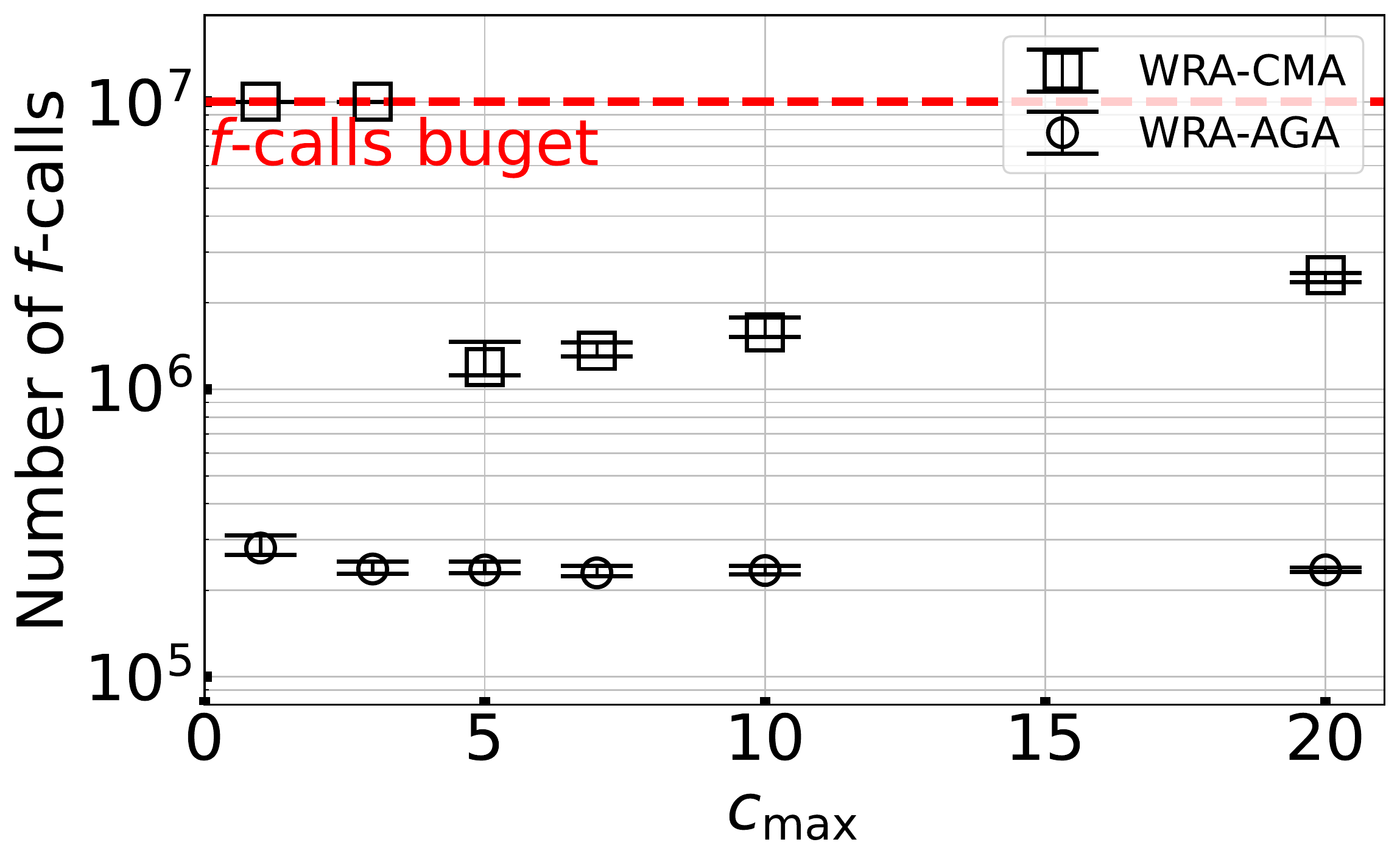}%
\caption{
Median and interquartile range of the number of $f$-calls
spent by \wracma{} and \wraslsqp{}  over $20$ trials on $f_{10}$ with $c_{\max} \in \{1, 3, 5, 7, 10, 20\}$.
Note that the interquartile ranges were so small that the gaps between bars are barely visible in most cases.
}%
\label{fig:ex2_f10}
\end{figure}

The relevance of $c_{\max}$ is pronounced in the results of \wracma{} for $f_{10}$.
\Cref{fig:f10} shows that \wracma{} with $c_{\max} = 1$ failed to converge for $f_{10}$. 
From \Cref{fig:ex2_kendall}, the Kendall's $\tau$ for $f_{10}$ is more frequently negative than those for $f_1$ and $f_5$ in which \wracma{} could converge successfully.
The reason for the low $\tau$ value is explained using \Cref{fig:ex2_landscape}, which visualizes the landscape of an approximated worst-case objective function $\max_{y \in Y_{local}} f(x,y)$ with $Y_{local} = \mathbb{Y}$ (i.e., the ground truth worst-case objective) and $Y_{local} = \{\yw(x) | x \in [-0.5, 0.5] \}$. 
\Cref{fig:ex2_landscape}  simulates the situation where the search distribution for $x$ is concentrated around $[-0.5, 0.5]$ and hence $\ny$ worst-case scenario candidates in WRA are concentrated at the corresponding worst-case scenario region. 
Differently from $f_1$ and $f_5$, the worst-case objective function values of candidate solutions outside $[-0.5, 0.5]$, which are generated by chance, are significantly underestimated for $f_{10}$. 
In such a situation, the worst-case scenario search with a small $c_{\max}$ may be insufficient to correctly rank such solutions and they may be regarded as the best solutions.
This will prevent convergence to the global min--max solution.
From \Cref{fig:ex2_f10}, the performance of \wracma{} is improved by setting a greater $c_{\max}$. However, a too large $c_{\max}$ value requires more $f$-calls.
\wraslsqp{} could converge successfully for $f_{10}$ even with $c_{\max}=1$. This could be because the objective function for $y$ in $f_{10}$ was relatively easy for \Cref{alg:SLSQPinwra}; therefore, the worst-case scenario could be approximated with high accuracy even for small $c_{\max}$. Differently with the result from \wracma{}, the number of $f$-calls spent by \wraslsqp{} was less sensitive on various $c_{\max}$. We confirmed that most of AGA in this experiments was terminated by $U_{\min}$ before the number of improvements reached to $c_{\max}$.

\section{Application to robust berthing control} \label{sec:berthing}

We confirm the effectiveness of \wracma{}, \wraslsqp{}, \wracmaadv{}, and \wraslsqpadv{} on a robust berthing control problem presented in \cite{akimoto2022berthing}. 

\subsection{Problem description}

We exactly follow the problem setup in \cite{akimoto2022berthing}. 
We briefly describe the problem. 
The objective of this problem is to obtain a controller to control a ship to a target state located near a berth while avoiding collision with the berth. 
The ship’s state is represented by $s \in \R^6$, and the control signal is represented by $a \in U \subset \R^4$.
The state equation is the maneuvering modelling group model used in \cite{Miyauchi2022}.
The feedback controller $u_x: \R^6 \to U$ is modeled by a neural network with ${d_x} = 99$ parameters. 
The domain of the network parameters is set to $\mathbb{X} = [-1, 1]^{d_x}$. 

We consider three cases of uncertainty sets. 
Case A: The wind condition is the uncertainty vector. 
The wind condition is parameterized by the wind direction in $[-\pi, \pi]$ [rad] and wind velocity in $[0, 0.5]$ [m/s]. 
Case B: The coefficient in the state equation regarding the wind force is the uncertain vector, which comprises a $10$-dimensional vector. 
Case C: Both uncertainties in Cases A and B.
In all cases, the search domain is scaled to $\mathbb{Y} = [-1, 1]^{d_y}$.

The objective function $f(x, y)$ comprises two components. The first component measures the difference between the ship’s final state and target state. The second component measures the penalty for a collision with the berth. If a ship collides with the berth during the control period comprising $200$ [s], it receives a penalty greater than $10$. Our objective is to minimize the worst-case objective function $\max_{y \in \mathbb{Y}} f(x, y)$, where the uncertainty set $\mathbb{Y}$ differs for Cases A, B, and C.

\subsection{Experimental settings}

The proposed approach and the existing approaches, \zopgda{} and \acma{}, were applied to the robust berthing control problem. 
For each problem, we run 20 independent trials with different random seeds. 
The maximum number of $f$-calls was set to $2 \times 10^6$.


All approaches were configured as in \Cref{sec:test}, where $u_x = 1$, $\ell_x = -1$, and $b_y = 1$ were plugged, except that we turned on the restart strategy of the proposed approach and \acma{} 
to tackle multimodality as it has been used in the previous study \cite{akimoto2022berthing},
and the diagonal acceleration in CMA-ES for the outer minimization in \wracma{} and \wraslsqp{}.
%
For fair comparison, we have implemented a simple restart strategy for 
\zopgda{}. The initial solution and the initial scenario vector are reset uniform randomly in the given domains when $\norm{\eta_x \widehat{\nabla_{x} f}(x^t, y^t)}_2^2 + \norm{\eta_y \widehat{\nabla_{y} f}(x^t, y^t)}_2^2 \leq 10^{-5}$ is satisfied, i.e., significant improvements of the solution candidate and the scenario vector are not expected.
We set the number of configurations as $\ny=34$ ($=2 \times \lambda_x$).
The termination thresholds were $V_{\min}^x=10^{-6}$ and $\Cond_{\max}^x=10^{14}$.
In addition, we terminated the proposed approach if the best worst-case objective function value were not significantly improved.
Precisely, we save $\tilde{F}_{\min}^t = \min_{i=1,\dots,\lambda}(\tilde{F}(x_i))$, where $\tilde{F}(x_i)$ is the approximated worst-case objective function value of $x_i$ computed in WRA, and terminate if $\max_{T=t-10,... ,t}\{\tilde{F}_{\min}^t\}-\min_{T=t-10,... ,t}\{\tilde{F}_{\min}^t\} < 0.01$ is satisfied.\footnote{
We often observe that (1+1)-CMA-ES converges significantly faster than the standard CMA-ES (non-elitism CMA-ES) when optimizing a neural network. 
Probably because of this effect, \acma{} could perform several restarts on this problem, whereas the proposed approach could not perform any restart without this termination condition. For the proposed approach to perform multiple restarts, we introduced the termination criterion at the risk of too early termination. As a result, we confirmed that the proposed approach performed restarts $1$ or $2$ times in each run for Case A and Case C, and $2--4$ times in each run for Case B.
}

For each trial, the obtained solution was evaluated on the worst-case objective function value as in the previous study \cite{akimoto2022berthing}. 
To estimate the worst-case objective function value for each solution, we ran the (1+1)-CMA-ES to approximate $\max_{y \in \mathbb{Y}} f(x, y)$ with $100$ different initial points. Then, by taking the maximum of the obtained worst-case scenario candidates, $y_1, \dots, y_{100}$, the worst-case objective function value is evaluated. For the configuration of the (1+1)-CMA-ES, we exactly followed the previous study \cite{akimoto2022berthing}.

\begin{figure}[t]
  \centering
  \begin{subfigure}{0.9\hsize}
  \centering
    \includegraphics[height=\hsize, angle=-90, clip, trim=5 10 10 0]{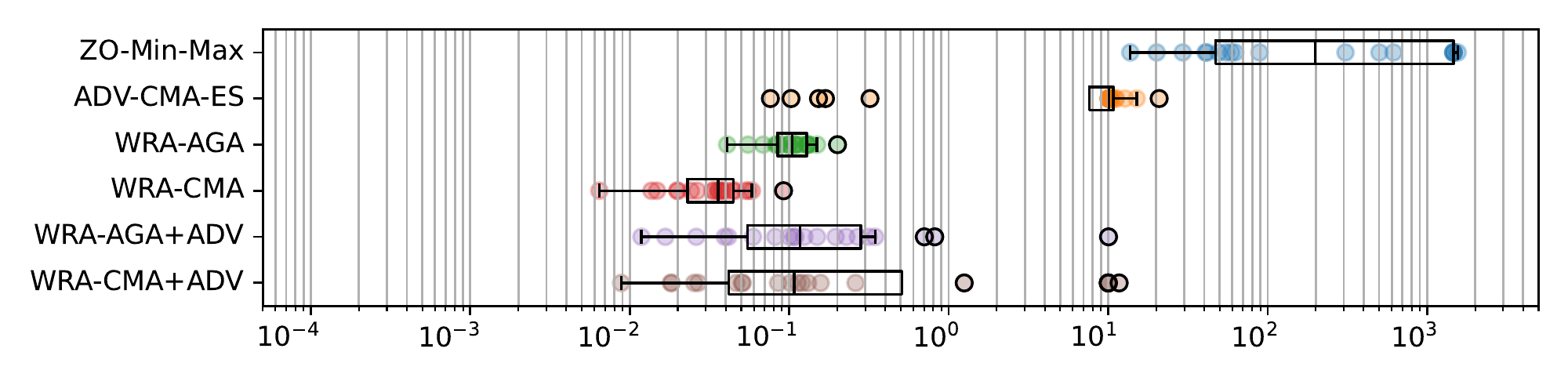}%
    \caption{A}
    \end{subfigure}
    \\
    \begin{subfigure}{0.9\hsize}
    \centering
    \includegraphics[height=\hsize, angle=-90, clip, trim=5 10 10 0]{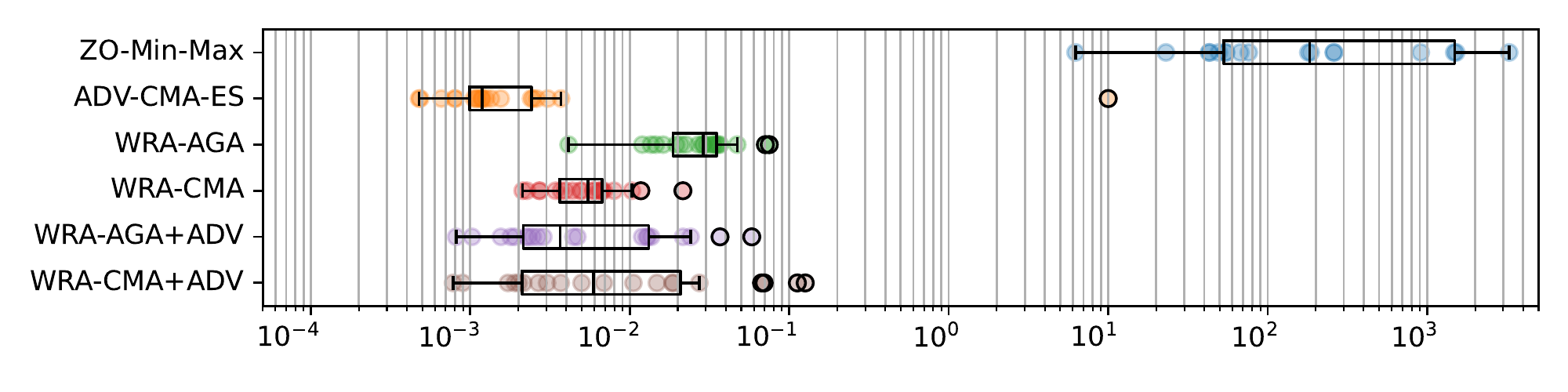}%
    \caption{B}
    \end{subfigure}    
    \\
    \begin{subfigure}{0.9\hsize}
    \centering
    \includegraphics[height=\hsize, angle=-90, clip, trim=5 10 10 0]{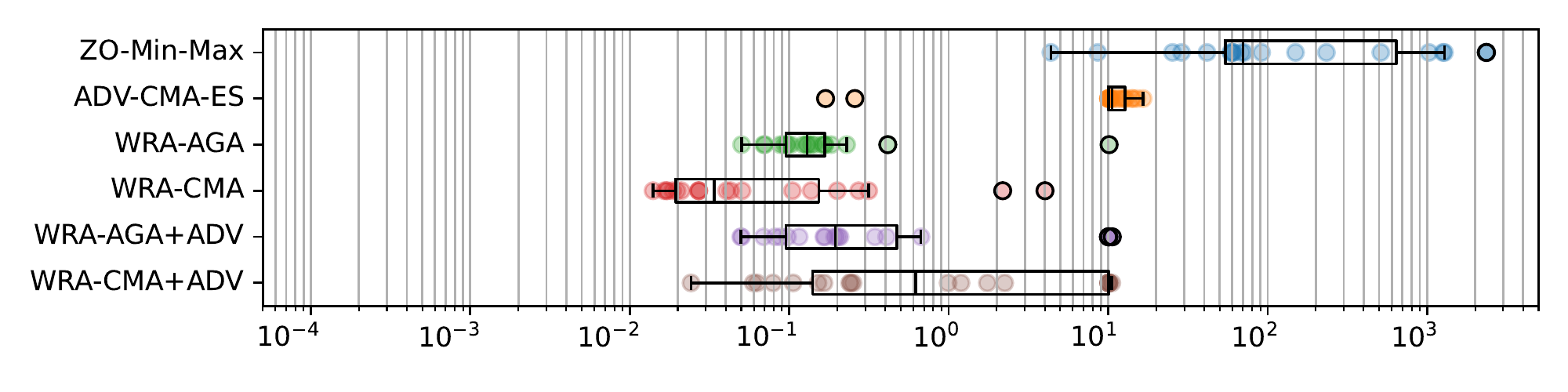}%
    \caption{C}
    \end{subfigure}
  \caption{Performance of the solution obtained in 20 independent trials of \zopgda{}, \acma{}, \wraslsqp{}, \wracmaadv{}, and \wraslsqpadv{} for Cases A, B, and C. Side edge on each box indicates the lower quartile Q1 and upper quartile Q3, and middle line in each box indicates the median. The lower and upper whiskers are the lowest datum above Q1-1.5(Q3-Q1) and the highest datum below Q3+1.5(Q3+Q1).}
\label{fig:berthing_result}
\end{figure}

\begin{figure*}
    \centering
    \begin{subfigure}{0.48\hsize}%
    \includegraphics[width=\hsize]{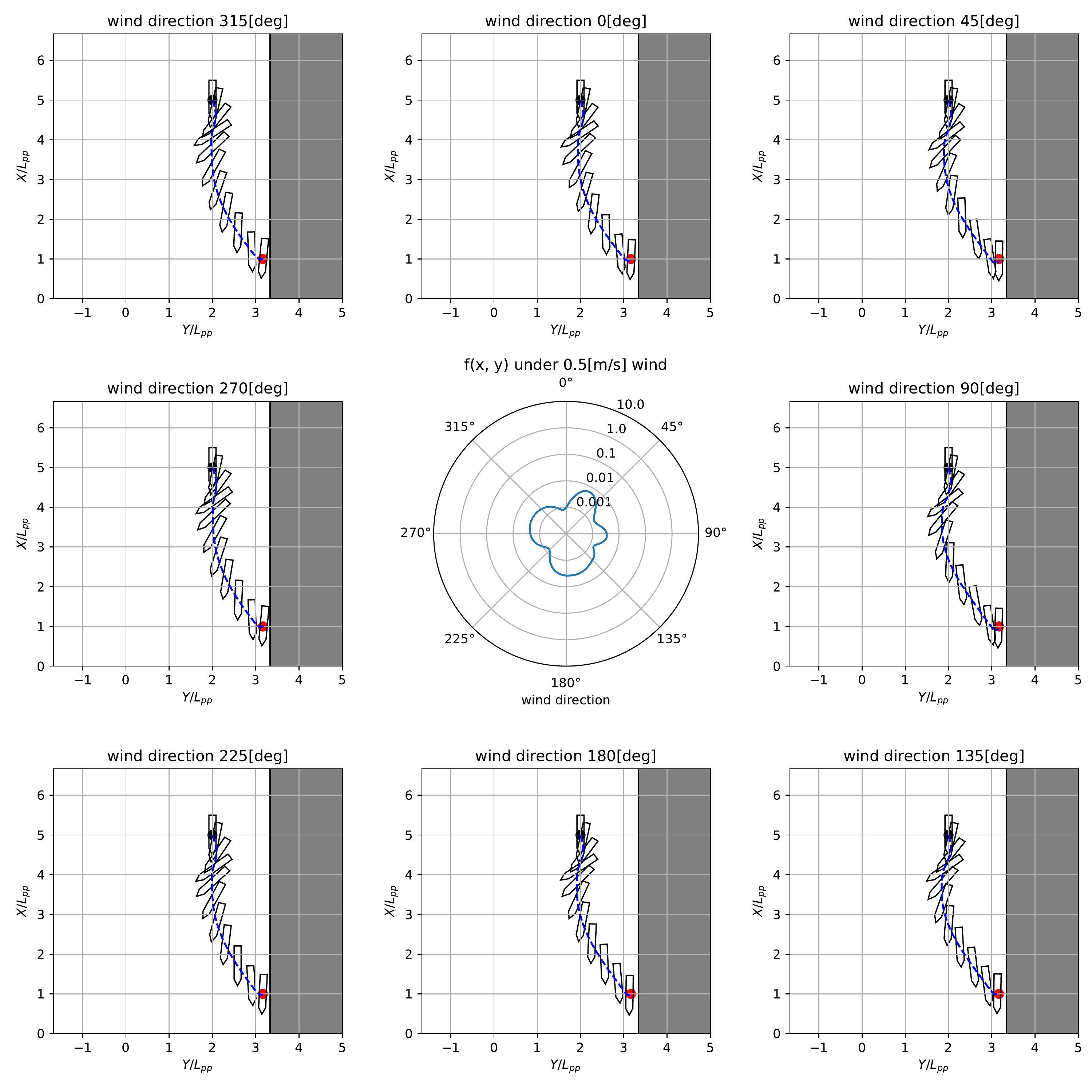}%
    \caption{\wracma{}}%
    \end{subfigure}%
    \hfill
    \begin{subfigure}{0.48\hsize}%
    \includegraphics[width=\hsize]{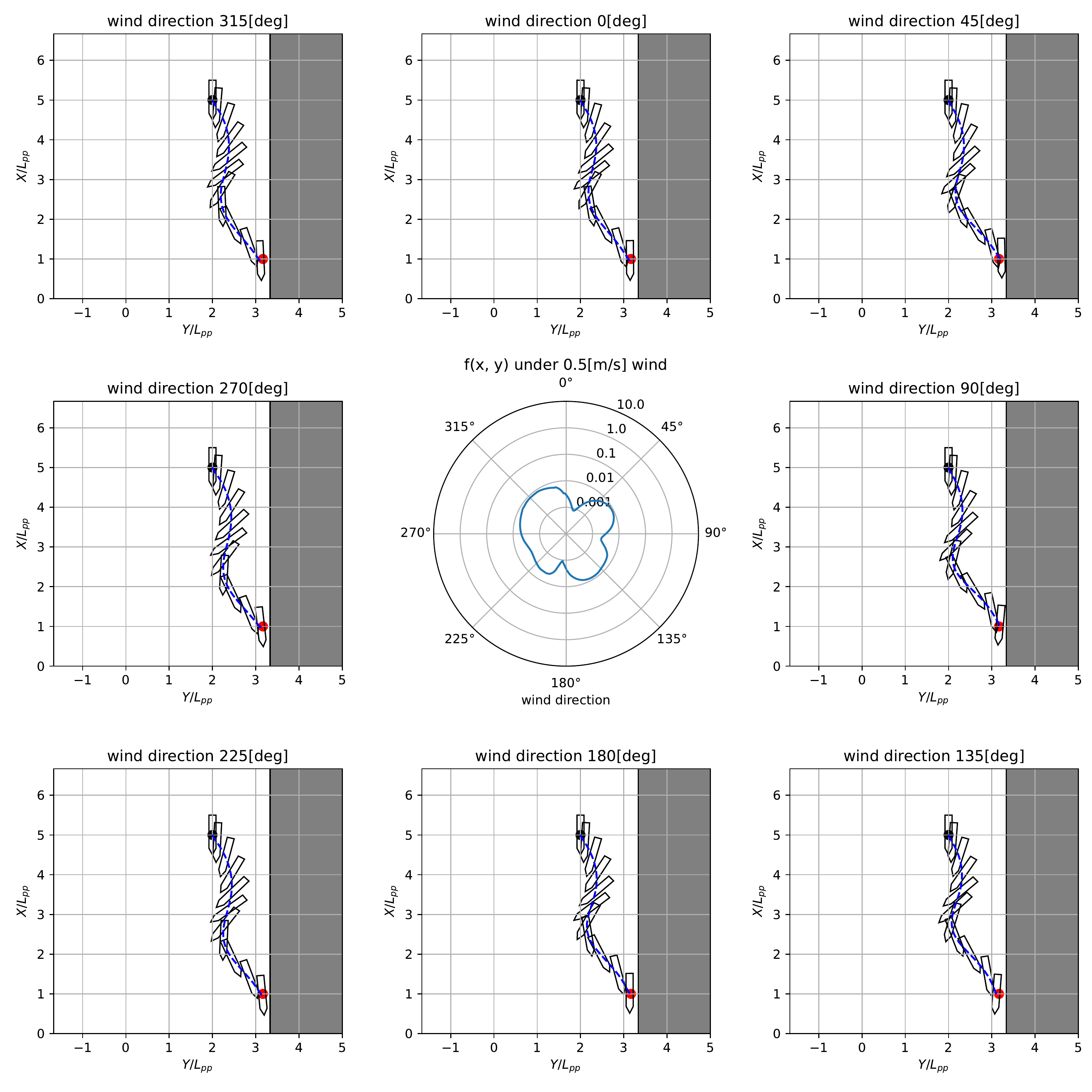}%
    \caption{\wracmaadv{}}%
    \end{subfigure}%
    \caption{Visualization of the trajectories obtained by the controllers for the worst wind condition with the maximum wind velocity of 0.5 [m/s]. The best controller obtained by (a) \wracma{} and (b) \wracmaadv{} are displayed. See the caption of \Cref{fig:berthing-trajectory-adv} for details of the figures.}
    \label{fig:berthing-trajectory-wra}
\end{figure*}

\subsection{Result and evaluation}

The worst-case performances of the obtained solutions are summarized in \Cref{fig:berthing_result}.

\emph{Results of \acma{} and \zopgda}.
\acma{} could find robust solutions in all but one trial in Case B. Meanwhile, the medians of the worst-case performances in Cases A and C were greater than $10$, indicating collision with the berth. 
These results agree with those of a previous study \cite{akimoto2022berthing}.
\zopgda{} failed to obtain solutions that could avoid collision with the berth in the worst-case scenario in most trials in all cases.

\emph{Results of \wracma{} and \wraslsqp}.
Except for a trial of \wraslsqp{} in Case C, \wracma{} and \wraslsqp{} could find controllers that could avoid collision with the berth in the worst-case scenarios. 
As discussed in \Cref{sec:intro}, we hypothesize that the problems in Cases A and C are such that the worst-case scenario around the optimal controller for $F$ changes discontinuously, and they are difficult for \acma{} and \zopgda{}. We confirmed \acma{} restarted significantly more often than \wracma{} and \wraslsqp{}, however, superior solutions were obtained by the proposed approaches.
We consider this is one of the reasons for the superior worst-case performances of \wracma{} and \wraslsqp{} in Cases A and C.
Moreover, the worst-case performances of \wracma{} and \wraslsqp{} were significantly worse than that of \acma{} in Case B. 
One reason for this result is the termination criterion introduced in the experiment, which prevents performing an intensive local search. In addition, we confirmed that the worst-case performances of \wracma{} and \wraslsqp{} were inferior to that of \acma{}, even without this termination condition, attributable to the slower convergence of \cmaes{} than (1+1)-CMA-ES for this problem.

\emph{Results of \wracmaadv{} and \wraslsqpadv}.
In Case B, \wracmaadv{} and \wraslsqpadv{} could obtain better worst-case performances in several trials. From these results, we confirm that the motivation of running \acma{} after \wracma{} and \wraslsqp{}, namely, improving the exploitation ability, was realized in Case B. 
The worst-case performances exhibited more variance in Case A, and their median was significantly degraded in Case C. 
The negative effect of running \acma{} after \wracma{} and \wraslsqp{} may be explained as follows.
The set $\bar{Y}$ of worst-case scenario candidates given to \acma{} is expected to approximate the worst-case scenario set $\ywset(\bar{x})$ of a given solution candidate $\bar{x}$ as a subset of $\bar{Y}$. 
During \acma{}, $\bar{Y}$ is fixed and the solution candidate $x$ is optimized under $\bar{Y}$ and a newly added scenario candidate $y_\mathrm{adv}$. 
Because $\ywset(x)$ may change with $x$, $\bar{Y}$ may not approximate $\ywset(x)$ well after \acma{} and a single scenario candidate $y_\mathrm{adv}$ may not be sufficient to recover $\ywset(x)$. 
Thus, the solution obtained by $\acma{}$ may be overfitting to $\bar{Y}$ and there may be scenarios where the performance is worse.

\Cref{fig:berthing-trajectory-wra} shows the ship trajectory observed under the best controllers obtained by \wracma{} and \wracmaadv{}. 
Under both controllers, collision is successfully avoided under wind from an arbitrary direction.

\section{Conclusion}

To address the limitation of existing approaches for black-box min--max optimization problems, \zopgda{} and \acma{}, we propose a novel approach to minimize the worst-case objective function using the CMA-ES while approximating the rankings of the worst-case objective function values of the solution candidates using a proposed WRA mechanism. 
To save $f$-calls inside the WRA mechanism, we implement a 
warm-starting
strategy and an early-stopping strategy. We developed \texttt{WRA-CMA} and \texttt{WRA-AGA} by combining the WRA mechanism with the CMA-ES and AGA, respectively. 
A restart strategy and a hybridization of the proposed approach and \acma{} are implemented for practical use. The proposed approach was evaluated for $11$ test problems and three cases of the robust berthing control problem.

The relevant findings from our numerical experiments are as follows.
On smooth 
strongly
convex--concave problems, where \zopgda{} and \acma{} have been analyzed for their convergence, the proposed approach exhibited slower convergence than existing approaches when the interaction between $x$ and $y$ is relatively weak. However, the $f$-calls were not increased significantly for the proposed approach when the interaction was stronger, whereas they were increased significantly for the existing approaches.
On nonsmooth strictly convex--concave problems and problems where the global min--max solution is not a strict global min--max saddle point, \zopgda{} and \acma{} failed to converge, whereas the proposed approach converged. For the former problems, the proposed approach could locate the global min--max solution even with $\ny = 1$; a sufficiently large $\ny$ was a key to the success of the proposed approach. 
When good initial configurations were not provided for some solution candidates, a greater $c_{\max}$ was helpful.

The effectiveness of \texttt{WRA-CMA} and \texttt{WRA-AGA} were demonstrated in three cases of the robust berthing control problem. For problems where wind direction was included in $y$, \texttt{WRA-CMA} and \texttt{WRA-AGA} could find controllers that avoid collision with the berth in the worst-case scenario, whereas controllers obtained by the existing approaches, \acma{} and \zopgda{}, often collided with the berth in the worst-case scenario. 
For problems where the wind direction is included in $y$, the worst-case scenario is expected to change discontinuously around the optimal controller for $F$, and they are difficult for the existing approaches. Moreover, the proposed approach can address such a difficulty. Therefore, we consider that controllers obtained using the proposed approach  were superior to those obtained using the existing approaches.
For a problem where the existing approaches obtained controllers that avoid a collision, we confirmed that the existing approaches found a better solution than the solutions obtained using the proposed approach. In addition, for such a problem, several trials showed that better controllers were obtained by running \acma{} after \texttt{WRA-CMA} and \texttt{WRA-AGA}.

Besides the above advantages of the proposed approach, one practical advantage of the proposed approach over \zopgda{} and \acma{} is that it is parallel-implementation friendly.
In WRA, $\lambda_x$ solvers $\mathcal{M}(\omega_k)$ ($k = 1,\dots,\lambda_x$) can be run in parallel.
The $\lambda_x \ny$ evaluations of $f(x_i, y_k)$ at the beginning of WRA can be performed in parallel.
Moreover, if \texttt{WRA-CMA} is used, $\lambda_y$ $f$-calls at each iteration of $\mathcal{M}(\omega_k)$ can be performed in parallel.
In total, roughly $\lambda_x \lambda_y$ times speedup in terms of the wall clock time can be achieved ideally.
For example, in Case C of the robust berthing control problem, we have ${d_x} = 99$ and ${d_y} = 12$; hence, $\lambda_x = 17$ and $\lambda_y = 11$, resulting in a possible speedup of factor $187$. 
Each $f$ evaluation took about $0.1$ s on average, amounting to about 2.3 days for each trial.
If the ideal speedup is achieved, the wall clock time reduces to about 18 min.
This compensates for the disadvantage of the proposed approach over \acma{}: slower convergence.

The main limitation of this study is the lack of theoretical guarantees. 
For the WRA mechanism to work effectively, we assume that $\yw(x)$ is continuous almost everywhere and $\max_{y \in \mathbb{Y}}f(x, y)$ can be solved efficiently for each $x$. 
However, questions as to how much $\yw$ can be sensitive or how efficiently the inner solver $\mathcal{M}$ should solve $\max_{y \in \mathbb{Y}}f(x, y)$ are not answered formally in this study. 
Such a theoretical investigation provides not only a guarantee of the performance of the proposed approach but also a seed to improve it.
Therefore, theoretical investigations of the WRA mechanism are important future research directions.

The black-box min--max optimization lacks the de facto standard benchmarking testbed, covering problems with different characteristics. 
In this study, we design $11$ test problems from the perspective of the characteristics of the global min--max solution (whether it is a strict min--max saddle point, a weak min--max saddle point, or not a min--max saddle point), and the perspective of the smoothness and the strong convexity of $f$. 
Moreover, we limit our focus on problems where $f(x, y)$ has relatively simple characteristics with respect to $x$ and $y$ and difficulties in black-box optimization, such as ruggedness, non-separability, and ill-conditioning, are yet to be considered. For example, all test problems are convex in $x$ except for $f_{10}$ and the effect of the multimodality in $x$ is not considered. The investigation of the effect of ill-conditioning of $f$ is limited to the comparison between $f_5$ and $f_{11}$.
Because approaches for black-box min--max optimization are designed and improved based on benchmarking and theoretical analyses on black-box min--max optimization are rather limited, developing 
benchmarking test cases is highly desired.
This is also an important future research direction.

\begin{acks}
This paper is partially supported by JSPS KAKENHI Grant Number 19H04179 and 22H01701.
\end{acks}

\bibliographystyle{ACM-Reference-Format}
\bibliography{thebibliography}

\appendix
\section{
Sensitivity analysis}\label{app:wrasensitivity}

The sensitivities of \wracma{} and \wraslsqp{} on their hyper-parameters $\tau_{\mathrm{threshold}}$, $\bar{p}_{+}$, and $\bar{p}_{-}$ are investigated on test problems. 
The experimental settings are the same as those described in \Cref{sec:common}. 
Among test problems, the proposed approaches under the setting in \Cref{sec:common} can converge to the optimal solution $x^*$ in $f_1$--$f_3$ and $f_5$--$f_8$. Test function $f_1$ has similar characteristic with $f_2$ in terms of that the worst-case scenarios around $x^*$ are discontinuously distributed, and $f_5$ has similar characteristic with $f_3$ and $f_5$--$f_8$ in which the worst-case scenarios around $x^*$ are continuously distributed. 
Therefore, we use $f_1$ and $f_5$ for this sensitivity analysis.
In this sensitivity analysis, the dimensions are set to ${d_x} = {d_y} = 20$, and the coefficient matrix of the interaction term $x^T B y$ is $B = \diag{(1, \dots, 1)}$.

\begin{figure}[t]
\centering
\begin{subfigure}{0.48\hsize}%
    \includegraphics[width=\hsize]{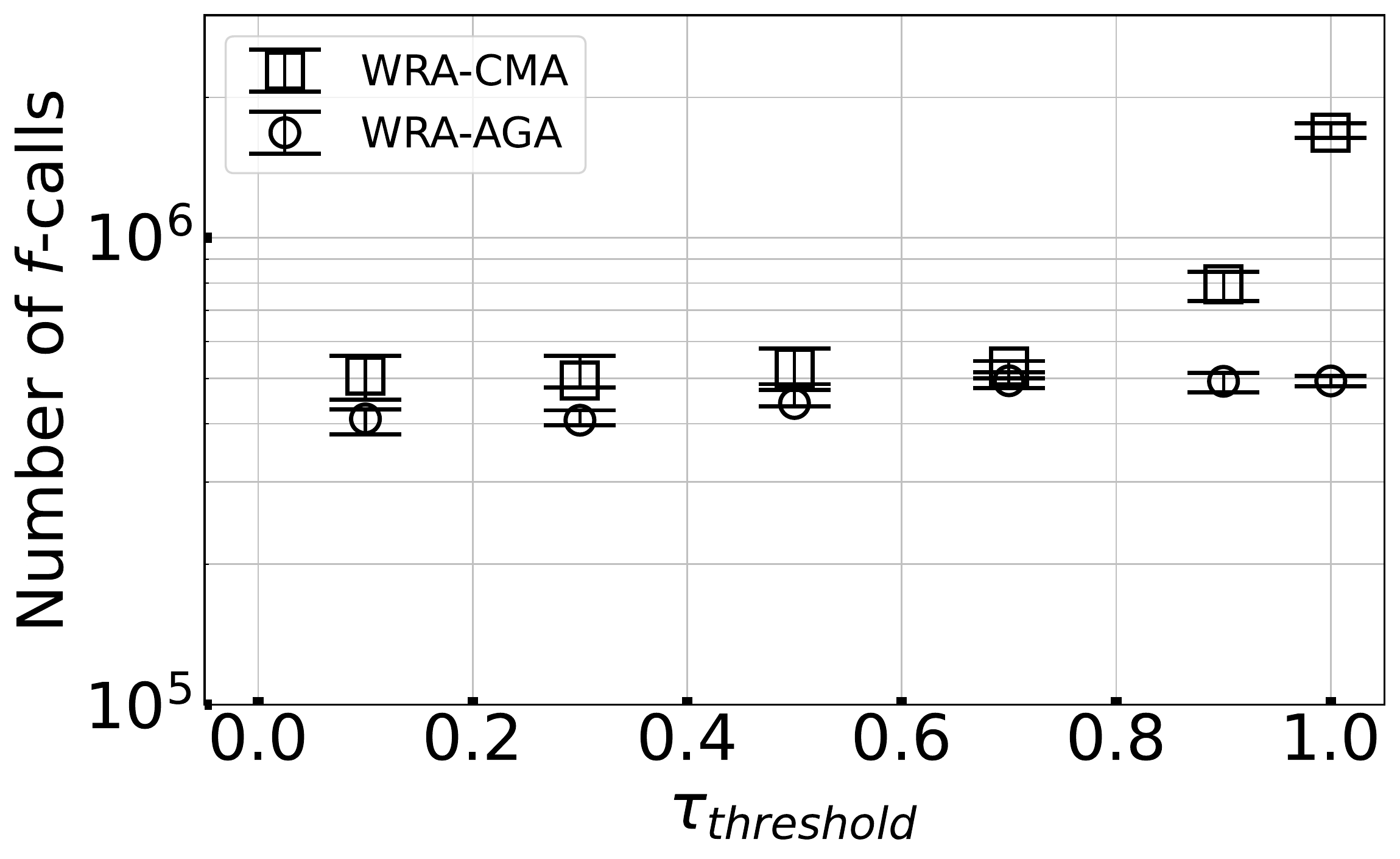}%
  \caption{$f_1$}%
\end{subfigure}%
\hfill
\begin{subfigure}{0.48\hsize}%
    \includegraphics[width=\hsize]{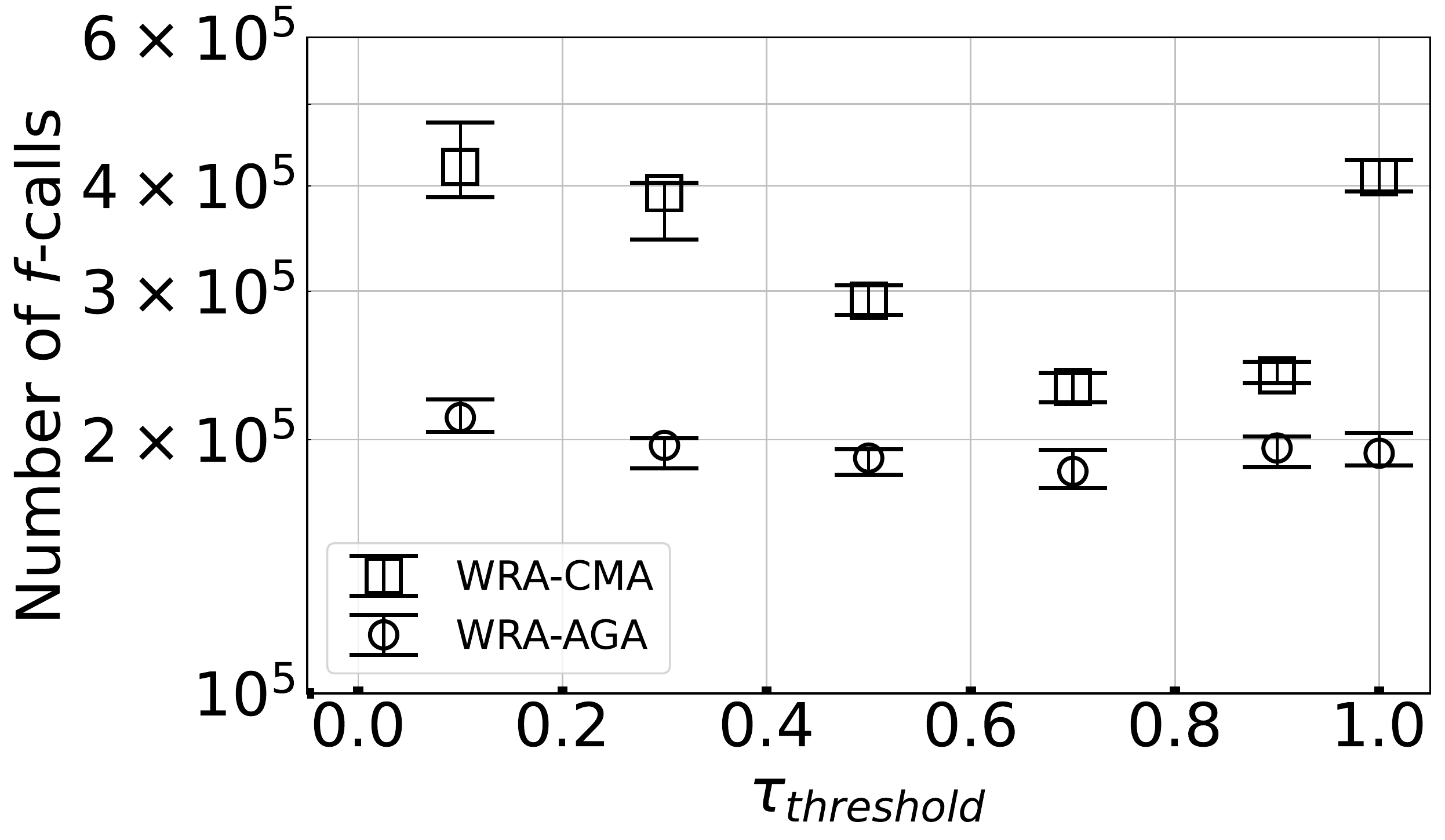}%
  \caption{$f_5$}%
\end{subfigure}%
\caption{Median and interquartile range of the number of $f$-calls over 20 trials obtained from \wracma{} and \wraslsqp{} with $\tau_{\mathrm{threshold}} \in \{0.1, 0.3, 0.5, 0.7, 0.9, 1.0\}$.
}%
\label{fig:p2tau}
\end{figure}

\subsection{Sensitivity to $\tau_{\mathrm{threshold}}$}\label{app:wratausensitivity}

A higher $\tau_{\mathrm{threshold}}$ is expected to estimate the worst-case function $F$ with higher accuracy, while requiring more $f$-calls for WRA. On the other hand, when $\tau_{\mathrm{threshold}}$ is set to a smaller value, WRA will spent fewer $f$-calls, and the estimation accuracy of the worst-case function $F$ is expected to be lower, leading to a difficulty for the proposed approaches to converge at the optimal solution.

The results of the sensitive analysis on $\tau_{\mathrm{threshold}}$ are shown in \Cref{fig:p2tau}.

The number of $f$-calls spent by \wracma{} depended on the setting of $\tau_{\mathrm{threshold}}$, however, the differences were at most the factor of two when $\tau_\mathrm{threshold} \in [0.3, 0.7]$. On $f_5$, we have observed a clear trend of the efficacy as we expected.
When $\tau_{\mathrm{threshold}} \leq 0.7$, the performance of \wracma{} was degraded on $f_5$, whereas it is more or less constant on $f_1$. 
The reason for such a constant behavior on $f_1$ may be because the candidates of the worst-case scenarios are at the corners of the domain $\mathbb{Y}$, independently of solution candidates. Therefore, once such scenario vectors are maintained in $N_\omega$ scenarios, the ranking of the worst-case objective function can be estimated accurately with the initial scenario vectors in WRA. In such a case, the estimated worst-case ranking will not change significantly from its initial estimates and it will results in $\tau \geq 0.7$ at the first round. 

The numbers of $f$-calls spent by \wraslsqp{} were nearly constant over different values of $\tau_{\mathrm{threshold}}$. 
This may be due to the fact that $f_1$ and $f_5$ are concave and linear with respect to $y$ and they can be maximized easily by AGA.
Then, the estimated ranking of the solution candidates on the worst-case objective quickly converges and $\tau$ will be $1$. 
In such a situation, the performance will not change for $\tau_\mathrm{threshold} < 1$.

\begin{figure}[t]
\centering
\begin{subfigure}{0.48\hsize}%
    \includegraphics[width=\hsize]{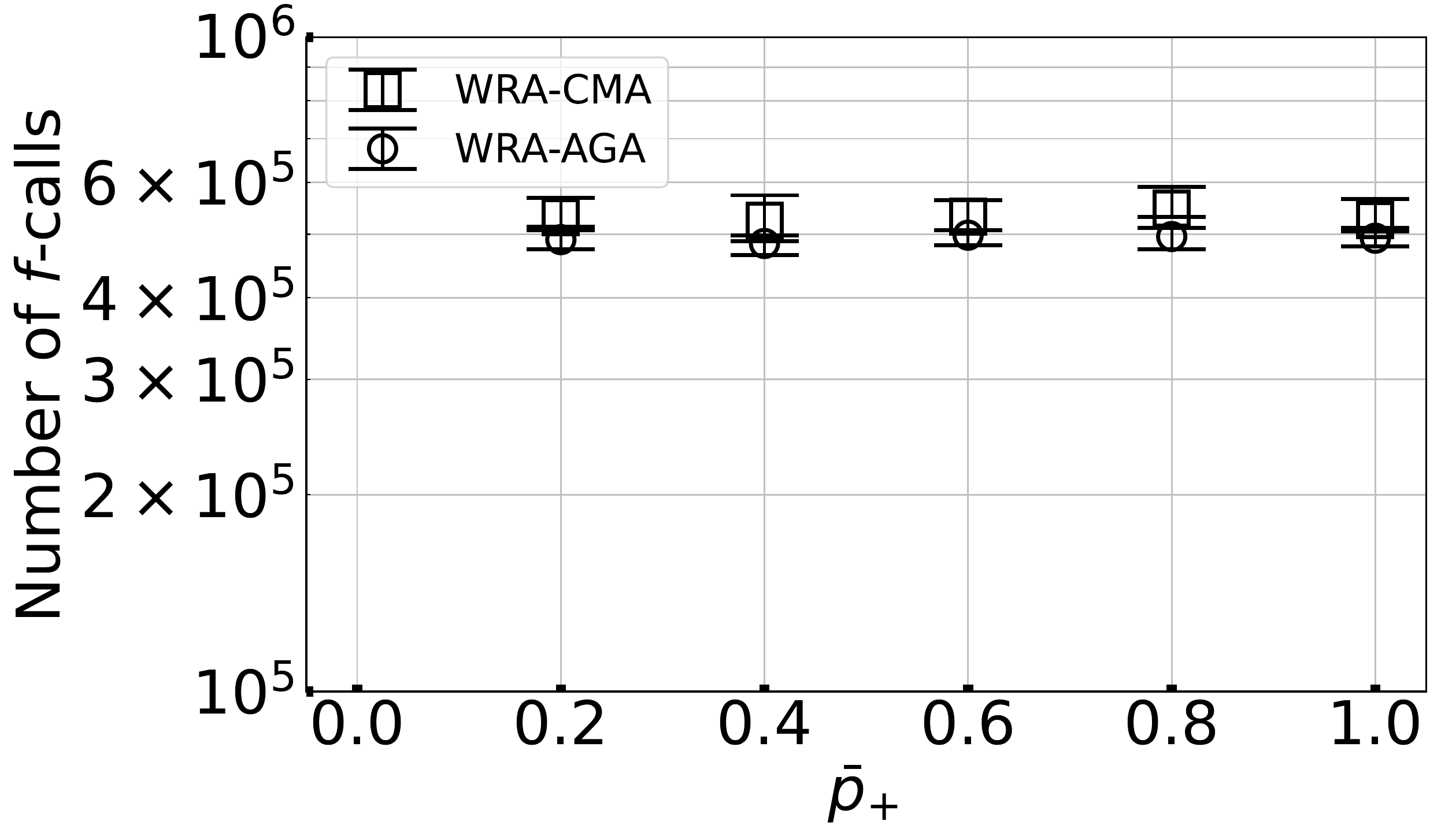}%
  \caption{$f_1$}%
\end{subfigure}%
\hfill
\begin{subfigure}{0.48\hsize}%
    \includegraphics[width=\hsize]{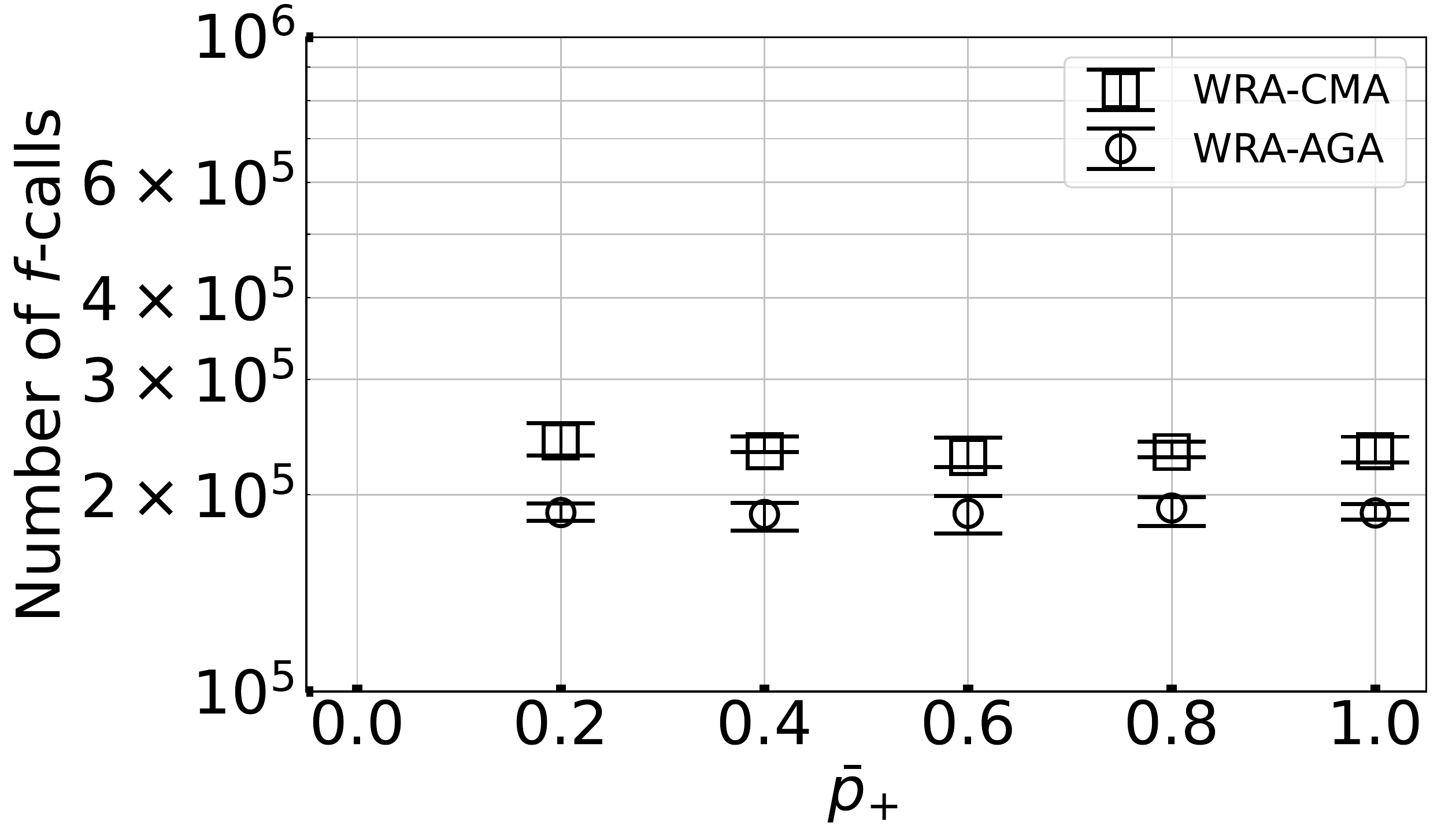}%
  \caption{$f_5$}%
\end{subfigure}%
\caption{
Median and interquartile range of the number of $f$-calls over 20 trials obtained from \wracma{} and \wraslsqp{} with $\bar{p}_{+} \in \{1, 0.8, 0.6, 0.4, 0.2\}$.
}%
\label{fig:p2pp}
\end{figure}

\begin{figure}[t]
\centering
\begin{subfigure}{0.48\hsize}%
    \includegraphics[width=\hsize]{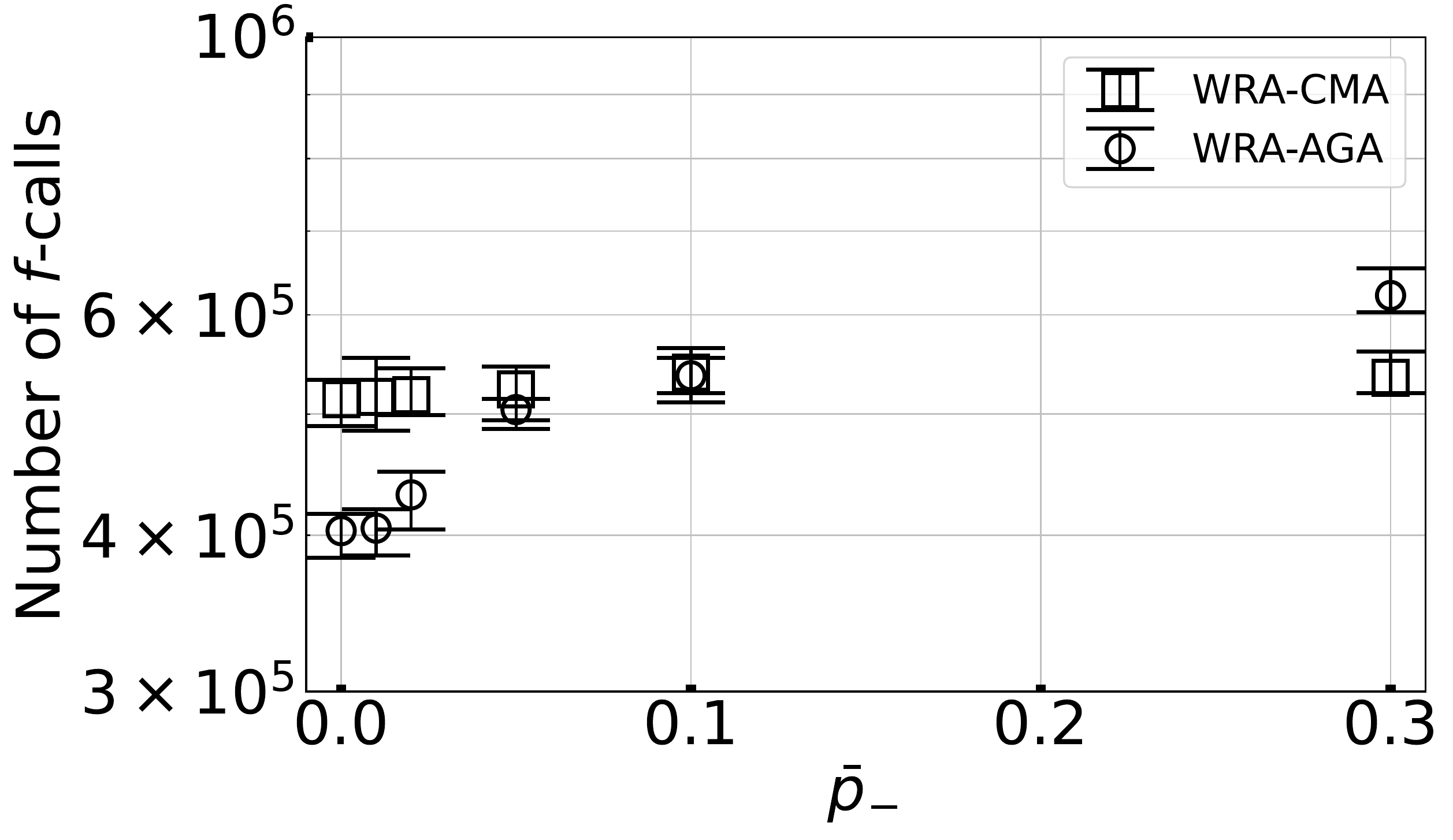}%
  \caption{$f_1$}%
\end{subfigure}%
\hfill
\begin{subfigure}{0.48\hsize}%
    \includegraphics[width=\hsize]{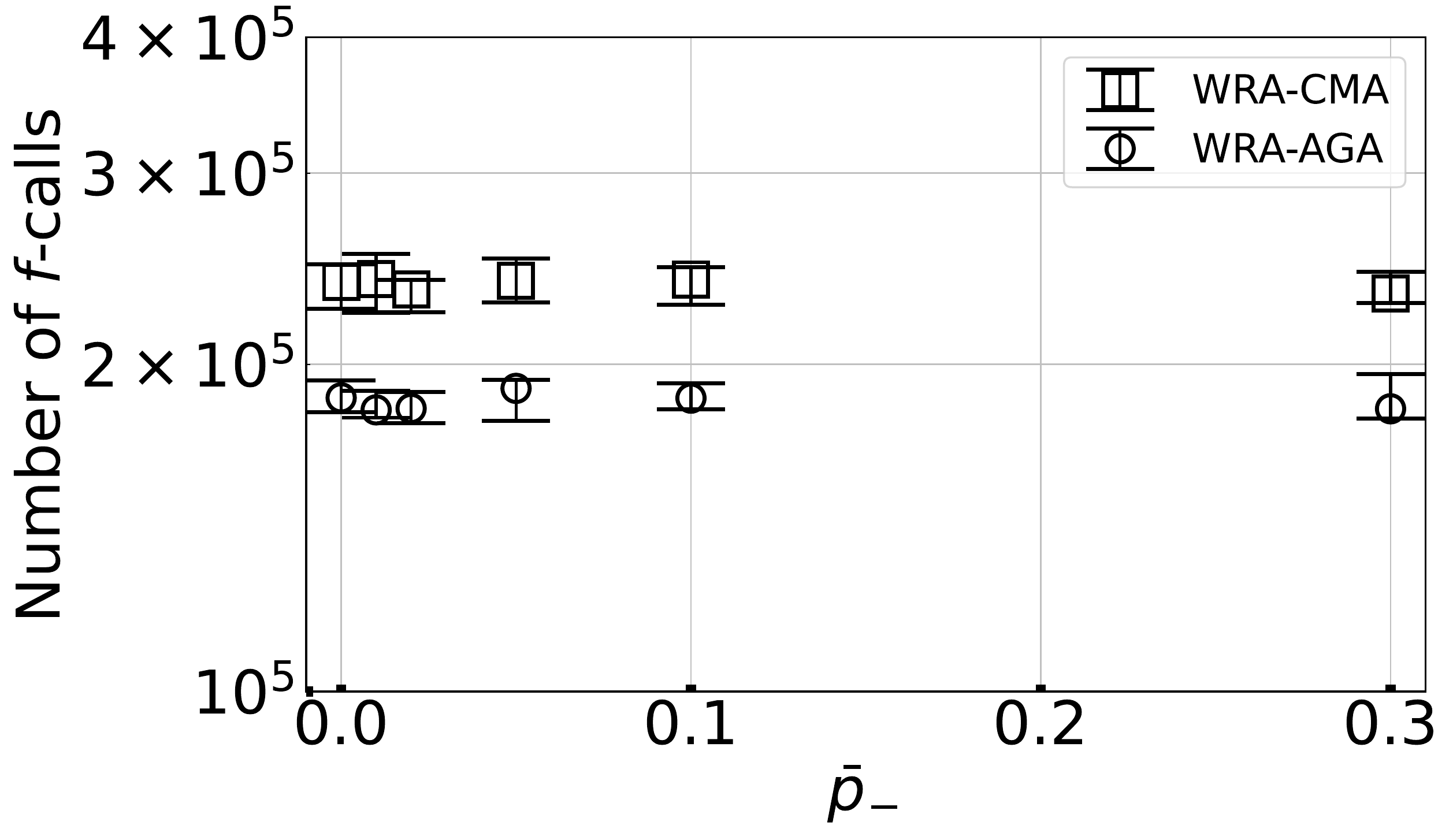}%
  \caption{$f_5$}%
\end{subfigure}%
\caption{
Median and interquartile range of the number of $f$-calls over 20 trials obtained from \wracma{} and \wraslsqp{} with $\bar{p}_{-} \in \{0.01, 0.02, 0.05, 0.1, 0.3\}$.
}%
\label{fig:p2pn}
\end{figure}


\subsection{Sensitivity to $\bar{p}_+$ and 
$\bar{p}_{-}$}\label{app:wrapsensitivity}

The frequency of refreshing configurations $\{(y_k, \omega_k)\}_{k=1}^{\ny}$ is considered to have the following impacts. 
If the configurations are too frequently refreshed, the warm starting strategy will become less effective. 
The frequency of refreshing configurations is controlled by $\bar{p}_{+}$, $\bar{p}_{-}$, and $p_\mathrm{threshold}$. 
The minimum number of iterations that a configuration is refreshed after the last use is given by $(\bar{p}_{+} - p_{\mathrm{threshold}}) / \bar{p}_{-}$. In this study, we fixed $p_{\mathrm{threshold}} = 0.1$ and changed $\bar{p}_{+}$ and $\bar{p}_{-}$. 

\Cref{fig:p2pp} and \Cref{fig:p2pn} shows the results of sensitivity analysis on $f_1$ and $f_5$. On both problems, we can confirm that the proposed approach is not sensitive to the change of $\bar{p}_{+}$ and $\bar{p}_{-}$. On $f_5$, the reason is simply because the refreshing strategy is not necessary as a single configuration is sufficient for this problem. Indeed, one among $N_\omega$ configurations has been selected as the worst case scenario almost all the time during optimization.
On $f_1$, because the objective function is linear with respect to $y$, searching for the worst-case scenario from an inherited worst-case scenario candidate that are not close to the worst-case scenario for a given $x$ and searching from a randomly refreshed scenario will requires nearly the same number of $f$-calls to locate a near worst-case scenario.

The effect of refreshing strategy is expected to appear when the objective function is multimodal with respect to $y$. Further investigation is required into this direction.

\section{
[R2C13]
Scalability analysis} \label{app:p2scal}

We investigate the efficiency of \wracma{} and \wraslsqp{} on $f_1$ and $f_5$ with various dimensions ${d_y}$ and ${d_x}$. 
The experimental setting is the same as in \Cref{sec:common} except that the coefficient matrix $B$ of the interaction term $x^T B y$ is set to a band matrix with the band width of $\abs{{d_y}-{d_x}}+1$ and the band elements are all $1$.

\subsection{Scalability to ${d_x}$}

\begin{figure}[t]
\centering
\begin{subfigure}{0.48\hsize}
    \includegraphics[width=\hsize]{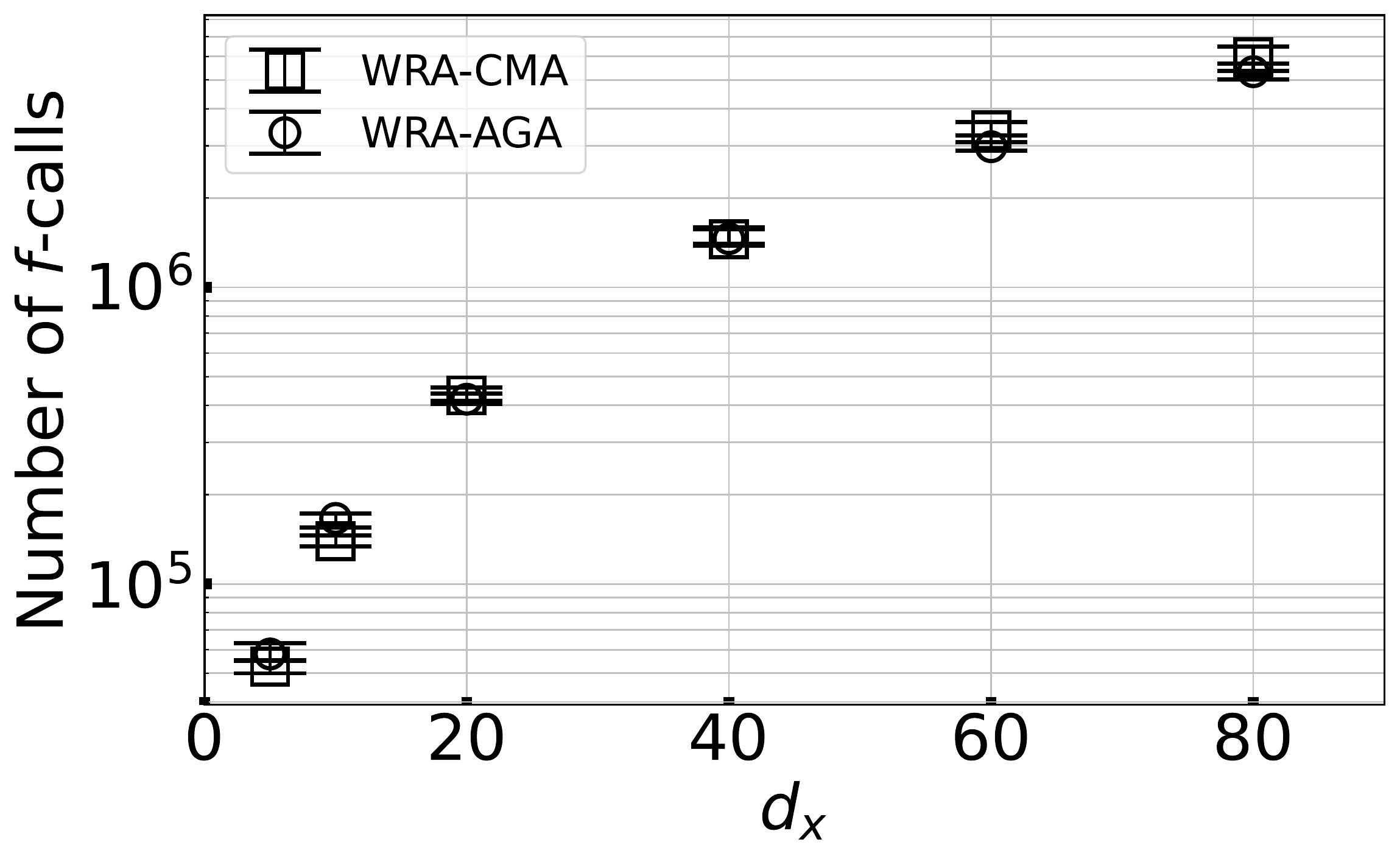}%
 \caption{$f_{1}$}%
\end{subfigure}
\hfill
\begin{subfigure}{0.48\hsize}
    \includegraphics[width=\hsize]{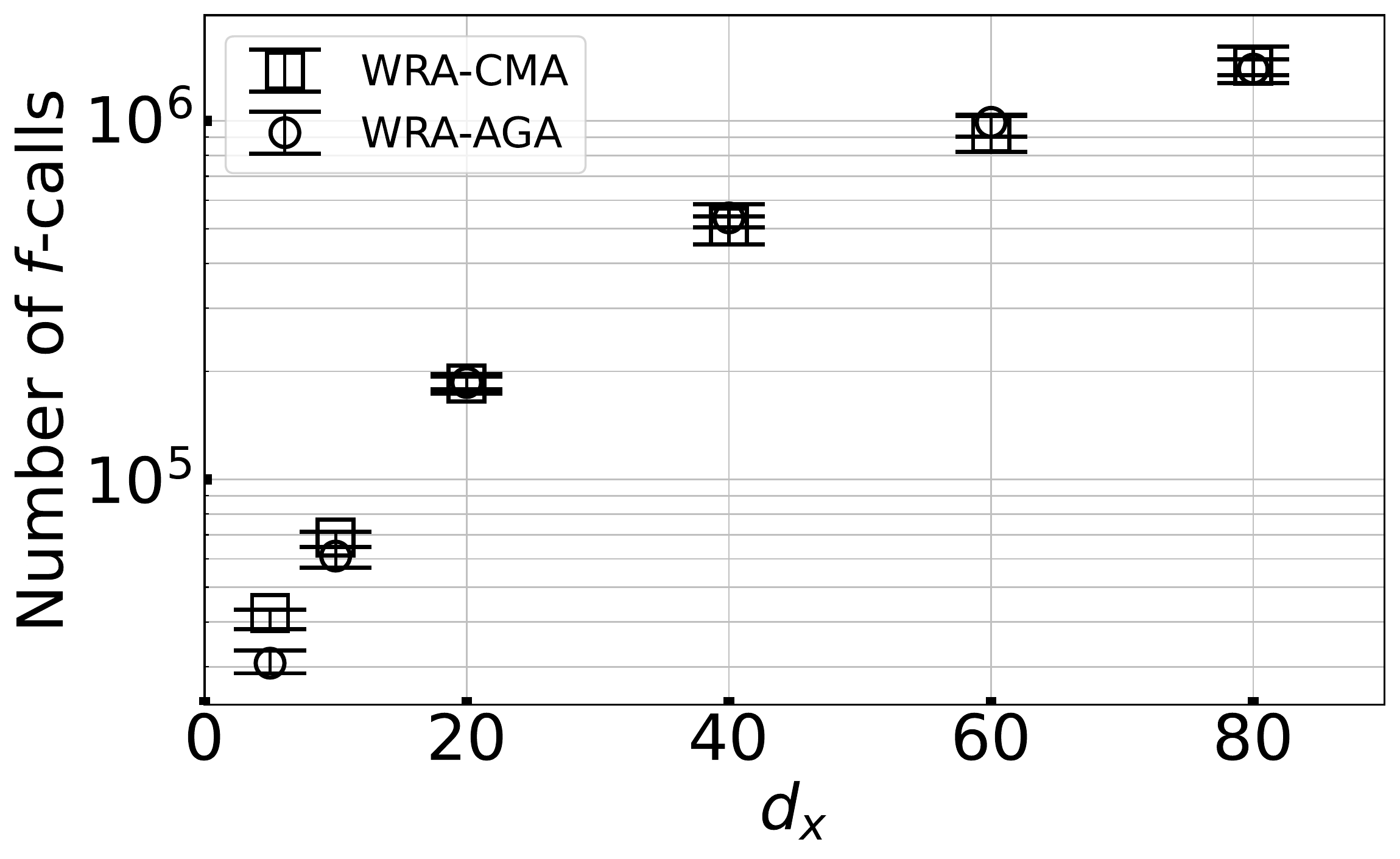}%
 \caption{$f_{5}$}%
\end{subfigure}
\caption{Median and interquartile range of the number of $f$-calls on ${d_x} \in \{5,10,20,40,60,80\}$ over $20$ trials.
}
\label{fig:p2scalx}
\end{figure}

\Cref{fig:p2scalx} shows the results of \wracma{} and \wraslsqp{} on $f_1$ and $f_5$ with different $d_x \in \{5,10,20,40,60,80\}$ and ${d_y}=10$.
The numbers of $f$-calls spent by these approaches increase as $d_x$ increases.
On $f_5$, the increase is near-linearly with respect to $d_x$.
This may be understood naturally as the CMA-ES used for the outer-minimization requires $O(d_x)$ $f$-calls on convex quadratic functions.
On the other hand, it is more than linearly on $f_1$.
This may be understood as the defect of the cumulative step-size adaptation (CSA) on problems with no-effective dimensions.
On $f_1$, the worst-case objective function has the effective dimension of $\mathrm{rank}(B) = \min(d_x, d_y)$ and it is $10$ if $d_x \geq 10$.
On such problem, it has been reported in \cite{akimoto2022cmaes} that the existence of no effective dimension slows down the convergence of the CSA used in the CMA-ES.

\begin{figure}[t]
\centering
\begin{subfigure}{0.48\hsize}
    \includegraphics[width=\hsize]{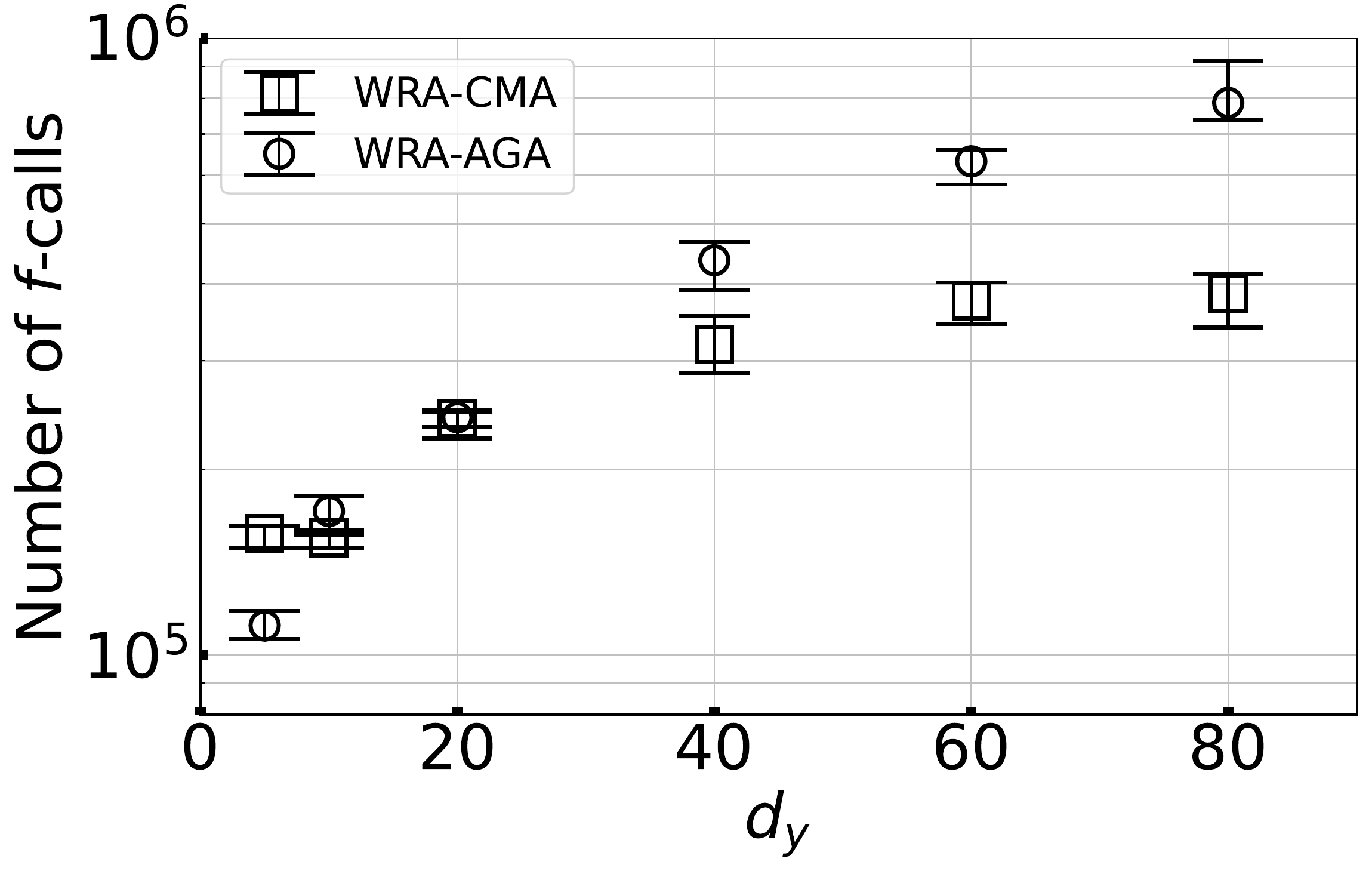}%
 \caption{$f_{1}$}%
\end{subfigure}
\hfill
\begin{subfigure}{0.48\hsize}
    \includegraphics[width=\hsize]{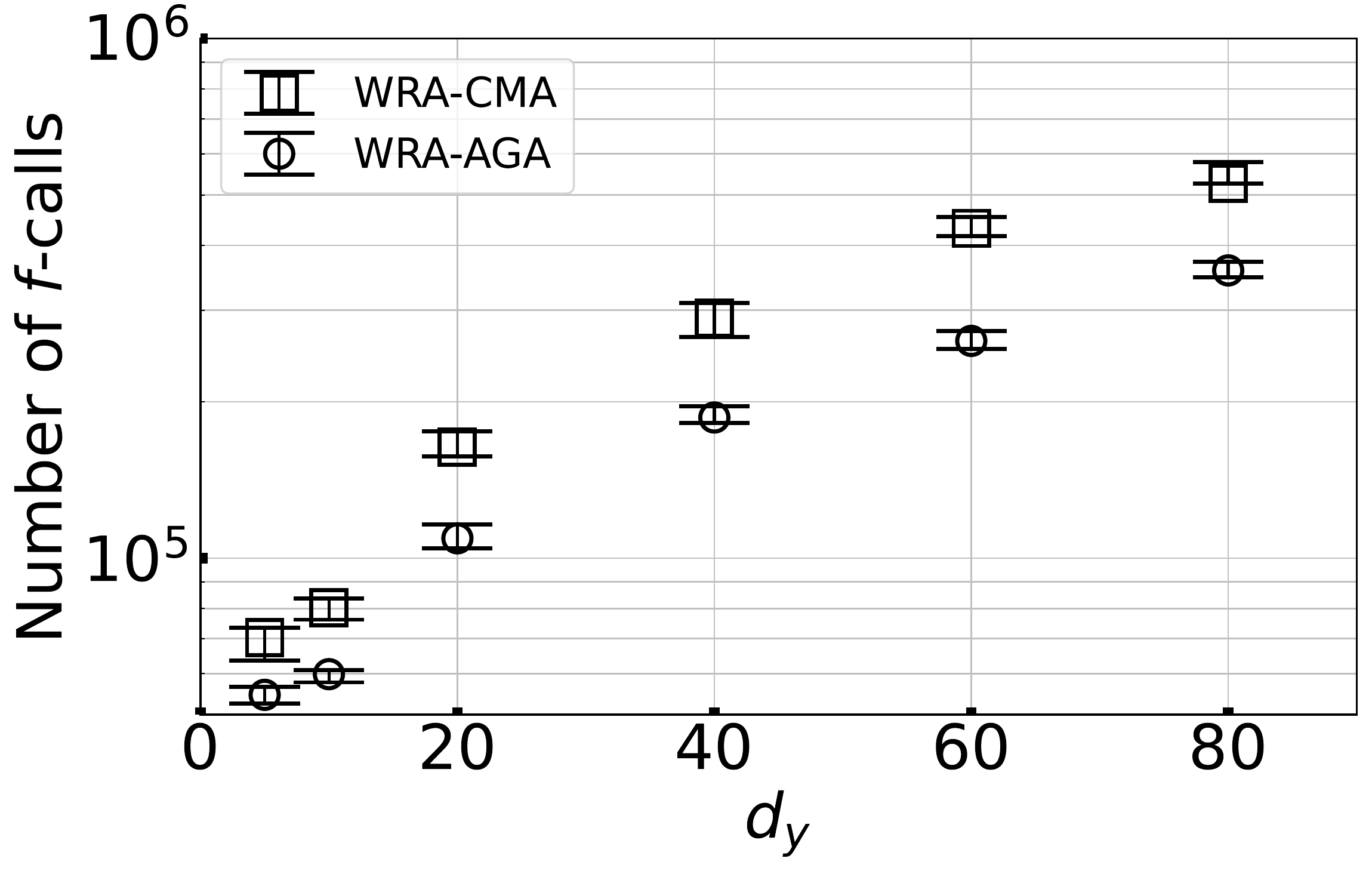}%
 \caption{$f_{5}$}%
\end{subfigure}
\caption{Median and interquartile range of the number of $f$-calls on ${d_y} \in \{5,10,20,40,60,80\}$ over $20$ trials.
}
\label{fig:p2scaly}
\end{figure}

\subsection{Scalability to ${d_y}$}\label{sec:p2scaly}

The scalability of $f$-calls spent by \wracma{} and \wraslsqp{} is investigated as the dimension of the scenario vector ${d_y}$ changes in $f_1$ and $f_5$.
The results on ${d_y} \in \{5, 10, 20, 40, 60, 80\}$ with fixing ${d_x}=10$ are shown in \Cref{fig:p2scaly}.


First, we focus on the results of \wraslsqp{}. 
The number of $f$-calls scaled up near linearly with respect to $d_y$ when $d_{y} \geq 10$. 
This may be simply because that $d_y$ $f$-calls are required to approximate a gradient at each update. 

Second, we focus on the results of \wracma{}.
Similarly to the results of \wraslsqp{}, the number of $f$-calls scaled up near linearly with respect to $d_y$ on $f_5$. 
Because the CMA-ES requires $O(d_y)$ $f$-calls to solve a convex quadratic function, this may be understood as such an effect. 
On the other hand, the number of $f$-calls on $f_1$ did not increased for $d_y \geq 40$. 
This may be because the effective dimension of the objective function with respect to $y$ is $\mathrm{rank}(B) = \min(d_x, d_y)$ and it is $10$ if $d_y \geq 10$. Differently from the result of the scalability analysis to $d_x$, the defect of CSA was considered successfully avoided by the early stopping strategy of WRA.

\end{document}